\documentclass[sigconf]{aamas}

\usepackage{balance} %

\usepackage{subcaption}         %
\usepackage{xcolor}
\usepackage{amsthm}

\usepackage[colorinlistoftodos]{todonotes}
\usepackage{enumitem}
\usepackage{multirow}
\usepackage[most]{tcolorbox}
\usepackage{longtable}
\usepackage{tablefootnote}

\usepackage{pifont}%

\definecolor{softbluegray}{HTML}{F2F6F9}
\definecolor{softbluegray1}{HTML}{D9E2EC}
\definecolor{softbluegray2}{HTML}{BCCCDC}
\definecolor{softbluegray3}{HTML}{607992}
\definecolor{textbluegray}{HTML}{629999}

\newtheoremstyle{decisionstyle}
  {0.6\baselineskip}   %
  {0.6\baselineskip}   %
  {\itshape}           %
  {}                   %
  {\bfseries\color{black}} %
  {.}                  %
  {0.5em}              %
  {}                   %

\theoremstyle{decisionstyle}
\newtheorem{decisionrule}{Decision Rule}

\definecolor{textbluegray}{RGB}{34,94,140}  

\newtheoremstyle{diagnosticstyle}
  {0.6\baselineskip}   %
  {0.6\baselineskip}   %
  {\normalfont}        %
  {}                   %
  {\bfseries\color{textbluegray}} %
  {.}                  %
  {0.5em}              %
  {%
    \thmname{#1}\thmnumber{ #2}%
    \thmnote{ \textnormal{(#3)}}%
  }

\theoremstyle{diagnosticstyle}
\newtheorem{diagnosticbold}{Diagnostic}

\usepackage{tikz}
\usetikzlibrary{fit, backgrounds}
\newcommand{\highlight}[3][softbluegray3]{%
\tikz[baseline=(X.base)]{
\node[
  fill=#1,
  fill opacity=#2,
  text opacity=1,
  text=black,
  rounded corners=2pt,
  inner xsep=2pt,
  inner ysep=1pt
] (X) {#3};
}}

\usetikzlibrary{calc,positioning,decorations.pathreplacing,arrows.meta,fit}

\definecolor{varA}{HTML}{2563EB}    %
\definecolor{varB}{HTML}{059669}    %
\definecolor{condC}{HTML}{D97706}   %
\definecolor{condClight}{HTML}{FEF3C7}
\definecolor{varAlight}{HTML}{DBEAFE}
\definecolor{varBlight}{HTML}{D1FAE5}
\definecolor{boxborder}{HTML}{4B5563}
\definecolor{timeline}{HTML}{F3F4F6}
\definecolor{tlborder}{HTML}{D1D5DB}
\definecolor{arrcol}{HTML}{374151}

\doi{ECCJ1033}

\makeatletter
\gdef\@copyrightpermission{
  \begin{minipage}{0.2\columnwidth}
   \href{https://creativecommons.org/licenses/by/4.0/}{\includegraphics[width=0.90\textwidth]{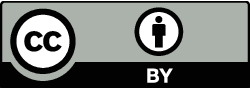}}
  \end{minipage}\hfill
  \begin{minipage}{0.8\columnwidth}
   \href{https://creativecommons.org/licenses/by/4.0/}{This work is licensed under a Creative Commons Attribution International 4.0 License.}
  \end{minipage}
  \vspace{5pt}
}
\makeatother

\setcopyright{ifaamas}
\acmConference[AAMAS '26]{Proc.\@ of the 25th International Conference
on Autonomous Agents and Multiagent Systems (AAMAS 2026)}{May 25 -- 29, 2026}
{Paphos, Cyprus}{C.~Amato, L.~Dennis, V.~Mascardi, J.~Thangarajah (eds.)}
\copyrightyear{2026}
\acmYear{2026}
\acmDOI{}
\acmPrice{}
\acmISBN{}

\acmSubmissionID{1235}

\title[AAMAS-2026 Formatting Instructions]{Probing Dec-POMDP Reasoning in Cooperative MARL}

\author{Kale-ab Abebe Tessera}
\affiliation{
  \institution{University of Edinburgh}
  \city{Edinburgh}
  \country{United Kingdom}}
\email{k.tessera@ed.ac.uk}

\author{Leonard Hinckeldey}
\affiliation{
  \institution{University of Edinburgh}
  \city{Edinburgh}
  \country{United Kingdom}}
\email{l.hinckeldey@ed.ac.uk}

\author{Riccardo Zamboni}
\affiliation{
  \institution{Politecnico di Milano}
  \city{Milan}
  \country{Italy}}
\email{riccardo.zamboni@polimi.it}

\author{David Abel}
\affiliation{
  \institution{University of Edinburgh}
  \city{Edinburgh}
  \country{United Kingdom}}
\email{david.abel@ed.ac.uk}

\author{Amos Storkey}
\affiliation{
  \institution{University of Edinburgh}
  \city{Edinburgh}
  \country{United Kingdom}}
\email{a.storkey@ed.ac.uk}

\begin{abstract}
Cooperative multi-agent reinforcement learning (MARL) is typically framed as a decentralised partially observable Markov decision process (Dec-POMDP), a setting whose hardness stems from two key challenges: \textit{partial observability} and \textit{decentralised coordination}. Genuinely solving such tasks requires \textit{Dec-POMDP reasoning}, where agents use history to infer hidden states and coordinate based on local information. Yet it remains unclear whether popular benchmarks actually demand this reasoning or permit success via simpler strategies. We introduce a diagnostic suite combining statistically grounded performance comparisons and information-theoretic probes to audit the behavioural complexity of baseline policies (IPPO and MAPPO) across 37 scenarios spanning MPE, SMAX, Overcooked, Hanabi, and MaBrax. Our diagnostics reveal that success on these benchmarks rarely requires genuine Dec-POMDP reasoning. Reactive policies match the performance of memory-based agents in over half the scenarios, and emergent coordination frequently relies on brittle, synchronous action coupling rather than robust temporal influence. These findings suggest that some widely used benchmarks may not adequately test core Dec-POMDP assumptions under current training paradigms, potentially leading to over-optimistic assessments of progress. We release our diagnostic tooling to support more rigorous environment design and evaluation in cooperative MARL.\footnote{The code is available at \url{https://github.com/KaleabTessera/probing-dec-pomdps}.}
\end{abstract}


\newcommand{\BibTeX}{\rm B\kern-.05em{\sc i\kern-.025em b}\kern-.08em\TeX}

\begin{document}

\pagestyle{fancy}
\fancyhead{}

\maketitle

\section{Introduction}

The widespread deployment of autonomous multi-agent systems is bounded by their ability to coordinate under uncertainty. In such settings, no single agent possesses a complete view of the world, yet outcomes depend on joint behaviour. This tension lies at the heart of cooperative multi-agent reinforcement learning~\citep[MARL,][]{albrech2024multiagent}. The standard formalism for these problems, decentralised partially observable Markov decision processes~\citep[Dec-POMDPs,][]{bernstein2002complexity,oliehoek2016concise}, capture this intrinsic hardness through two fundamental characteristics: \emph{partial observability}, where agents cannot directly observe the full global state, and \emph{decentralised coordination}, where agents must cooperate based on local and private information.

The intrinsic hardness of this setting stems directly from the interaction of these two factors. In principle, to act optimally, each agent must recover a \emph{Markovian signal} by maintaining a \emph{multi-agent belief} over the joint state and the policies (or histories) of other agents~\citep{oliehoek2016concise}. However, exact multi-agent belief computation is typically infeasible~\citep{bernstein2002complexity}. Consequently, practical model-free methods approximate this reasoning using finite-memory or recurrent policies (e.g., GRUs)~\citep{hausknecht2015deep}, often instantiated within the \emph{centralised training with decentralised execution}~\citep[CTDE,][]{oliehoek2008optimal,kraemer2016multi} paradigm to leverage extra information during learning.

The empirical success of MARL approaches in benchmarks~\citep[among others,][]{yu2022surprising,papoudakis2020benchmarking} is often interpreted as evidence that practical approximations (e.g., recurrent policies) effectively capture the Dec-POMDP reasoning these problems demand. \emph{We challenge this interpretation.} High returns can mask a failure to learn the underlying coordination challenge, as agents may exploit reactive shortcuts permitted by the task design rather than employing genuine history-based reasoning. This distinction is critical. If valid solutions exist that ignore the theoretical challenges of partial observability and coordination, then the environment can become a weak proxy for the Dec-POMDP formalism, yielding an illusion of progress on coordination under uncertainty. We therefore use trained policies as \emph{diagnostic probes} to ask:
\begin{tcolorbox}[colback=softbluegray, colframe=softbluegray,  boxrule=0.5pt, arc=4pt, width=\linewidth, before skip=10pt]\begin{center} \emph{Do modern cooperative MARL environments truly test the Dec-POMDP properties that make these problems hard, or do they permit success via strategies that bypass them?}\end{center}\end{tcolorbox}

To answer this, we introduce a suite of \emph{MARL diagnostics} that couple statistically grounded performance comparisons with information-theoretic probes to measure history dependence, private information flow, synchronous action coupling, and directed temporal influence. Together, these reveal whether learned policies genuinely employ Dec-POMDP reasoning, or bypass it entirely.

We apply these diagnostics to policies learned by standard baselines in \emph{37 popular MARL scenarios}, across MPE~\citep{lowe2017multi}, SMAX\footnote{Both SMAC-V1~\citep{samvelyan2019starcraft} and SMAC-V2~\citep{ellis2023smacv2} maps were tested.}~\citep{rutherford2023jaxmarl}, Overcooked (V1 and V2)~\citep{carroll2019utility,gessler2025overcookedv}, Hanabi~\citep{bard2020hanabi} and MaBrax~\citep{rutherford2023jaxmarl,peng2021facmac}.  Across these settings, our analysis reveals three main takeaways: \textbf{(i)}~history dependence rarely translates to history utility---while all learned policies encode some history dependence, only 43\% actually need memory to achieve high returns, indicating that current observations often suffice for strong performance; \textbf{(ii)}~hidden environment state and hidden teammate information act as separate drivers of difficulty, which our metrics successfully disentangle (e.g., empirically validating the design shift from Overcooked V1 to V2); and \textbf{(iii)}~while coordination is common, its structure is highly variable---synchronous and temporal mechanisms dissociate across benchmarks. Notably, MPE emerges as the only suite where every scenario satisfies all four diagnostic criteria, consistently requiring both meaningful history use and decentralised coordination.

Ultimately, these findings suggest that, under current training paradigms, success on popular benchmarks often does not require the Dec-POMDP reasoning these tasks are intended to evaluate.

\smallskip
\noindent\textbf{Contributions.}
\begin{enumerate}[leftmargin=*, nosep, itemsep=2pt]
\item \textbf{Diagnostic framework.} We introduce information-theoretic probes -- measuring history dependence, private information flow, synchronous action coupling, and directed temporal influence -- that audit whether learned policies actually exhibit Dec-POMDP reasoning, beyond what raw returns reveal.
\item \textbf{Systematic benchmark audit.} We evaluate 37 scenarios across seven benchmark suites, revealing that history dependence is ubiquitous but rarely performance-critical, coordination structures vary qualitatively across domains, and few environments jointly test both partial observability and coordination.
\item\textbf{Open-source tooling and implications.} We release diagnostic tools for researchers to audit their own environments, and discuss implications for designing tasks where partial observability and coordination are non-optional.
\end{enumerate}

\section{Background}
\label{sec:background}

We introduce key concepts that will be needed throughout the paper.

\vspace{3pt}
\noindent\textbf{Interaction Protocol.}~~As a base model for interaction, we consider a discounted Dec-POMDP~\citep{bernstein2002complexity}, defined by the tuple $\mathcal{M} = (\mathcal N, \mathcal{S}, \mathbb{T}, \mathbb{O}, \mu, \{ \mathcal{A}^i \}_{i\in \mathcal N}, \{ \mathcal{O}^i \}_{i \in \mathcal N}, R, \gamma)$. Here, $\mathcal{N}$ is the set of $N \in \mathbb{N}$ agents and $\mathcal{S}$ is the set of global states. At each time step $t$, the system is in some state $s_t \in \mathcal{S}$. Each agent $i \in \mathcal{N}$ selects an action $a_t^i \in \mathcal{A}^i$, forming a joint action $\mathbf{a}_t = (a_t^1, \dots, a_t^N)$ in the joint action space $\mathcal{A} = \times_{i=1}^N \mathcal{A}^i$. This action leads to a state transition according to the probability function $\mathbb{T}(s_{t+1}|s_t, \mathbf{a}_t)$ and a shared reward $R(s_t, \mathbf{a}_t)$. Agents do not observe the global state $s_t$, instead they receive a local observation $o_t^i \in \mathcal{O}^i$. The joint observation $\mathbf{o}_t$ is drawn according to the observation function $\mathbb{O}(\mathbf{o}_t|s_t, \mathbf{a}_{t-1})$. The goal is to learn a joint policy $\boldsymbol{\pi}$ at which no agent has any incentive to deviate, while maximising the expected discounted return $\mathbb{E}_{\mathbf{a}_t \sim \boldsymbol{\pi}, \mathcal M} \left[ \sum_{t=0}^{\infty} \gamma^t R(s_t, \mathbf{a}_t) \right]$. These solution concepts are usually described through various notions of \emph{equilibria}:  we report a brief description in  Appendix~\ref{apx:solution_concepts}. %

\vspace{3pt}
\noindent\textbf{Mutual Information.}~~To study the information embedded in agents' policies, we propose metrics based on mutual information (MI). 
For two discrete random variables $X$ and $Y$ with joint probability mass function\footnote{For continuous variables, we use the probability density function.} $p(x,y)$ and marginals $p(x)$, $p(y)$, we can measure MI as follows:
\begin{align}
    \mathbb{I}(X; Y) &= H(X)- \mathbb{H}(X\mid Y) = \mathbb{H}(Y)- \mathbb{H}(Y\mid X), \\
    &= \sum_{x,y} p(x,y) \log \frac{p(x,y)}{p(x)p(y)}.
\end{align}
where $H$ is the Shannon entropy. Intuitively, $\mathbb{I}(X;Y)$ is the average amount of information that $X$ conveys about $Y$, or vice versa. MI is symmetric and non-negative, and $\mathbb{I}(X;Y)=0$ iff $X$ and $Y$ are independent. 

We will also use metrics based on conditional mutual information (CMI), $\mathbb{I}(X; Y \mid Z )$. Intuitively, CMI measures the extra information that $X$ tells us about $Y$, excluding what we know about $Y$ given $Z$. $\mathbb{I}(X; Y \mid Z )=0$ iff $X$ and $Y$ are conditionally independent given $Z$.

\definecolor{varA}{HTML}{2563EB}          %
\definecolor{varB}{HTML}{059669}          %
\definecolor{condC}{HTML}{D97706}         %
\definecolor{condClight}{HTML}{FEF3C7}
\definecolor{varAlight}{HTML}{DBEAFE}
\definecolor{varBlight}{HTML}{D1FAE5}
\definecolor{boxborder}{HTML}{6B7280}
\definecolor{timeline}{HTML}{F9FAFB}
\definecolor{tlborder}{HTML}{D1D5DB}
\definecolor{arrcol}{HTML}{374151}

\begin{figure*}[t]
\centering
\resizebox{\textwidth}{!}{%
\begin{tikzpicture}[
    >=Stealth,
    cell/.style={
        minimum width=0.56cm, minimum height=0.40cm,
        inner sep=0pt, outer sep=0pt,
        font=\fontsize{5.8}{7}\selectfont,
        draw=tlborder, thin, fill=timeline,
    },
    hlA/.style  ={fill=varAlight,  draw=varA,  semithick},
    hlB/.style  ={fill=varBlight,  draw=varB,  semithick},
    hlCond/.style={fill=condClight, draw=condC, semithick, densely dashed},
    fbox/.style ={draw=boxborder, rounded corners=2pt,
                  inner sep=2.5pt, font=\small, fill=white, thin},
    agl/.style  ={font=\fontsize{5.8}{7}\selectfont\bfseries, text=black!55,
                  anchor=east},
    subl/.style ={font=\fontsize{6.5}{8}\selectfont\bfseries, text=black!80},
    dot/.style  ={font=\fontsize{5.8}{7}\selectfont},
]

\def\cA{0}     \def\cB{0.56}  \def\cC{1.12}  \def\cD{1.68}
\def\dotL{2.16}
\def\cE{2.64}  \def\cF{3.20}
\def\dotR{3.68}
\def\cG{4.16}  \def\cH{4.72}
\def\pctr{2.36}

\def\colII{6.4}
\def\colIII{12.8}
\def\rowII{-3.80}

\def\subcapI{-2.2}
\def\subcapII{-2.20}

\def\agX{-0.25}

\begin{scope}
  \node[agl] at (\agX,0) {Ag.\,$i$};
  \node[cell]         at (\cA,0) {$O_1^i$};
  \node[cell]         at (\cB,0) {$A_1^i$};
  \node[cell]         at (\cC,0) {$O_2^i$};
  \node[cell]         at (\cD,0) {$A_2^i$};
  \node[dot]          at (\dotL,0) {$\cdots$};
  \node[cell,hlA] (aO) at (\cE,0) {$O_t^i$};
  \node[cell,hlB] (aA) at (\cF,0) {$A_t^i$};
  \node[dot]          at (\dotR,0) {$\cdots$};
  \node[cell]         at (\cG,0) {$O_T^i$};
  \node[cell]         at (\cH,0) {$A_T^i$};
  \draw[->,arrcol,thick,shorten >=1.5pt,shorten <=1.5pt]
    (aO.south) .. controls +(0,-0.32) and +(-0.1,-0.5) .. (aA.south);
  \node[fbox] at (\pctr,-0.90)
    {$\mathrm{OAR}\triangleq
      \mathbb{I}(\textcolor{varA}{O_t^i}\,;\,\textcolor{varB}{A_t^i})$};
  \node[subl] at (\pctr,\subcapI) {(a) Observation--Action Relevance (OAR)};
\end{scope}

\begin{scope}[shift={(\colII,0)}]
  \node[agl] at (\agX,0) {Ag.\,$i$};
  \node[cell,hlA] (bO1)  at (\cA,0) {$O_1^i$};
  \node[cell,hlA]        at (\cB,0) {$A_1^i$};
  \node[cell,hlA]        at (\cC,0) {$O_2^i$};
  \node[cell,hlA] (bA2)  at (\cD,0) {$A_2^i$};
  \node[dot,text=varA]   at (\dotL,0) {$\cdots$};
  \node[cell,hlCond]     at (\cE,0) {$O_t^i$};
  \node[cell,hlB]  (bAt) at (\cF,0) {$A_t^i$};
  \node[dot]             at (\dotR,0) {$\cdots$};
  \node[cell]            at (\cG,0) {$O_T^i$};
  \node[cell]            at (\cH,0) {$A_T^i$};
  \draw[decorate,decoration={brace,amplitude=3pt,mirror},varA,semithick]
    ([yshift=-2pt]bO1.south west) -- ([yshift=-2pt]bA2.south east)
    node[midway,below=4pt,font=\fontsize{5.5}{6.5}\selectfont,text=varA]
    {$H_t^i$};
  \draw[->,arrcol,thick,shorten >=1.5pt]
    ([yshift=-0.34cm]bA2.south west)
    .. controls +(0.45,-0.1) and +(-0.1,-0.5) .. (bAt.south);
  \node[fbox] at (\pctr,-1.18)
    {$\mathrm{HAR}\triangleq
      \mathbb{I}(\textcolor{varA}{H_t^i}\,;\,
      \textcolor{varB}{A_t^i}\mid\textcolor{condC}{O_t^i})$};
  \node[subl] at (\pctr,\subcapI) {(b) History--Action Relevance (HAR)};
\end{scope}

\begin{scope}[shift={(\colIII,0)}]
  \node[agl] at (\agX,0.44) {Ag.\,$i$};
  \node[cell,hlA] (ci1)    at (\cA,0.44) {$O_1^i$};
  \node[cell,hlA]           at (\cB,0.44) {$A_1^i$};
  \node[dot,text=varA]     at (\dotL,0.44) {$\cdots$};
  \node[cell,hlA] (ci4)    at (\cD,0.44) {$A_{t\text{-}1}^i$};   %
  \node[cell,hlA] (ci5)    at (\cE,0.44) {$O_t^i$};              %
  \node[cell]              at (\cF,0.44) {$A_t^i$};
  \node[dot]               at (\dotR,0.44) {$\cdots$};
  \node[cell]              at (\cG,0.44) {$O_T^i$};
  \node[cell]              at (\cH,0.44) {$A_T^i$};
  \draw[decorate,decoration={brace,amplitude=3pt},varA,semithick]
    ([yshift=2pt]ci1.north west) -- ([yshift=2pt]ci4.north east)
    node[midway,above=4pt,font=\fontsize{5.5}{6.5}\selectfont,text=varA]
    {$\tau_{t-1}^i$};
  \node[agl] at (\agX,-0.44) {Ag.\,$j$};
  \node[cell,hlCond] (cj1)  at (\cA,-0.44) {$O_1^j$};
  \node[cell,hlCond]        at (\cB,-0.44) {$A_1^j$};
  \node[dot,text=condC]     at (\dotL,-0.44) {$\cdots$};
  \node[cell,hlCond] (cj4)  at (\cD,-0.44) {$A_{t\text{-}1}^j$};  %
  \node[cell,hlCond] (cj5)  at (\cE,-0.44) {$O_t^j$};             %
  \node[cell,hlB]   (cj6)   at (\cF,-0.44) {$A_t^j$};
  \node[dot]                at (\dotR,-0.44) {$\cdots$};
  \node[cell]               at (\cG,-0.44) {$O_T^j$};
  \node[cell]               at (\cH,-0.44) {$A_T^j$};
  \draw[decorate,decoration={brace,amplitude=3pt,mirror},condC,semithick]
    ([yshift=-2pt]cj1.south west) -- ([yshift=-2pt]cj4.south east)
    node[midway,below=4pt,font=\fontsize{5.5}{6.5}\selectfont,text=condC]
    {$\tau_{t-1}^j$};
  \draw[->,arrcol,thick,shorten >=1.5pt,shorten <=2pt]
    (ci5.south) .. controls +(0.12,-0.32) and +(-0.12,0.45) .. (cj6.north);
  \node[fbox] at (\pctr,-1.58)
    {$\mathrm{PIF}_{i\to j}\triangleq
      \mathbb{I}\big(\textcolor{varA}{(\tau_{t-1}^i,\, O_t^i)}\,;\,
      \textcolor{varB}{A_t^j}\,\big|\,\textcolor{condC}{(\tau_{t-1}^j,\, O_t^j)}\big)$};
  \node[subl] at (\pctr,\subcapI) {(c) Private Information Flow (PIF)};
\end{scope}

\begin{scope}[shift={(0,\rowII)}]
  \node[agl] at (\agX,0.58) {Ag.\,$i$};
  \node[cell]              at (\cA,0.58) {$O_1^i$};
  \node[cell]              at (\cB,0.58) {$A_1^i$};
  \node[cell]              at (\cC,0.58) {$O_2^i$};
  \node[cell]              at (\cD,0.58) {$A_2^i$};
  \node[dot]               at (\dotL,0.58) {$\cdots$};
  \node[cell,hlCond]       at (\cE,0.58) {$O_t^i$};
  \node[cell,hlA]   (dAi)  at (\cF,0.58) {$A_t^i$};
  \node[dot]               at (\dotR,0.58) {$\cdots$};
  \node[cell]              at (\cG,0.58) {$O_T^i$};
  \node[cell]              at (\cH,0.58) {$A_T^i$};
  \node[agl] at (\agX,-0.58) {Ag.\,$j$};
  \node[cell]              at (\cA,-0.58) {$O_1^j$};
  \node[cell]              at (\cB,-0.58) {$A_1^j$};
  \node[cell]              at (\cC,-0.58) {$O_2^j$};
  \node[cell]              at (\cD,-0.58) {$A_2^j$};
  \node[dot]               at (\dotL,-0.58) {$\cdots$};
  \node[cell,hlCond]       at (\cE,-0.58) {$O_t^j$};
  \node[cell,hlB]   (dAj)  at (\cF,-0.58) {$A_t^j$};
  \node[dot]               at (\dotR,-0.58) {$\cdots$};
  \node[cell]              at (\cG,-0.58) {$O_T^j$};
  \node[cell]              at (\cH,-0.58) {$A_T^j$};
  \draw[<->,arrcol,thick,shorten >=1.5pt,shorten <=1.5pt]
    (dAi.south) -- (dAj.north);
  \node[fbox] at (\pctr,-1.30)
    {$\mathrm{AA}\triangleq
      \mathbb{I}\big(\textcolor{varA}{A_t^i}\,;\,
      \textcolor{varB}{A_t^j}\,\big|\,\textcolor{condC}{O_t^i,\,O_t^j}\big)$};
  \node[subl] at (\pctr,\subcapII) {(d) Action--Action Coupling (AA)};
\end{scope}

\begin{scope}[shift={(\colII,\rowII)}]
  \node[agl] at (\agX,0.44) {Ag.\,$i$};
  \node[cell,hlA] (ei1)    at (\cA,0.44) {$O_1^i$};
  \node[cell,hlA] (ei2)    at (\cB,0.44) {$A_1^i$};
  \node[dot,text=varA]     at (\dotL,0.44) {$\cdots$};
  \node[cell,hlA]          at (\cD,0.44) {$O_{t\text{-}1}^i$};   %
  \node[cell,hlA] (ei5)    at (\cE,0.44) {$A_{t\text{-}1}^i$};   %
  \node[cell]              at (\cF,0.44) {$A_t^i$};
  \node[dot]               at (\dotR,0.44) {$\cdots$};
  \node[cell]              at (\cG,0.44) {$O_T^i$};
  \node[cell]              at (\cH,0.44) {$A_T^i$};
  \draw[decorate,decoration={brace,amplitude=3pt},varA,semithick]
    ([yshift=2pt]ei1.north west) -- ([yshift=2pt]ei5.north east)
    node[midway,above=4pt,font=\fontsize{5.5}{6.5}\selectfont,text=varA]
    {$\tau_{t-1}^i$};
  \node[agl] at (\agX,-0.44) {Ag.\,$j$};
  \node[cell,hlCond] (ej1)  at (\cA,-0.44) {$O_1^j$};
  \node[cell,hlCond]        at (\cB,-0.44) {$A_1^j$};
  \node[dot,text=condC]     at (\dotL,-0.44) {$\cdots$};
  \node[cell,hlCond]        at (\cD,-0.44) {$O_{t\text{-}1}^j$};
  \node[cell,hlCond] (ej5)  at (\cE,-0.44) {$A_{t\text{-}1}^j$};
  \node[cell,hlB]   (ej6)   at (\cF,-0.44) {$A_t^j$};
  \node[dot]                at (\dotR,-0.44) {$\cdots$};
  \node[cell]               at (\cG,-0.44) {$O_T^j$};
  \node[cell]               at (\cH,-0.44) {$A_T^j$};
  \draw[decorate,decoration={brace,amplitude=3pt,mirror},condC,semithick]
    ([yshift=-2pt]ej1.south west) -- ([yshift=-2pt]ej5.south east)
    node[midway,below=3pt,font=\fontsize{5.5}{6.5}\selectfont,text=condC]
    {$\tau_{t-1}^j$};
  \draw[->,arrcol,thick,shorten >=1.5pt,shorten <=2pt]
    (ei5.south) .. controls +(0.12,-0.32) and +(-0.12,0.45) .. (ej6.north);
  \draw[->,arrcol!50,semithick,shorten >=1pt,shorten <=1pt,densely dotted]
    ([xshift=1pt]ei2.south east) .. controls +(0.12,-0.45) and +(-0.16,0.30) ..
    ([xshift=-4pt]ej6.north west);
  \draw[->,arrcol!50,semithick,shorten >=1pt,shorten <=1pt,densely dotted]
    (\dotR,0.22)  .. controls +(0.12,-0.26) and +(-0.14,0.45) ..
    ([xshift=-2pt]{\cH},-0.24);
  \node[fbox] at (\pctr,-1.6)
    {$\mathrm{DAI}_{i\to j}\triangleq
      \textstyle \frac{1}{T}\sum_{t=1}^{T}\mathbb{I}\big(\textcolor{varA}{\tau_{t-1}^i}\,;\,
      \textcolor{varB}{A_t^j}\,\big|\,\textcolor{condC}{\tau_{t-1}^j}\big)$};
  \node[subl] at (\pctr,\subcapII) {(e) Directed Action Information (DAI)};
\end{scope}

\begin{scope}[shift={(\colIII,\rowII)}]
  \node[anchor=north west, inner sep=0pt] at (-0.1, 0.80) {%
    \fontsize{7}{8}\selectfont
    \setlength{\tabcolsep}{3pt}%
    \renewcommand{\arraystretch}{1.15}%
    \begin{tabular}{@{}l@{\hspace{4pt}}p{6cm}@{}}
    \toprule
    & \textbf{If high, suggests\ldots} \\
    \midrule
    \textcolor{varA}{\textbf{OAR}}
      & Actions are largely predictable from the current observations, consistent with reactive policies. \\
    \textcolor{varA}{\textbf{HAR}}
      & History helps predict actions beyond the current observation $O_t^i$. \\ 
    \textcolor{varA}{\textbf{PIF}} &
    Agent $i$'s private information adds predictive information about $j$'s action beyond $j$'s own history. \\
    \textcolor{varA}{\textbf{AA}}
    & Residual same-timestep action coupling not explained by $O_t^i, O_t^j$, consistent with conventions or symmetry breaking. \\ 
    
    \textcolor{varA}{\textbf{DAI}} & Agent $i$'s past carries additional predictive information about $j$'s future actions beyond $j$'s own past. \\

    \bottomrule
    \end{tabular}%
  };
\end{scope}

\end{tikzpicture}
}
\caption{Summary of information-theoretic diagnostics.
Colours denote \textcolor{varA}{\textbf{source}} (blue),
\textcolor{varB}{\textbf{target}} (green), and
\textcolor{condC}{\textbf{conditioning}} (amber, dashed) variables.
All quantities are expectations under the converged joint policy $p^{\boldsymbol{\pi}}$;
$O_t^i$ is agent~$i$'s observation, $A_t^i$ its action,
$H_t^i$ the local history (RNN hidden state or finite window),
$\tau^{i}_{t-1}$ agent~$i$'s action--observation history up to $t{-}1$,
and $T$ the episode horizon.}
\label{fig:diagnostics}
\end{figure*}
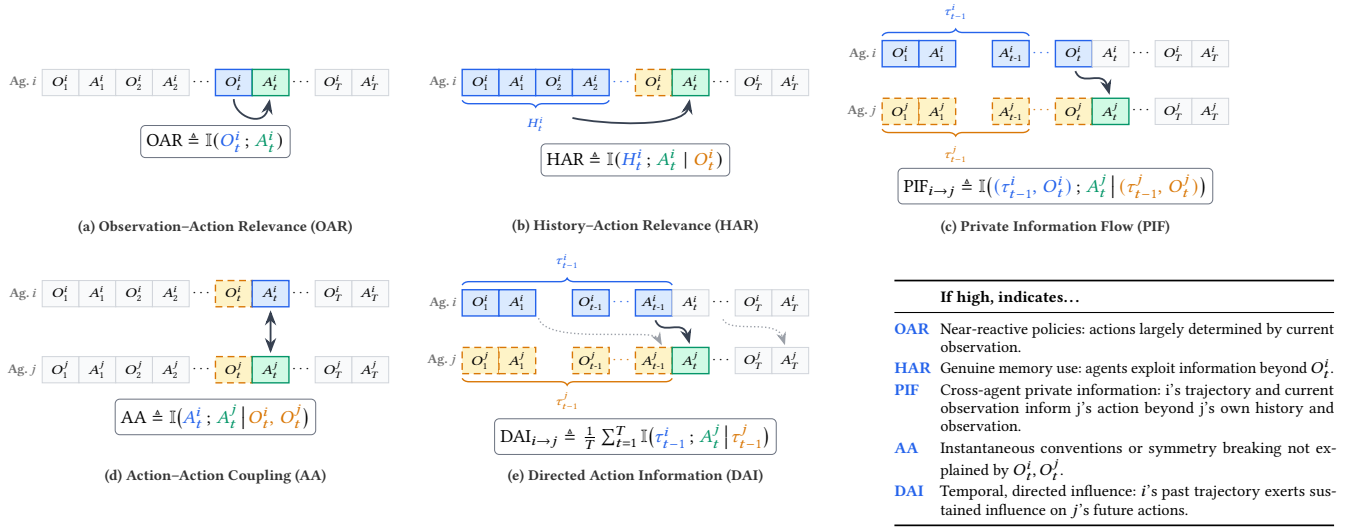

\section{Related Work}
\noindent\textbf{Benchmarking Partial Observability.} \citet{ellis2023smacv2} found that many SMAC~\citep{samvelyan2019starcraft} maps admit open-loop solutions that ignore local observations. While they redesigned these maps to enforce "meaningful partial observability", they provided no metric to quantify it. In single-agent RL, \citet{tao2025benchmarking} formalised \emph{memory improvability} based on performance gaps between agents with access to more or less state information. Our framework provides quantitative tools for the multi-agent case, moving beyond raw performance metrics. We disentangle history dependence, private information flow, and coordination as separate dimensions of Dec-POMDP difficulty.

\noindent\textbf{Conventions.}~~Co-trained agents typically develop conventions that are efficient but arbitrary and brittle when paired with unfamiliar partners~\citep{foerster2019bayesian,hu2020other}. Prior work shows that grounding these conventions in observations makes coordination more robust~\citep{hu2021off}. Our AA and DAI diagnostics explicitly quantify these dynamics, disentangling instantaneous, ungrounded conventions from coordination that is temporally responsive to a partner's trajectory.

\section{Probing Dec-POMDPs}\label{sec:diag_decpomdps}

To probe the reasoning demands specific to MARL environments, we focus on two core properties of Dec-POMDPs -- \emph{partial observability} and \emph{decentralised coordination}. While the interaction of these factors renders the general problem class NEXP-complete\footnote{The worst case complexity of DEC-MDPs is the same as Dec-POMDPs~\citep{bernstein2002complexity}, as such hardness comes from decentralisation as well, and not (only) from the presence of hidden states.}, theoretical worst-case hardness does not necessarily imply practical difficulty in specific benchmarks.

Our goal is therefore to characterise these properties \emph{functionally}, measuring them only \emph{as they matter for solving a task}. Consequently, we define every diagnostic as an expectation under the trajectory distribution of a joint policy $\boldsymbol{\pi}$ after convergence. We do not define \emph{purely structural} properties of Dec-POMDPs independent of behaviour, but rather, we quantify \emph{the specific reasoning capabilities necessitated by the task}. Figure~\ref{fig:diagnostics} presents a summary and interpretation of our proposed diagnostic measures, and we discuss the technical details in the following sections.

\subsection{Partial Observability}\label{diag:po}

\textbf{Is Partial Observability Relevant?}
\vspace{2pt}

\noindent While many environments are \emph{structurally} partially observable (states are hidden), this does not guarantee that the missing information is \emph{functionally relevant} to solving a task. For example, the hidden state may not affect the rewards or dynamics, or it may be redundant given the current observations. 

We are therefore interested in identifying when partial observability strictly affect success. If a task requires memory, it confirms that immediate observations are insufficient and that history contains decision-relevant information. Therefore,  we measure this using \emph{history dependence}.

\begin{definition}[Relevant Partial Observability]\label{def:partial_relevance}
An environment exhibits relevant partial observability if:
\begin{enumerate}[label=(\roman*), nosep]
    \item memory-based agents ($\pi_{\text{RNN}}$) outperform reactive agents ($\pi_{\text{FF}}$) under matched training conditions; and
    \item learned policies actively exploit history, rather than relying solely on immediate observations.
\end{enumerate}
\end{definition}

This definition requires that memory is both \emph{beneficial} (producing higher returns) and \emph{active} (influencing decisions). We quantify this with a performance diagnostic, and with two complementary information-theoretic probes. 

\begin{diagnosticbold}[Memory--Reactive Gap]\label{diag:gap}
We test whether memory results in a performance gain by comparing recurrent and feed-forward policies. For matched training runs (sharing seed, environment, and algorithm), let $J(\pi)$ denote the mean evaluation return. We define the paired performance gap as:
$$ \Delta_{\mathrm{Mem}} \triangleq J(\pi_{\mathrm{RNN}}) - J(\pi_{\mathrm{FF}}). $$
We test $\mathbf{H_0}:\ \mathrm{median}(\Delta_{\mathrm{Mem}})\le 0$ vs.\ $\mathbf{H_1}:\ \mathrm{median}(\Delta_{\mathrm{Mem}})>0$ using a one-sided Wilcoxon signed-rank test~\citep{wilcoxon1945individual} over the paired differences. A significant result ($p<0.05$) indicates a reliable performance advantage from memory under matched training.
\end{diagnosticbold}

\begin{diagnosticbold}[History--Action Relevance (HAR)]\label{diag:har}
We quantify memory use beyond the current observation via conditional mutual information:
\begin{equation}
\mathrm{HAR} \triangleq \mathbb{I}(H_t^i;A_t^i\mid O_t^i),\qquad
\mathrm{HAR}^{\mathrm{norm}} \triangleq
\frac{\mathrm{HAR}}{\mathbb{H}(A_t^i\mid O_t^i)} \in [0,1].
\end{equation}
Here, $H_t^i$ denotes the agent’s history representation, for reactive policies,
$H_t^i = O^i_{t-k:t-1}$ (a length-$k$ window excluding $O_t^i$), and for recurrent policies,
$H_t^i$ is the RNN hidden state.
\end{diagnosticbold}

\begin{diagnosticbold}[Observation--Action Relevance (OAR)]\label{diag:oar}
We quantify reactivity by measuring how informative the current observation is about the agent's action:
\begin{equation}
    \mathrm{OAR} \triangleq \mathbb{I}(O_t^i; A_t^i), \qquad
    \mathrm{OAR}^{\mathrm{norm}} \triangleq \frac{\mathrm{OAR}}{\mathbb{H}(A_t^i)} \in [0,1].
\end{equation}
High $\mathrm{OAR}^{\mathrm{norm}}$ indicates that $A_t^i$ is largely predictable from the current observation $O_t^i$ (i.e., near-reactive behaviour). Conversely, low $\mathrm{OAR}^{\mathrm{norm}}$ combined with high $\mathrm{HAR}^{\mathrm{norm}}$ provides evidence that history contributes information for selecting $A_t^i$ beyond what is contained in $O_t^i$.
\end{diagnosticbold}

\vspace{1pt}
\noindent\textbf{Is Partial Observability Reliant on Private Information?}
\vspace{3pt}

\noindent The previous diagnostics measure whether agents benefit from \emph{history} or \emph{memory}, which acts as a behavioural proxy for functionally relevant partial observability. Crucially, history dependence alone does not imply that the hidden information is relevant for \emph{coordination}. An agent may use its history only to infer latent \emph{environment} state, as in single-agent POMDPs~\citep{aastrom1965optimal,kaelbling1998planning}, even if this provides no additional information about coordinating with teammates.

We therefore introduce a cross-agent diagnostic that quantifies whether the private information of one agent helps predict the actions of another. This metric is related to the intuition behind \emph{meaningful partial observability}~\citep{ellis2023smacv2}, where hidden information observed by one agent is critical for the actions of another. Such cross-agent information asymmetries are central to the hardness of Dec-POMDPs~\citep{bernstein2002complexity}.

\begin{diagnosticbold}[Private Information Flow (PIF)]\label{diag:pif}
We measure how much additional information agent $i$'s history provides about agent $j$'s action, beyond what is already contained in $j$'s own history. We define this using conditional mutual information:

\begin{equation}
\begin{split}
    \mathrm{PIF}_{i \to j} &\triangleq \mathbb{I}(\tau_{t-1}^i, O_t^i;\, A_t^j \mid \tau_{t-1}^j, O_t^j), \\
    \mathrm{PIF}^{\mathrm{norm}}_{i \to j} &\triangleq 
    \frac{\mathrm{PIF}_{i \to j}}{\mathbb{H}(A_t^j \mid \tau_{t-1}^j, O_t^j)} \in [0, 1].
\end{split}
\end{equation}

Here, $\tau_{t-1}$ denotes an agent's action-observation history\footnote{In practice, we approximate $\tau_{t-1}$ using the RNN hidden state (for recurrent policies) or a finite window of size $k$, $(O^i_{t-k:t-1},A^i_{t-k:t-1})$ (for reactive policies).}. We explicitly condition on the current observations $O_t^i, O_t^j$ alongside the past $\tau_{t-1}$ to capture information asymmetries at decision time.

$\mathrm{PIF}_{i \to j}$ quantifies how much information about $A_t^j$ is contained in agent $i$'s trajectory that is \emph{not} already captured by agent $j$. $\mathrm{PIF}^{\mathrm{norm}}_{i \to j}$ rescales this as the fraction of agent $j$'s residual action uncertainty (given its own history $\tau_{t-1}^j$ and observation $O_t^j$) that is explained by agent $i$.

\end{diagnosticbold}

\subsection{Decentralised Coordination}

The previous diagnostics quantify whether \emph{hidden information} is relevant to decision-making, specifically, whether agents require memory of local state (HAR; Diagnostic~\ref{diag:har}) or access to a teammate's private information (PIF; Diagnostic~\ref{diag:pif}). However, they do not characterise the \emph{form} of coordination that emerges in the joint behaviour induced by the converged policies (if any). We therefore introduce coordination probes that separate instantaneous action coupling from temporally extended, more directional dependence. 

\vspace{3pt}
\noindent\textbf{Is Coordination Synchronous?}

\begin{diagnosticbold}[Action–Action Coupling (AA)]
\label{diag:aa}
We quantify \emph{instantaneous} action dependence via the coupling of actions at time $t$:
\begin{equation}
\mathrm{AA} \triangleq \mathbb{I}\left(A_t^i ; A_t^j \mid O_t^i, O_t^j\right), \quad \mathrm{AA}^{\mathrm{norm}}\triangleq\frac{\mathrm{AA}}{\mathbb{H}\left(A_t^j \mid O_t^i,O_t^j\right)}.
\end{equation}
$\mathrm{AA}$ measures same-timestep dependence between agents' actions beyond what their current observations explain, and $\mathrm{AA}>0$ is consistent with symmetry breaking or instantaneous conventions (e.g., agents taking distinct roles such as heading to different landmarks).
\end{diagnosticbold}

\vspace{0.3em}
\noindent\textbf{Is Coordination Temporally Responsive?}
\vspace{2pt}

\noindent $\mathrm{AA}$ alone cannot distinguish \emph{task-driven role differentiation} from \emph{arbitrary, ungrounded conventions}, as it detects instantaneous coupling beyond shared observations, but cannot distinguish static conventions (e.g., fixed roles) from agents adapting to evolving partner behaviours. 

To probe this \emph{temporally extended, directional dependence}, we test whether agent $i$’s \emph{past} provides additional predictive information about agent $j$’s \emph{current} action, conditioned on agent $j$’s own history. While a lagged $\mathrm{AA}$ could measure this, it would rely on fixed windows that are brittle to unknown or variable delays. We instead use \emph{Directed Information}~\citep{massey1990causality}, which aggregates directional cross-timestep dependence over the episode, capturing dependencies regardless of the temporal lag.

\begin{diagnosticbold}[Directed Action Information (DAI)]
\label{diag:di}
We measure the average directional, cross-timestep dependence from agent $i$ to agent $j$ as follows:
\begin{equation}
\begin{split}
   \mathrm{DAI}_{i\to j}\triangleq
\frac{1}{T}\sum_{t=1}^{T} \mathbb{I}\left(\tau^i_{t-1} ; A_t^j \middle| \tau^j_{t-1}\right) ,\\
    \mathrm{DAI}^{\mathrm{norm}}_{i\to j}\triangleq
\frac{\mathrm{DAI}_{i\to j}}{\frac{1}{T}\sum_{t=1}^{T} \mathbb{H}\left(A_t^j \middle| \tau^j_{t-1}\right)} \in [0, 1].
\end{split}
\end{equation}

Here, $\tau^i_{t-1}$ is agent $i$'s action-observation history up to $t-1$, including $A^i_{t-1}$, the last act before agent $j$ selects $A^j_t$. Conditioning on $\tau^j_{t-1}$ controls for what is already predictable from agent $j$'s own past, so $\mathrm{DAI}_{i\to j}>0$ indicates that agent $i$'s past carries additional predictive information about agent $j$'s current action. Unlike PIF, which includes current observations to capture information at decision time, DAI conditions only on the causal past (the trajectory completed before j's action), isolating strictly temporal, directional influence.

\end{diagnosticbold}

\section{Case Study: How Observation Structure shapes Behaviour}\label{sec:case_study}

\begin{figure*}[ht]
\centering
\captionsetup[sub]{font=footnotesize}
\begin{minipage}[b]{0.18\textwidth}
  \centering
  \includegraphics[width=\linewidth]{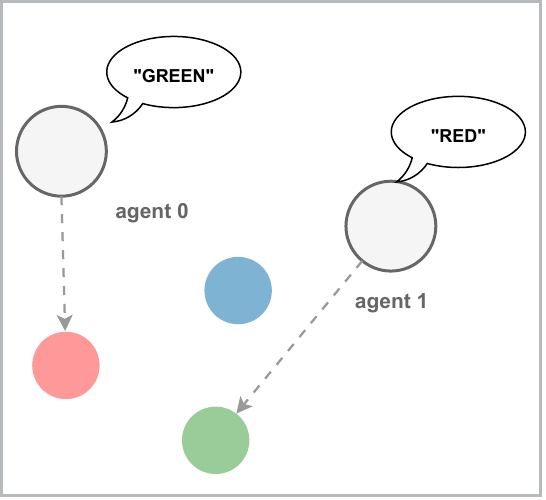}
  \vspace{-4pt}
  \subcaption{Simple Reference}\label{fig:mpe_reference}
\end{minipage}\hfill
\begin{minipage}[b]{0.18\textwidth}
  \centering
  \includegraphics[width=\linewidth]{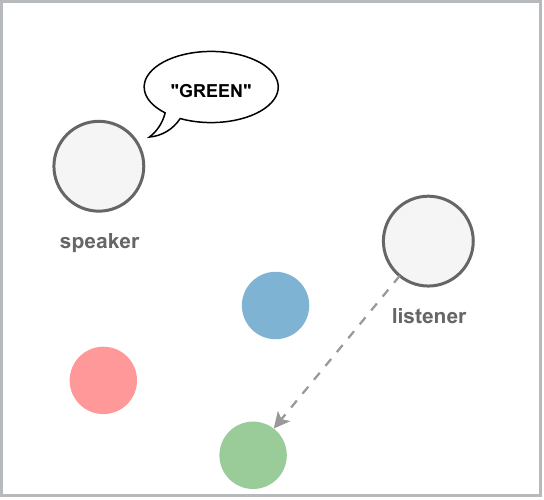}
  \vspace{-4pt}
  \subcaption{Speaker Listener}\label{fig:mpe_speaker}
\end{minipage}\hfill
\begin{minipage}[b]{0.18\textwidth}
  \centering
  \includegraphics[width=\linewidth]{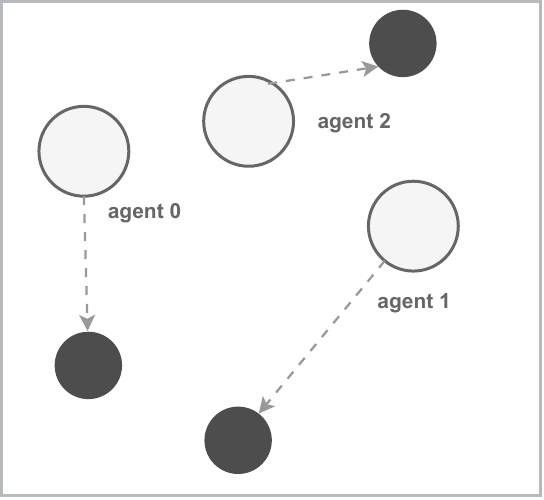}
  \vspace{-4pt}
  \subcaption{Simple Spread}\label{fig:mpe_spread}
\end{minipage}\hfill
\begin{minipage}[b]{0.22\textwidth}
  \centering
  \footnotesize
  \setlength{\tabcolsep}{4pt}
  \renewcommand{\arraystretch}{1.0}
  \begin{tabular}{lc}
    \toprule
    Scenario & $\Delta$ (RNN$-$FF) \\
    \midrule
    \texttt{Simple Reference} & \textbf{6.50} \\
    \texttt{Speaker Listener} & \textbf{14.84} \\
    \texttt{Simple Spread} & \textbf{2.50} \\
    \bottomrule
  \end{tabular}
  \vspace{2pt}
  \subcaption{$\Delta_{\text{Mem}}$; bold: $p{<}0.05$.}
  \label{tab:mpe_per_env_deltas_inline}
\end{minipage}
\caption{MPE tasks and per-environment performance deltas (RNN$-$FF). Memory improves performance across tasks.}
\label{fig:mpe_row_with_table}
\vspace{-12pt}
\end{figure*}

\begin{figure*}[ht]
\centering
\begin{minipage}{\textwidth}
 \centering
        \includegraphics[width=0.6\linewidth]{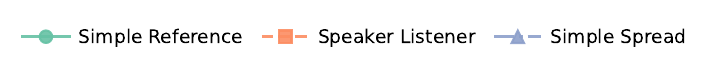}
    \end{minipage}
    \vspace{-6pt}

\begin{subfigure}[b]{0.20\textwidth}
  \centering
  \includegraphics[width=\textwidth]{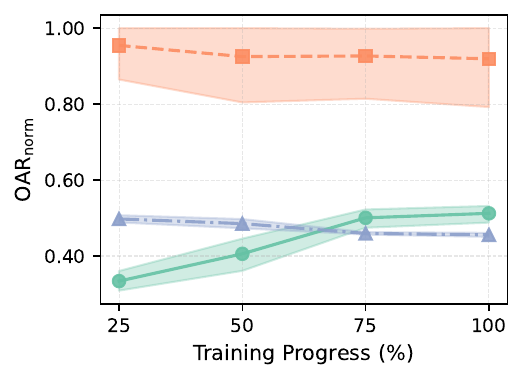}
  \subcaption{Observation–Action Relevance (OAR)}
  \label{fig:diag_oar}
\end{subfigure}\hfill
\begin{subfigure}[b]{0.20\textwidth}
  \centering
  \includegraphics[width=\textwidth]{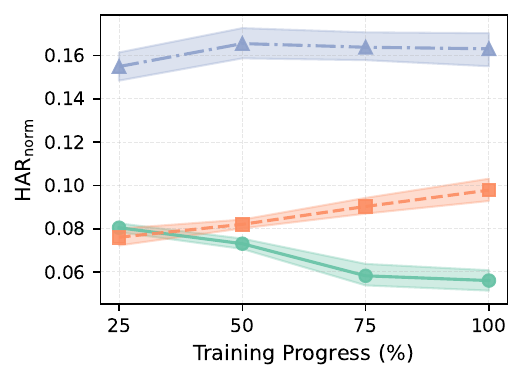}
  \subcaption{History–Action Relevance (HAR)}
  \label{fig:diag_har}
\end{subfigure}\hfill
\begin{subfigure}[b]{0.20\textwidth}
  \centering
  \includegraphics[width=\textwidth]{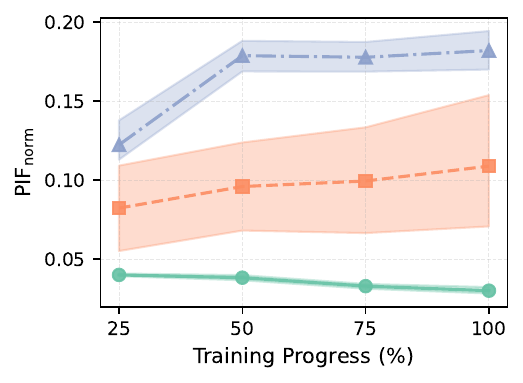}
  \subcaption{Private Information Flow (PIF)}
  \label{fig:diag_pif}
\end{subfigure}\hfill
\begin{subfigure}[b]{0.20\textwidth}
  \centering
  \includegraphics[width=\textwidth]{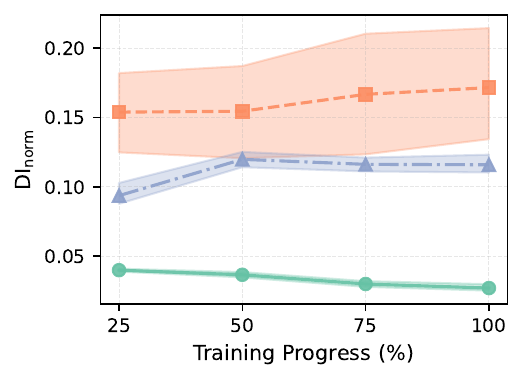}
  \subcaption{Directed Action Information (DAI)}
  \label{fig:diag_dai}
\end{subfigure}

\caption{Evolution of diagnostic metrics \emph{during} training in MPE with recurrent MAPPO (mean and 95\% CI): (\textbf{a}) Observation--Action Relevance, (\textbf{b}) History--Action Relevance, (\textbf{c}) Private Information Flow, and (\textbf{d}) Directed Action Information. For the same algorithm and training paradigm, environment modifications can have a large impact on the kinds of behaviour learned.}
\label{fig:mpe_diagnostics_over_time}
\end{figure*}

Multi-Particle Environments (MPE)~\citep{lowe2017multi} provide a controlled testbed
with differing observation and communication structures. We examine three cooperative tasks---\emph{Simple Reference}, \emph{Speaker--Listener} and \emph{Simple Spread}---using our diagnostics (Section~\ref{sec:diag_decpomdps}) and MAPPO.

\noindent\textbf{Performance.}
We see from Tbl.~\ref{tab:mpe_per_env_deltas_inline}, recurrent policies (RNN) outperform feed-forward (FF) baselines in all three tasks ($p<0.05$, one-tailed Wilcoxon), confirming that memory provides a reliable advantage across MPE.

\noindent\textbf{What the diagnostics reveal.}
Viewing MPE through our diagnostics shows that learned behaviour varies sharply across tasks, not because the algorithm changes, but because the observation/communication structure does.

\emph{Simple Reference (Fig.~\ref{fig:mpe_reference}).} 
In \emph{Simple Reference}, two agents move and observe the \emph{other's} goal alongside a rich communication channel ($\mathrm{dim}_c{=}10$). Goal information is thus redundantly available at every timestep, reducing the need for history: $\mathrm{HAR}^{\mathrm{norm}}$ is the lowest across tasks and declines over training (${\approx}\,0.06$, Fig.~\ref{fig:diag_har}), and $\mathrm{PIF}$/$\mathrm{DAI}$ remain low (Fig.~\ref{fig:diag_pif},~\ref{fig:diag_dai}).

\emph{Speaker--Listener (Fig.~\ref{fig:mpe_speaker}).} 
In this scenario, a stationary speaker observes a hidden goal and must guide a listener that receives no goal information except through a narrow message channel ($\mathrm{dim}_c{=}3$). This dependency produces the highest $\mathrm{DAI}^{\mathrm{norm}}$ across tasks ($>0.15$, Fig.~\ref{fig:diag_dai}), reflecting sustained directional influence from speaker to listener. $\mathrm{HAR}^{\mathrm{norm}}$ rises over training but remains moderate (${\approx}\,0.10$), suggesting that the listener's history use, while present, is secondary to the cross-agent information channel.

\emph{Simple Spread (Fig.~\ref{fig:mpe_spread}).}
Here, agents must cover distinct landmarks without explicit communication. Consequently, $\mathrm{HAR}^{\mathrm{norm}}$ and $\mathrm{PIF}^{\mathrm{norm}}$ are the highest across tasks (Fig.~\ref{fig:diag_har}, Fig.~\ref{fig:diag_pif}), indicating that agents condition on each other's private trajectories to avoid overlapping landmarks. $\mathrm{DAI}^{\mathrm{norm}}$ is also substantial (${\approx}\,0.12$), confirming coordination is both temporally extended and reliant on private information.
\begin{tcolorbox}[colback=softbluegray, colframe=softbluegray,
  boxrule=0.5pt, arc=4pt, width=\linewidth, top=2pt, bottom=2pt, before skip=10pt]
\textbf{Takeaway.}
The form of coordination that emerges is shaped primarily by \textbf{information bottlenecks in the environment}. When task-relevant information is fully available at each timestep (\emph{Simple Reference}), agents default to \textbf{reactive behaviour} despite having recurrent architectures. Conversely, when information is restricted, e.g., funnelled through a narrow channel (\emph{Speaker--Listener}) or left implicit in a partner's trajectory (\emph{Simple Spread}), agents develop \textbf{qualitatively different coordination structures}: higher directional influence in the former, and higher private information flow in the latter. 
\end{tcolorbox}

\section{Results}

\begin{figure*}[ht!]
    \centering
    \vspace{-6pt}
    \begin{minipage}{\textwidth}
        \centering
        \includegraphics[width=0.6\linewidth]{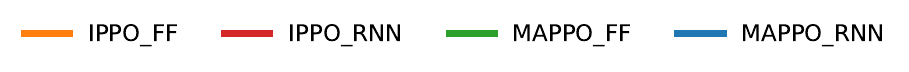}
    \end{minipage}
    
    \vspace{0.5em}
    
    \begin{subfigure}{0.24\textwidth}
        \centering
        \includegraphics[width=\linewidth]{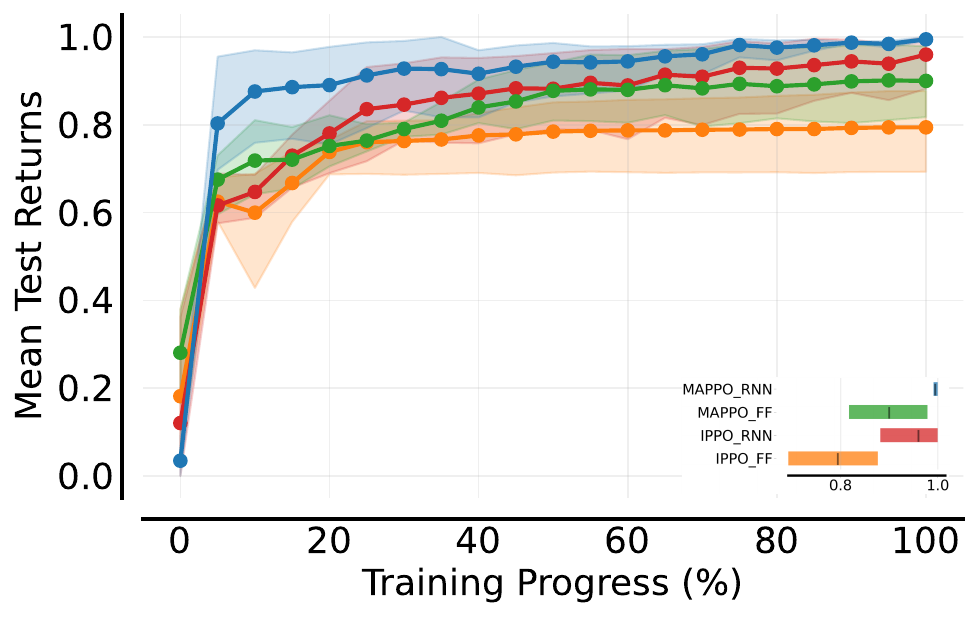}
        \vspace{-0.6em}
        \subcaption{MPE}
        \label{fig:mpe}
    \end{subfigure}
    \hfill
    \begin{subfigure}{0.24\textwidth}
        \centering
        \includegraphics[width=\linewidth]{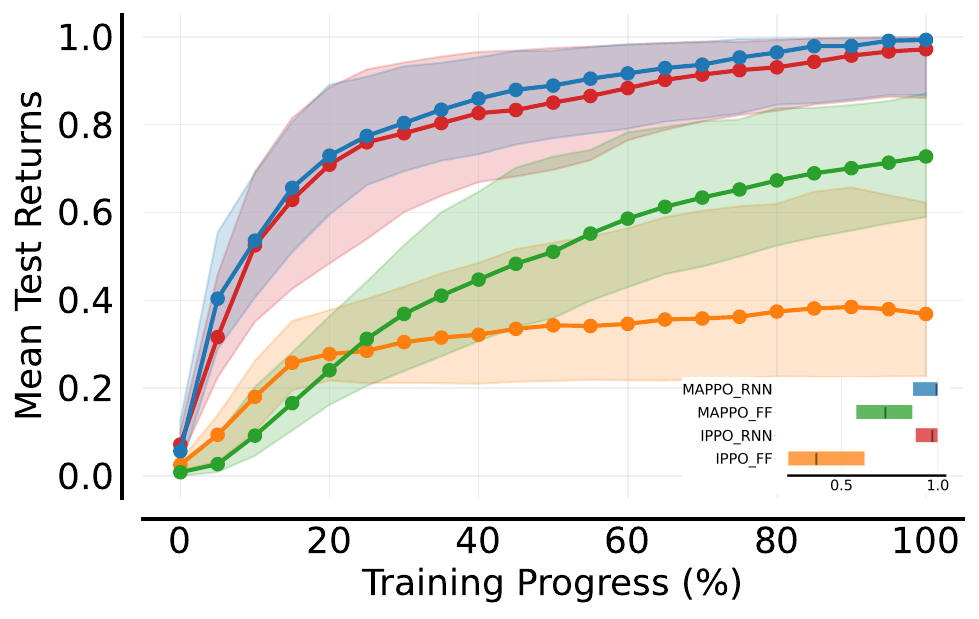}
        \vspace{-0.6em}
        \subcaption{SMAX V1 maps}
        \label{fig:smaxv1}
    \end{subfigure}
    \hfill
    \begin{subfigure}{0.24\textwidth}
        \centering
        \includegraphics[width=\linewidth]{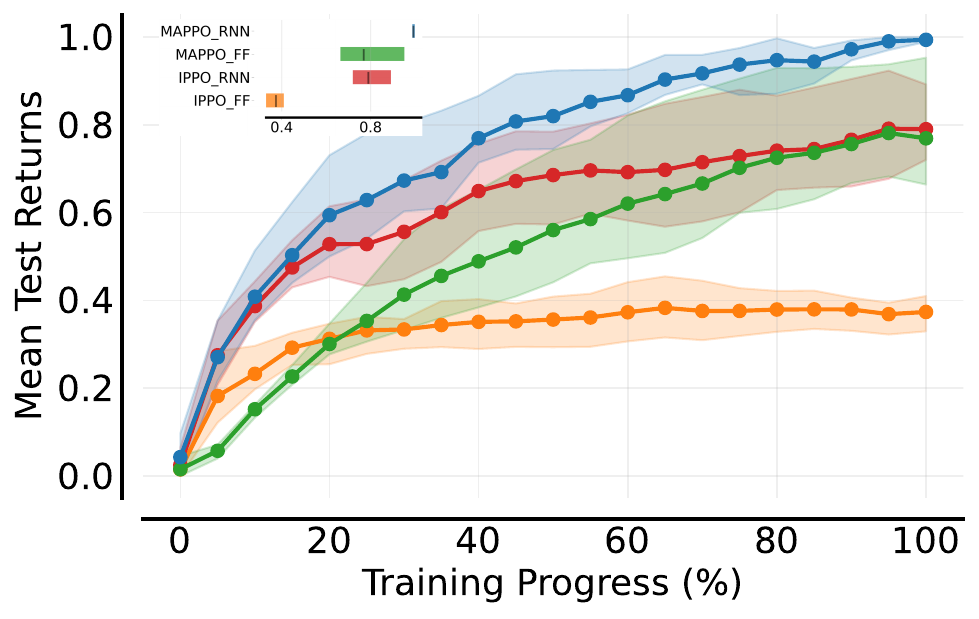}
        \vspace{-0.6em}
        \subcaption{SMAX V2 maps}
        \label{fig:smaxv2}
    \end{subfigure}
    \hfill
    \begin{subfigure}{0.24\textwidth}
        \centering
        \includegraphics[width=\linewidth]{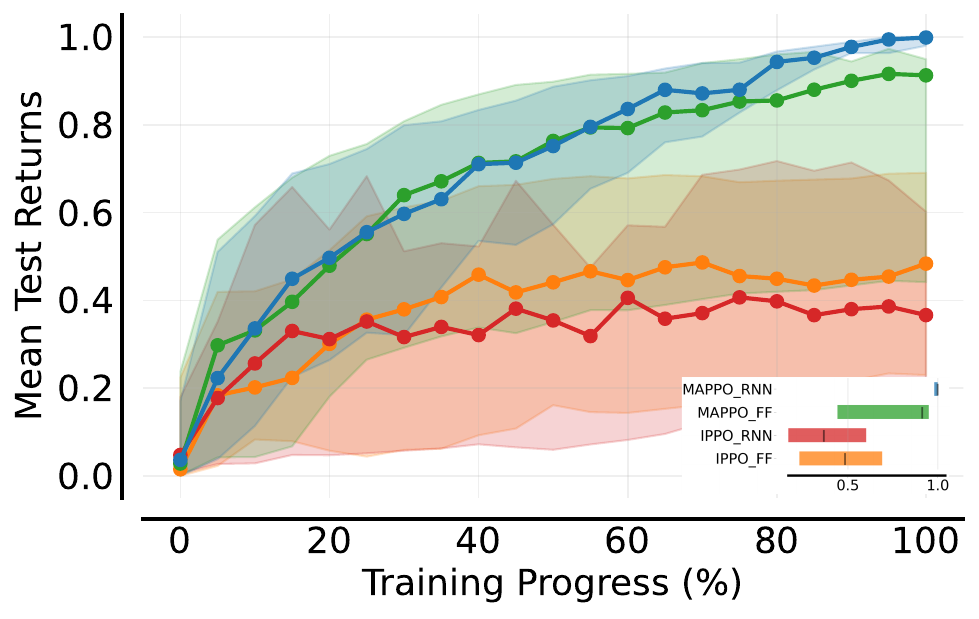}
        \vspace{-0.6em}
        \subcaption{MaBrax}
        \label{fig:mabrax}
    \end{subfigure}
    
    \vspace{0.7em}
    \begin{subfigure}{0.32\textwidth}
        \centering
        \includegraphics[width=\linewidth]{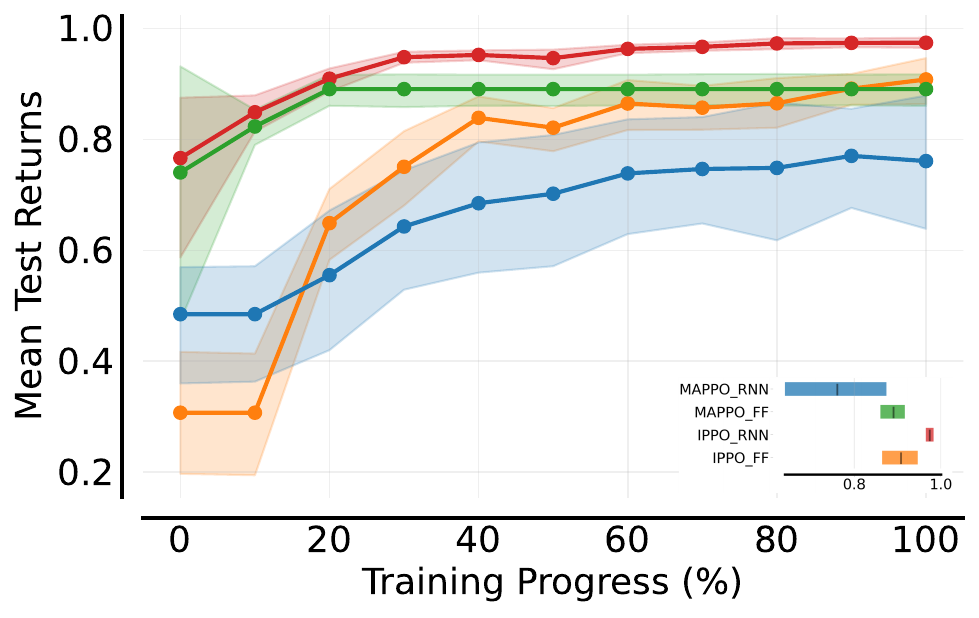}
        \vspace{-0.6em}
        \subcaption{Hanabi}
        \label{fig:hanabi}
    \end{subfigure}
    \hfill
    \begin{subfigure}{0.32\textwidth}
        \centering
        \includegraphics[width=\linewidth]{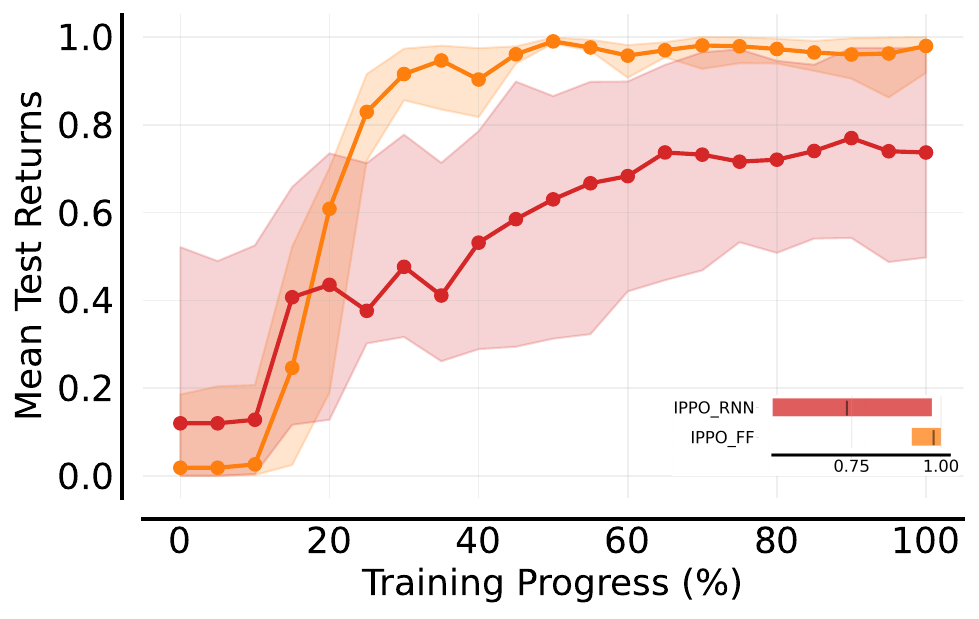}
        \vspace{-0.6em}
        \subcaption{Overcooked V1}
        \label{fig:overcooked_v1}
    \end{subfigure}
    \hfill
    \begin{subfigure}{0.32\textwidth}
        \centering
        \includegraphics[width=\linewidth]{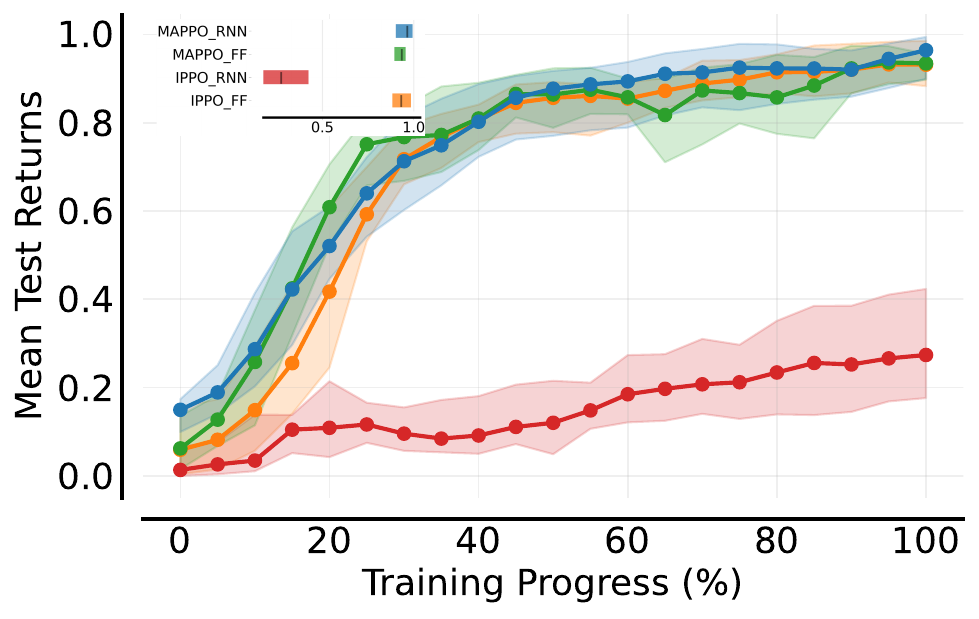}
        \vspace{-0.6em}
        \subcaption{Overcooked V2}
        \label{fig:overcooked_v2}
    \end{subfigure}
    
    \vspace{0.5em}
    \caption{Sample efficiency of IPPO and MAPPO across diverse MARL benchmarks. We show the min-max normalised interquartile mean (IQM) with 95\% stratified bootstrap confidence intervals (CIs). Detailed plots in App.~\ref{app:detailed_results}.}
    \label{fig:all_sample_efficiency_results}
\end{figure*}\footnotetext{MAPPO is omitted for Overcooked V1 as full observability renders a centralised critic redundant.}

\newcommand{\rot}[1]{\rotatebox{0}{#1}}

We apply our diagnostics (Sec.~\ref{sec:diag_decpomdps}) to widely used cooperative MARL benchmarks, using learned policies as probes of \emph{partial observability} and \emph{decentralised coordination} as they arise in behaviour. Concretely, we ask a fundamental question: do these tasks genuinely elicit \emph{Dec-POMDP reasoning}, where agents exploit history to infer decision-relevant hidden states and coordinate based on private information, or do they permit solutions that largely bypass these demands?

\textbf{Experimental Setup.}~~We evaluate 37 scenarios across MPE~\citep{lowe2017multi}, SMAX (V1 maps and V2-style maps)~\citep{rutherford2023jaxmarl,samvelyan2019starcraft}, Overcooked (V1 and V2)~\citep{carroll2019utility,gessler2025overcookedv}, Hanabi~\citep{bard2020hanabi} and MaBrax~\citep{rutherford2023jaxmarl,peng2021facmac}.

\textbf{Evaluation Protocol.}~~We train with 10 seeds, matching original training budgets, and evaluate every 5\% of training (mean evaluation return over 32 episodes)~\citep{gorsane2022towards}. For aggregate comparisons, we report min–max normalised interquartile mean (IQM) with 95\% stratified bootstrap CIs~\citep{agarwal2021deep}. Hyperparameters are tuned per scenario, full details in App.~\ref{app:hyperparms}.

\textbf{Algorithms.}~~We use Independent PPO~\citep[IPPO,][]{de2020independent} and Multi-Agent PPO~\citep[MAPPO,][]{yu2022surprising} as they are widely used MARL baselines. We treat them as two training paradigms: IPPO uses independent critics, whereas MAPPO uses a centralised critic. Additionally, we compare feed-forward (FF) and recurrent (RNN) policies to study the role of memory and temporal information flow in these settings. Finally, to avoid confounders from optimisation and representation choices associated with shared weights in heterogeneous tasks~\citep{christianos2021scaling,tessera2025hypermarl}, we do not use parameter sharing in any baseline.

\vspace{-0.8em}
\subsection{Diagnostic Probes}\label{sec:decision_rules}

To answer the questions from Section~\ref{sec:diag_decpomdps}, we use a two-stage protocol. First, we compute diagnostics on \emph{converged} policies. Then, we determine whether each value reflects genuine structure or finite-sample noise by comparing against a \emph{permutation null baseline}.

\textbf{Permutation null baselines.}~~Information-theoretic estimators (e.g., kNN/KSG~\citep{kraskov2004estimating,ross2014mutual}) can exhibit bias when working with finite samples, resulting in non-zero values even under independence. We therefore construct an empirical null by independently permuting each agent's action sequence within each episode, which destroys temporal and cross-agent dependencies while preserving each agent's marginal action distributions. We recompute each diagnostic on the permuted data and deem the result meaningful only if its value on the original trajectories \emph{exceeds} the mean of the corresponding permutation baseline. 

\textbf{Aggregation.}~~We apply a two-stage aggregation to probe for the emergence of Dec-POMDP reasoning capabilities. First, within each run, we compute the \emph{maximum} diagnostic value across agents, asking whether \emph{any} agent exhibits the property. Second, we maximise across training configurations (IPPO/MAPPO $\times$ FF/RNN) to determine if \emph{any} algorithm elicits the behaviour. This returns a conservative, per-scenario verdict: a property is flagged as absent only if no agent under any tested paradigm displays it.

\textbf{Decision Rules.}~~We now translate the conceptual questions from Section~\ref{sec:diag_decpomdps} into concrete decision rules, evaluating partial observability and coordination directly through agent behaviour.

\begin{table*}[h!]
\centering
\caption{Diagnostics of learned behaviour across cooperative MARL benchmarks. We report the share of scenarios (count/total) where trained policies satisfy our decision criteria (Sec.~\ref{sec:decision_rules}). Crucially, these reflect dependencies \emph{induced by the policy} rather than strict environment requirements. Per-scenario metrics are detailed in App.~\ref{app:detailed_measures}.}
\label{tab:key-diagnostics}
\small
\begin{tabular}{p{0.30\textwidth} ccccccc}
\toprule
\rule[-0.8em]{0pt}{2.5em}
& \rot{MPE} 
& \rot{SMAX V1} 
& \rot{SMAX V2} 
& \rot{MaBrax} 
& \rot{Hanabi} 
& \rot{Overcooked V1} 
& \rot{Overcooked V2} \\
\midrule
Do agents benefit from memory?& \highlight{1.0}{100\% (3/3)} & \highlight{1.0}{100\% (9/9)}& \highlight{1.0}{100\% (3/3)}  &  \highlight{0.2}{20\% (1/5)}   & 0\% (0/1) & \highlight{0}{0\% (0/5)} & \highlight{0}{0\% (0/11)}  \\
Does teammate info. help predict actions? & \highlight{1.0}{100\% (3/3)} & \highlight{0.67}{67\% (6/9)} & \highlight{0.67}{67\% (2/3)}  &  \highlight{1.0}{100\% (5/5)} & 0\% (0/1) & \highlight{0.2}{20\% (1/5)} & \highlight{0.82}{82\% (9/11)} \\
Does synchronous coordination emerge?& \highlight{1.0}{100\% (3/3)} & \highlight{0.44}{44\% (4/9)} & 0\% (0/3)  & \highlight{0.60}{60\% (3/5)} & 0\% (0/1) & \highlight{1.0}{100\% (5/5)} & \highlight{0.82}{82\% (9/11)} \\
Does temporal coordination emerge?& \highlight{1.0}{100\% (3/3)} & \highlight{0.67}{67\% (6/9)} & \highlight{0.67}{67\% (2/3)} &  \highlight{1.0}{100\% (5/5)}& \highlight{1.0}{100\% (1/1)} & \highlight{0.4}{40\% (2/5)} & \highlight{1.0}{100\% (11/11)} \\
\bottomrule
\end{tabular}
\end{table*}

\begin{decisionrule}[Do agents benefit from memory?]
\label{dr:po}
Following Definition~\ref{def:partial_relevance}, agents \emph{benefit from memory} iff both:
\begin{enumerate}[topsep=2pt, itemsep=1pt, leftmargin=*]
\item \textbf{Significant performance gap:} The memory–reactive gap $\Delta_{\text{Mem}}$ is significant (one-tailed Wilcoxon signed-rank, $p<0.05$), see Diagnostic~\ref{diag:gap}.
\item \textbf{Meaningful history use:} Under the memory-based policy, $\mathrm{HAR}^{\mathrm{norm}}$ exceeds its permutation null baseline, see Diagnostic~\ref{diag:har}.
\end{enumerate}
\end{decisionrule}

\noindent Criterion~(1) establishes a reliable performance advantage from memory, while criterion~(2) confirms that this advantage reflects active use of history rather than other confounding factors, such as optimisation dynamics.

\begin{decisionrule}[Does teammate information help predict actions?]\label{dr:pif}

Teammate information \emph{helps predict the actions of other agents} iff $\mathrm{PIF}^{\mathrm{norm}}$ exceeds its permutation null
baseline, indicating that agent $i$'s trajectory and observation inform agent $j$'s action beyond agent $j$'s own history (Diag.~\ref{diag:pif}).
\end{decisionrule}
\begin{decisionrule}[Does synchronous coordination emerge?]
\label{dr:aa}
Instantaneous, synchronous coordination \emph{emerges} iff $\mathrm{AA}^{\mathrm{norm}}$ exceeds its permutation null baseline, indicating coupling beyond shared observations (Diag.~\ref{diag:aa}).
\end{decisionrule}
\begin{decisionrule}[Does temporal coordination emerge?]
\label{dr:di}
Temporal, directional coordination \emph{emerges} iff $\mathrm{DAI}^{\mathrm{norm}}$ exceeds its permutation null baseline, indicating genuine causal influence from past actions (Diag.~\ref{diag:di}).
\end{decisionrule}

\subsection{The Relevance of Partial Observability}\label{sec:results-po}
\noindent\textbf{How often does memory really matter?}

Applying Decision Rule~\ref{dr:po}, we find that memory-based policies yield a statistically significant performance advantage in \textbf{43.2\%} (16/37) of tested scenarios ($\Delta_{\mathrm{Mem}}>0$; see Tbls.~\ref{tab:key-diagnostics}, ~\ref{tab:combined_wilcoxon_results}, and Fig.~\ref{fig:all_sample_efficiency_results}). However, we observe a clear dissociation between history \emph{dependence} and \emph{utility}. $\mathrm{HAR}^{\mathrm{norm}}$ exceeds its permutation null in all 37 scenarios (App. Tbl.~\ref{tab:all_norm_metrics_long}), confirming that trained policies universally encode \emph{some} history dependence, yet this dependence translates into a measurable performance gain in less than half of the cases. Hanabi illustrates this disconnect. Despite being a canonical partially observable task, the memory--reactive gap is not significant under our baselines ($\Delta_{\mathrm{Mem}}=0.279$, Tbls.~\ref{tab:key-diagnostics}, ~\ref{tab:combined_wilcoxon_results}), as IPPO/MAPPO fail to meaningfully exploit recurrent architectures to improve performance on this task (Fig.~\ref{fig:hanabi_results}).

This suggests that much of the observed history dependence could be redundant, i.e., policies learn to track past information that offers no functional advantage over current observations $O_t^i$. Consequently, to genuinely test Dec-POMDP reasoning, environments should ensure decision-relevant information is \emph{exclusively} available through history, rendering reactive policies insufficient.

 \smallskip
\noindent\textbf{Is partial observability reliant on private information?}

From applying Decision Rule~\ref{dr:pif}, we find that $\mathrm{PIF}^{\mathrm{norm}}$ exceeds its permutation null in \textbf{70.3\%} (26/37) of tested scenarios (Tbl.~\ref{tab:key-diagnostics}). Notably, many of these are \emph{not} the same scenarios flagged by the HAR criterion, confirming that hidden environment state and hidden teammate information are distinct drivers of difficulty that our metrics can successfully disentangle (App. Tbl.~\ref{tab:all_norm_metrics_long}).

This separation is especially visible in Overcooked. Overcooked~V1 is fully observable and triggers PIF in only \textbf{20\%} of layouts, while Overcooked~V2, which introduces hidden teammate information by design~\citep{gessler2025overcookedv}, rises to \textbf{82\%}. This serves as an external validation of our diagnostic, as PIF recovers the design intentions of the environment authors. SMAX~V2 maps, following SMACv2, were similarly motivated by "meaningful partial observability"~\citep{ellis2023smacv2}, however, PIF is detected in 67\% of both V1 and V2 maps. This suggests that, at least under current baselines, several V1 maps already exhibit meaningful cross-agent information flow, and the redesign may not have widened this gap as intended.

\vspace{-0.3cm}
\subsection{Decentralised Coordination}
\label{sec:results-coordination}

\textbf{Synchronous vs. Temporal coordination.}

Decision Rules~\ref{dr:aa} and~\ref{dr:di} probe two distinct coordination mechanisms. Synchronous coordination ($\mathrm{AA}$) captures instantaneous action coupling conditioned on current observations, and \textbf{64.9\%} (24/37) of scenarios exceed the null permutation. While high $\mathrm{AA}$ indicates action-action dependence, this coupling can be brittle, e.g. when it reflects rigid, ungrounded conventions that do not generalise~\citep{hu2020other}. Nonetheless, it remains a signature of coordination.

Directed Action Information ($\mathrm{DAI}$), by contrast, measures temporal influence between agents. Under this measure, \textbf{81.1\%} (30/37) of scenarios exceed the null permutation. Notably, \textbf{10/37} scenarios lack synchronous coupling yet exhibit significant temporal influence (App. Tbl.~\ref{tab:all_norm_metrics_long}), indicating that meaningful sequential coordination can arise without simultaneous conventions.

These two mechanisms dissociate systematically across benchmarks, revealing the underlying coordination structure each environment induces. SMAX V2 maps show the starkest separation---none trigger AA, yet 67\% elicit DAI, suggesting that SMAX V2-style combat micro-management relies on sequential positioning rather than synchronous actions. Overcooked V1 presents a contrasting profile (100\% AA, 40\% DAI), reflecting rigid positional conventions in many scenarios. However, Overcooked V2's introduction of hidden information strengthens temporal dependence (100\% DAI) while retaining synchronous coupling (82\% AA). Finally, MPE stands out as the only suite where every scenario demands both coordination forms (100\% AA and 100\% DAI).

\begin{tcolorbox}
[colback=softbluegray, colframe=softbluegray,  boxrule=0.5pt, arc=4pt, before skip=3pt, width=\linewidth]
\textbf{Summary.}\label{sec:results-summary}
Our audit yields four main takeaways:
\begin{enumerate}[leftmargin=*]
\item \textbf{History dependence $\neq$ history utility.} All policies exhibit detectable history dependence ($\mathrm{HAR}^{\mathrm{norm}} > \mathrm{null}$ in 37/37 scenarios), yet only \textbf{43.2\%} show a significant performance gain from memory (Fig.~\ref{fig:all_sample_efficiency_results}, Tbl.~\ref{tab:combined_wilcoxon_results}).
\item \textbf{Hidden state and private information are separable.} PIF flags \textbf{70.3\%} of scenarios, often different ones from HAR, confirming these are separate drivers of difficulty. The Overcooked V1$\to$V2 contrast (\textbf{20\%} $\to$ \textbf{82\%}) validates PIF as an environment-agnostic audit tool.
\item \textbf{Coordination is structurally diverse.} AA (\textbf{64.9\%}) and DAI (\textbf{81.1\%}) dissociate across benchmarks: SMAX V2 exhibits temporal but not synchronous coordination, Overcooked V1 the reverse, while MPE, SMAX V1, MaBrax, and Overcooked V2 trigger both.
\item \textbf{Few benchmarks jointly test partial observability and coordination.} MPE is the only suite in which every scenario satisfies all diagnostic criteria. Most scenarios do not require meaningful history use for strong performance despite being framed as Dec-POMDP challenges. 
\end{enumerate}

\end{tcolorbox}

Our diagnostics expose the divergence between what a benchmark \emph{intends} to test and what it \emph{actually requires}. By characterising \emph{how} agents coordinate rather than just how well, these tools enable researchers to verify Dec-POMDP demands and deliberately select environments that stress-test specific capabilities. Furthermore, as demonstrated in Section~\ref{sec:case_study}, our metrics capture the behavioural impact of structural environment changes, potentially providing actionable guidance for designing more rigorous cooperative environments.

\section{Implications}

\begin{figure}[t]
    \centering
    \vspace{-6pt}

    \begin{subfigure}[t]{0.48\columnwidth}
        \centering
        \includegraphics[width=\linewidth]{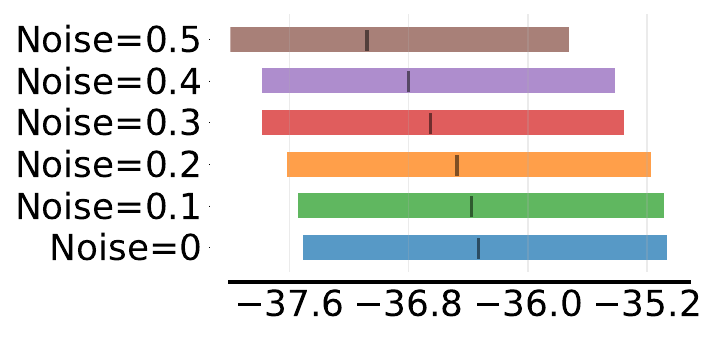}
        \vspace{-0.6em}
        \subcaption{MPE Simple Reference}
        \label{fig:mpe_ref_noise}
    \end{subfigure}
    \hfill
    \begin{subfigure}[t]{0.48\columnwidth}
        \centering
        \includegraphics[width=\linewidth]{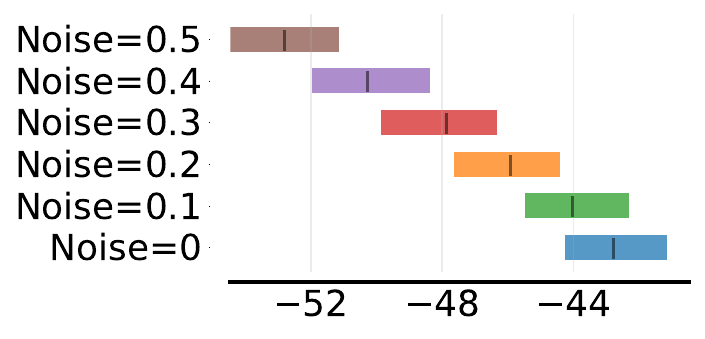}
        \vspace{-0.6em}
        \subcaption{MPE Simple Spread}
        \label{fig:mpe_spread_noise}
    \end{subfigure}

    \caption{Robustness to observational noise in MPE. We report the mean return and 95\% stratified bootstrap confidence intervals for IPPO FeedForward (FF) across varying noise scales. Scenarios with higher $\mathrm{OAR}^{\mathrm{norm}}$ (\emph{Simple Spread}) correlate with greater sensitivity to sensory perturbations.}

    \label{fig:mpe_noise_robustness}
    \vspace{-6pt}
\end{figure}

\vspace{0.5pt}

In many real-world cooperative systems, agents are expected to adapt to changes in their environment and to the behaviour of other agents. While our probes and metrics do not directly measure generalisation, they allow us to detect when policies exhibit weak statistical dependence between observations/histories and actions, i.e., low $\text{OAR}^\text{norm}$ and $\text{HAR}^\text{norm}$. Such instances suggest that agents may be relying on learned conventions or implicit coordination strategies rather than actively conditioning on current observations.

This distinction has nuanced implications. On the one hand, environments in which agents can solve the task via conventions without relying on observations may yield policies that are robust to sensory noise or partial occlusion. On the other hand, such policies may be brittle under structural changes to the environment, to the behaviour of other agents, or even to minimal variations in the task definition~\citep{zamboni2025principledunsupervised}, since coordination may depend on fixed joint strategies rather than observation-driven adaptation.

To examine how our diagnostics relate to behavioural robustness, we conduct controlled evaluations under noisy observations in two MPE tasks with differing $\text{OAR}^\text{norm}$ values: \textit{Simple Spread} and \textit{Simple Reference}. The former exhibits substantially higher estimated mutual information between observations and actions than the latter (IPPO FF, App.~\ref{app:detailed_measures}).

To test robustness to noise, we perturb observations $x$ with additive Gaussian noise scaled by the feature-wise standard deviation $\sigma_x$, computed over $N$ initial rollouts. For noise scale $k \in [0, 0.5]$. For more details on how we add noise see the (App.~\ref{sec:mpe_noise}).

Fig.~\ref{fig:mpe_noise_robustness} shows that performance in \textit{Simple Spread} degrades more substantially under increasing noise than in \textit{Simple Reference}. This is consistent with the higher $\text{OAR}^\text{norm}$ observed in \textit{Simple Spread}, when using IPPO FF.

A key take-away is that information-theoretic diagnostics can provide structured signals about how policies utilise observations and interact with other agents under the training distribution. When interpreted jointly, they can indicate whether behaviour appears observation-driven or convention-driven. However, these metrics quantify statistical dependence rather than causal relationships. As a result, high mutual information does not guarantee sensitivity to noise, and low values do not necessarily imply the absence of structured coordination. Careful behavioural evaluation alongside the use of diagnostics can however provide indications of robustness and generalisation of learned policies.

\section{Challenges and Limitations}\label{sec:challenges_and_limitations}

\noindent\textbf{Policy-dependent probes.} All diagnostics are expectations under the converged joint policy $p^{\boldsymbol\pi}$ and therefore characterise \emph{learned behaviour} under IPPO/MAPPO with FF/RNN architectures, not worst-case or best-case properties of the environment. This is deliberate, as we probe behaviours induced by widely used algorithms; however, stronger or weaker algorithms may yield different diagnostic profiles for the same scenario.

\noindent\textbf{Estimation noise.} Our MI/CMI/DI estimators (kNN and KSG~\citep{kraskov2004estimating}) are biased in finite samples, especially with long histories or large action spaces. We mitigate this via permutation null baselines that account for estimator-specific bias, and report bootstrap confidence intervals throughout. Nonetheless, these probes are diagnostic tools, not hard pass/fail filters, and borderline cases should be interpreted with caution.

\section{Conclusion}

In this work, we introduce a principled diagnostic framework to probe whether cooperative MARL agents genuinely exhibit Dec-POMDP reasoning. By coupling information-theoretic metrics with simple decision rules, our diagnostics evaluate \emph{how} policies solve tasks, not just \emph{how well}, moving evaluation beyond raw returns.

Applied to 37 scenarios across seven environments, our analysis reveals that: (\textbf{i})~history dependence is ubiquitous but rarely yields a performance advantage; (\textbf{ii})~hidden state and private teammate information are separable drivers of difficulty; and (\textbf{iii})~synchronous and temporal coordination frequently dissociate across domains. Notably, MPE is the only environment in which every scenario satisfies all diagnostic criteria. Our case study further demonstrates that the form of emergent coordination is shaped primarily by information bottlenecks in the environment design.

These findings motivate a shift toward benchmarks that strictly compel agents to exploit historical context and coordinate under private information---making \textbf{partial observability} and \textbf{decentralised coordination} non-optional for success.

\begin{acks}
This project received funding from the European Union’s Horizon Europe research and innovation programme under grant agreement No. 101120726. This work was also supported by UK Research and Innovation (UKRI) under the UK government’s Horizon Europe funding Guarantee 10085198.
\end{acks}

\bibliographystyle{ACM-Reference-Format} 
\balance
\bibliography{sample}

\clearpage
\onecolumn
\appendix 

\section*{Appendix}

\section{Solution Concepts}\label{apx:solution_concepts}

As in the single-agent case, we can define value functions for any joint policy $\boldsymbol{\pi}$ composed of  per-agent policies. The value function for agent $i$ are defined as
\begin{equation}
V_{i}^{\boldsymbol{\pi}}(s) := \mathbb{E}_{\boldsymbol{\pi}}\left[\sum_{t=0}^{\infty}\gamma^t R^i(s_{t}, \mathbf{a}_{t}) \mid s_t = s\right]
\end{equation}

We define a best-response policy for agent $i$ as $\pi^{\dagger}(\pi^{-i}) : \mathcal O^i \to \Delta_{{\mathcal A}_i}$ such that
\begin{equation}
V_{i}^{\pi^{i,\dagger}(\pi^{-i}) \times \pi^{-i}}(s) = \sup_{\bar \pi^i} V_{i}^{\bar \pi^i \times \pi^{-i}}(s), \quad \forall s \in \mathcal S,
\end{equation}
\noindent and we denote
\begin{equation}
V_i^{\dagger, \pi_{-i}}(s) := V_i^{\pi^{i,\dagger}(\pi_{-i}) \times \pi_{-i}}(s), \quad V_i^{\dagger, \pi_{-i}}(\mu) := \mathbb{E}_{s \sim \mu} \left[ V_i^{\dagger, \pi_{-i}}(s) \right].
\end{equation}

We are now ready to define the following solution concepts for Dec-POMDPs:

\begin{definition}[(Markov) Equilibria]\label{def:solution_concept} 
For $\epsilon > 0$, a (Markov) policy $\boldsymbol{\pi}$ is a (Markov) $\epsilon$-Approximate Coarse Correlated Equilibrium (CCE) if 
\begin{equation}
\max_{i \in [\vert{\mathcal N}\vert]} \left\{ V_i^{\dagger, \pi_{-i}}(\mu) - V_i^{\boldsymbol{\pi}}(\mu) \right\} \leq \epsilon.
\end{equation}
It is a (Markov) Coarse Correlated Equilibrium if $\epsilon = 0$. A product policy $\boldsymbol{\pi}$ satisfying the same condition is called a ($\epsilon$-Approximate) Nash Equilibrium (NE). If the policy is non-Markov, then the equilibrium is said to be non-Markov.
\end{definition}

\clearpage
\subsection{Hyperparameters}\label{app:hyperparms}
\begin{table}[h]
  \centering
  \caption{Default hyperparameters for MPE experiments.}
  \label{tab:default-hparams-mpe}
  \begin{tabular}{lcccc}
  \toprule
  \textbf{Hyperparameter} & \textbf{IPPO-FF} & \textbf{IPPO-RNN} & \textbf{MAPPO-FF} & \textbf{MAPPO-RNN} \\
  \midrule
  Total Timesteps        & \multicolumn{4}{c}{$1 \times 10^7$} \\
  Num. Parallel Envs     & 16     & 16     & 128    & 16     \\
  Num. Steps             & 128    & 128    & 128    & 128    \\
  FC Hidden Dim          & 128    & 128    & 128    & 128    \\
  GRU Hidden Dim         & --     & 128    & --     & 128    \\
  Num. Minibatches       & \multicolumn{4}{c}{4} \\
  $\gamma$               & \multicolumn{4}{c}{0.99} \\
  GAE $\lambda$          & \multicolumn{4}{c}{0.95} \\
  Entropy Coef.          & \multicolumn{4}{c}{0.01} \\
  Value Function Coef.   & \multicolumn{4}{c}{0.5} \\
  Max Grad Norm          & \multicolumn{4}{c}{0.5} \\
  Activation             & tanh   & tanh   & relu   & tanh   \\
  Anneal LR              & \multicolumn{4}{c}{True} \\
  \bottomrule
  \end{tabular}
  \end{table}

  \begin{table}[h]
  \centering
  \caption{Default hyperparameters for Overcooked (V1) experiments.}
  \label{tab:default-hparams-overcooked-V1}
  \begin{tabular}{lcc}
  \toprule
  \textbf{Hyperparameter} & \textbf{IPPO-FF} & \textbf{IPPO-RNN} \\
  \midrule
  Total Timesteps        & \multicolumn{2}{c}{$1 \times 10^7$} \\
  Num. Parallel Envs     & \multicolumn{2}{c}{64} \\
  Num. Steps             & \multicolumn{2}{c}{256} \\
  FC Hidden Dim          & \multicolumn{2}{c}{64} \\
  GRU Hidden Dim         & --     & 64     \\
  Num. Minibatches       & \multicolumn{2}{c}{16} \\
  $\gamma$               & \multicolumn{2}{c}{0.99} \\
  GAE $\lambda$          & \multicolumn{2}{c}{0.95} \\
  Entropy Coef.          & \multicolumn{2}{c}{0.04} \\
  Value Function Coef.   & \multicolumn{2}{c}{0.5} \\
  Max Grad Norm          & \multicolumn{2}{c}{0.5} \\
  Activation             & \multicolumn{2}{c}{relu} \\
  LR Warmup              & \multicolumn{2}{c}{0.05} \\
  Rew.\ Shaping Horizon  & \multicolumn{2}{c}{$5 \times 10^6$} \\
  Anneal LR              & \multicolumn{2}{c}{True} \\
  \bottomrule
  \end{tabular}
  \end{table}

  \begin{table}[h]
  \centering
  \caption{Default hyperparameters for Overcooked (V2) experiments. All four algorithms share the same defaults.}
  \label{tab:default-hparams-overcooked-V2}
  \begin{tabular}{lc}
  \toprule
  \textbf{Hyperparameter} & \textbf{All Algorithms} \\
  \midrule
  Total Timesteps        & $1 \times 10^7$ \\
  Num. Parallel Envs     & 128 \\
  Num. Steps             & 256 \\
  FC Hidden Dim          & 128 \\
  GRU Hidden Dim         & 128 \\
  Num. Minibatches       & 64 \\
  $\gamma$               & 0.99 \\
  GAE $\lambda$          & 0.95 \\
  Entropy Coef.          & 0.01 \\
  Value Function Coef.   & 0.5 \\
  Max Grad Norm          & 0.25 \\
  Activation             & relu \\
  LR Warmup              & 0.05 \\
  Rew.\ Shaping Horizon  & $5 \times 10^6$ \\
  Anneal LR              & True \\
  \bottomrule
  \end{tabular}
  \end{table}

  \begin{table}[h]
  \centering
  \caption{Default hyperparameters for SMAX experiments.}
  \label{tab:default-hparams-smax}
  \begin{tabular}{lcccc}
  \toprule
  \textbf{Hyperparameter} & \textbf{IPPO-FF} & \textbf{IPPO-RNN} & \textbf{MAPPO-FF} & \textbf{MAPPO-RNN} \\
  \midrule
  Total Timesteps        & \multicolumn{4}{c}{$1 \times 10^7$} \\
  Num. Parallel Envs     & \multicolumn{4}{c}{128} \\
  Num. Steps             & \multicolumn{4}{c}{128} \\
  FC Hidden Dim          & \multicolumn{4}{c}{128} \\
  GRU Hidden Dim         & --     & 128    & --     & 128    \\
  Num. Minibatches       & \multicolumn{4}{c}{4} \\
  $\gamma$               & \multicolumn{4}{c}{0.99} \\
  GAE $\lambda$          & \multicolumn{4}{c}{0.95} \\
  Entropy Coef.          & \multicolumn{4}{c}{0.0} \\
  Value Function Coef.   & \multicolumn{4}{c}{0.5} \\
  Max Grad Norm          & \multicolumn{4}{c}{0.5} \\
  Activation             & \multicolumn{4}{c}{relu} \\
  Anneal LR              & \multicolumn{4}{c}{True} \\
  \bottomrule
  \end{tabular}
  \end{table}

  \begin{table}[h]
\centering
\caption{Default hyperparameters for Hanabi experiments.}
\label{tab:default-hparams-hanabi}
\begin{tabular}{lcccc}
\toprule
\textbf{Hyperparameter} & \textbf{IPPO-FF} & \textbf{IPPO-RNN} & \textbf{MAPPO-FF} & \textbf{MAPPO-RNN} \\
\midrule
Total Timesteps         & \multicolumn{4}{c}{$1 \times 10^{10}$} \\
Num. Parallel Envs      & \multicolumn{4}{c}{1024} \\
Num. Steps              & \multicolumn{4}{c}{128} \\
FC Hidden Dim           & \multicolumn{4}{c}{128} \\
GRU Hidden Dim          & -- & 128 & -- & 128 \\
Num. Minibatches        & \multicolumn{4}{c}{4} \\
Update Epochs           & \multicolumn{4}{c}{4} \\
LR                      & \multicolumn{4}{c}{$5 \times 10^{-4}$} \\
$\gamma$                & \multicolumn{4}{c}{0.99} \\
GAE $\lambda$           & \multicolumn{4}{c}{0.95} \\
Clip $\varepsilon$      & \multicolumn{4}{c}{0.2} \\
Entropy Coef.           & \multicolumn{4}{c}{0.01} \\
Value Function Coef.    & \multicolumn{2}{c}{1.0} & \multicolumn{2}{c}{0.5} \\
Max Grad Norm           & \multicolumn{4}{c}{0.5} \\
Activation              & tanh & \multicolumn{3}{c}{relu} \\
Anneal LR               & \multicolumn{4}{c}{True} \\
Adam $\varepsilon$      & \multicolumn{4}{c}{$10^{-8}$} \\
\bottomrule
\end{tabular}
\end{table}

\begin{table}[h]
\centering
\caption{Default hyperparameters for MABrax experiments.}
\label{tab:default-hparams-mabrax}
\begin{tabular}{lcccc}
\toprule
\textbf{Hyperparameter} & \textbf{IPPO-FF} & \textbf{IPPO-RNN} & \textbf{MAPPO-FF} & \textbf{MAPPO-RNN} \\
\midrule
Total Timesteps         & \multicolumn{4}{c}{$1 \times 10^{8}$} \\
Num. Parallel Envs      & \multicolumn{4}{c}{64} \\
Num. Steps              & \multicolumn{4}{c}{300} \\
FC Hidden Dim           & \multicolumn{4}{c}{128} \\
GRU Hidden Dim          & -- & 128 & -- & 128 \\
Num. Minibatches        & \multicolumn{4}{c}{4} \\
Update Epochs           & \multicolumn{4}{c}{4} \\
LR                      & \multicolumn{4}{c}{$1 \times 10^{-3}$} \\
$\gamma$                & \multicolumn{4}{c}{0.99} \\
GAE $\lambda$           & \multicolumn{4}{c}{0.95} \\
Clip $\varepsilon$      & \multicolumn{4}{c}{0.2} \\
Entropy Coef.           & \multicolumn{4}{c}{$2 \times 10^{-6}$} \\
Value Function Coef.    & \multicolumn{4}{c}{4.5} \\
Max Grad Norm           & \multicolumn{4}{c}{0.5} \\
Activation              & \multicolumn{4}{c}{tanh} \\
Anneal LR               & \multicolumn{4}{c}{True} \\
\bottomrule
\end{tabular}
\end{table}

  \begin{table}[h]
  \centering
  \caption{Tuned hyperparameters for MPE environments, selected from sweep ranges shown in the sub-header.}
  \label{tab:tuned-hparams-mpe}
  \begin{tabular}{llccc}
  \toprule
  \textbf{Environment} & \textbf{Algorithm} & \textbf{LR} & \textbf{Clip $\epsilon$} & \textbf{Update Epochs} \\
   & & $\{10^{-4}, 3{\times}10^{-4}, 10^{-3}\}$ & $\{0.1, 0.2\}$ & $\{2, 4\}$ \\
  \midrule
  \multirow{4}{*}{Spread}
    & IPPO-FF   & $1 \times 10^{-3}$ & 0.1 & 4 \\
    & IPPO-RNN  & $1 \times 10^{-3}$ & 0.1 & 4 \\
    & MAPPO-FF  & $1 \times 10^{-3}$ & 0.1 & 4 \\
    & MAPPO-RNN & $1 \times 10^{-3}$ & 0.1 & 2 \\
  \midrule
  \multirow{4}{*}{Reference}
    & IPPO-FF   & $1 \times 10^{-4}$ & 0.2 & 4 \\
    & IPPO-RNN  & $1 \times 10^{-3}$ & 0.2 & 4 \\
    & MAPPO-FF  & $1 \times 10^{-3}$ & 0.1 & 4 \\
    & MAPPO-RNN & $1 \times 10^{-3}$ & 0.2 & 4 \\
  \midrule
  \multirow{4}{*}{Speaker-Listener}
    & IPPO-FF   & $1 \times 10^{-3}$ & 0.2 & 2 \\
    & IPPO-RNN  & $1 \times 10^{-3}$ & 0.2 & 2 \\
    & MAPPO-FF  & $1 \times 10^{-3}$ & 0.2 & 4 \\
    & MAPPO-RNN & $1 \times 10^{-3}$ & 0.2 & 4 \\
  \midrule
  \multirow{4}{*}{Tag}
    & IPPO-FF   & $1 \times 10^{-3}$ & 0.1 & 2 \\
    & IPPO-RNN  & $1 \times 10^{-4}$ & 0.2 & 2 \\
    & MAPPO-FF  & $1 \times 10^{-3}$ & 0.1 & 2 \\
    & MAPPO-RNN & $1 \times 10^{-4}$ & 0.2 & 4 \\
  \bottomrule
  \end{tabular}
  \end{table}

  \begin{table}[h]
  \centering
  \caption{Tuned hyperparameters for Overcooked (V1) layouts, selected from sweep ranges shown in the sub-header.}
  \label{tab:tuned-hparams-overcooked-V1}
  \begin{tabular}{llcccc}
  \toprule
  \textbf{Layout} & \textbf{Algorithm} & \textbf{LR} & \textbf{Clip $\epsilon$} & \textbf{Update Epochs} & \textbf{Rew.\ Shaping} \\
   & & $\{10^{-4}, 3{\times}10^{-4}, 10^{-3}\}$ & $\{0.1, 0.2\}$ & $\{2, 4\}$ & $\{2.5{\times}10^{6}, 1.5{\times}10^{7}\}$ \\
  \midrule
  \multirow{2}{*}{Cramped Room}
    & IPPO-FF   & $5 \times 10^{-4}$ & 0.2 & 4 & $5 \times 10^{6}$ \\
    & IPPO-RNN  & $1 \times 10^{-3}$ & 0.2 & 4 & $5 \times 10^{6}$ \\
  \midrule
  \multirow{2}{*}{Asymm.\ Advantages}
    & IPPO-FF   & $5 \times 10^{-4}$ & 0.2 & 4 & $5 \times 10^{6}$ \\
    & IPPO-RNN  & $1 \times 10^{-3}$ & 0.2 & 4 & $5 \times 10^{6}$ \\
  \midrule
  \multirow{2}{*}{Coord.\ Ring}
    & IPPO-FF   & $1 \times 10^{-3}$ & 0.2 & 4 & $5 \times 10^{6}$ \\
    & IPPO-RNN  & $5 \times 10^{-4}$ & 0.2 & 4 & $5 \times 10^{6}$ \\
  \midrule
  \multirow{2}{*}{Counter Circuit}
    & IPPO-FF   & $5 \times 10^{-4}$ & 0.2 & 4 & $5 \times 10^{6}$ \\
    & IPPO-RNN  & $1 \times 10^{-3}$ & 0.2 & 4 & $5 \times 10^{6}$ \\
  \midrule
  \multirow{2}{*}{Forced Coord.}
    & IPPO-FF   & $5 \times 10^{-4}$ & 0.2 & 4 & $5 \times 10^{6}$ \\
    & IPPO-RNN  & $1 \times 10^{-3}$ & 0.2 & 4 & $5 \times 10^{6}$ \\
  \bottomrule
  \end{tabular}
  \end{table}

  \begin{table}[h]
  \centering
  \caption{Tuned hyperparameters for Overcooked (V2) existing layouts. Only LR was swept; Clip $\epsilon = 0.2$ and Update Epochs $= 4$ were fixed.}
  \label{tab:tuned-hparams-overcooked-V2-existing}
  \begin{tabular}{llc}
  \toprule
  \textbf{Layout} & \textbf{Algorithm} & \textbf{LR} \\
   & & $\{10^{-4}, 4{\times}10^{-4}, 5{\times}10^{-4}, 10^{-3}\}$ \\
  \midrule
  \multirow{4}{*}{Cramped Room}
    & IPPO-FF   & $1 \times 10^{-3}$ \\
    & IPPO-RNN  & $5 \times 10^{-4}$ \\
    & MAPPO-FF  & $5 \times 10^{-4}$ \\
    & MAPPO-RNN & $5 \times 10^{-4}$ \\
  \midrule
  \multirow{4}{*}{Asymm.\ Advantages}
    & IPPO-FF   & $5 \times 10^{-4}$ \\
    & IPPO-RNN  & $1 \times 10^{-3}$ \\
    & MAPPO-FF  & $5 \times 10^{-4}$ \\
    & MAPPO-RNN & $1 \times 10^{-3}$ \\
  \midrule
  \multirow{4}{*}{Coord.\ Ring}
    & IPPO-FF   & $1 \times 10^{-3}$ \\
    & IPPO-RNN  & $1 \times 10^{-3}$ \\
    & MAPPO-FF  & $5 \times 10^{-4}$ \\
    & MAPPO-RNN & $4 \times 10^{-4}$ \\
  \midrule
  \multirow{4}{*}{Counter Circuit}
    & IPPO-FF   & $1 \times 10^{-3}$ \\
    & IPPO-RNN  & $1 \times 10^{-3}$ \\
    & MAPPO-FF  & $4 \times 10^{-4}$ \\
    & MAPPO-RNN & $5 \times 10^{-4}$ \\
  \midrule
  \multirow{4}{*}{Forced Coord.}
    & IPPO-FF   & $1 \times 10^{-3}$ \\
    & IPPO-RNN  & $4 \times 10^{-4}$ \\
    & MAPPO-FF  & $5 \times 10^{-4}$ \\
    & MAPPO-RNN & $1 \times 10^{-4}$ \\
  \bottomrule
  \end{tabular}
  \end{table}

  \begin{table}[h]
  \centering
  \caption{Tuned hyperparameters for Overcooked (V2) new layouts. Only LR was swept; Clip $\epsilon = 0.2$ and Update Epochs $= 4$ were fixed. Total timesteps $= 3 \times 10^7$ and reward
  shaping horizon $= 1.5 \times 10^7$.}
  \label{tab:tuned-hparams-overcooked-V2-new}
  \begin{tabular}{llc}
  \toprule
  \textbf{Layout} & \textbf{Algorithm} & \textbf{LR} \\
   & & $\{10^{-4}, 4{\times}10^{-4}, 5{\times}10^{-4}, 10^{-3}\}$ \\
  \midrule
  \multirow{4}{*}{Demo Cook Simple}
    & IPPO-FF   & $1 \times 10^{-3}$ \\
    & IPPO-RNN  & $1 \times 10^{-3}$ \\
    & MAPPO-FF  & $5 \times 10^{-4}$ \\
    & MAPPO-RNN & $5 \times 10^{-4}$ \\
  \midrule
  \multirow{4}{*}{Demo Cook Wide}
    & IPPO-FF   & $1 \times 10^{-3}$ \\
    & IPPO-RNN  & $3 \times 10^{-4}$ \\
    & MAPPO-FF  & $1 \times 10^{-3}$ \\
    & MAPPO-RNN & $4 \times 10^{-4}$ \\
  \midrule
  \multirow{4}{*}{Grounded Coord.\ Ring}
    & IPPO-FF   & $1 \times 10^{-3}$ \\
    & IPPO-RNN  & $1 \times 10^{-4}$ \\
    & MAPPO-FF  & $1 \times 10^{-3}$ \\
    & MAPPO-RNN & $4 \times 10^{-4}$ \\
  \midrule
  \multirow{4}{*}{Grounded Coord.\ Simple}
    & IPPO-FF   & $1 \times 10^{-3}$ \\
    & IPPO-RNN  & $1 \times 10^{-3}$ \\
    & MAPPO-FF  & $1 \times 10^{-3}$ \\
    & MAPPO-RNN & $4 \times 10^{-4}$ \\
  \midrule
  \multirow{4}{*}{Test Time Simple}
    & IPPO-FF   & $1 \times 10^{-3}$ \\
    & IPPO-RNN  & $1 \times 10^{-3}$ \\
    & MAPPO-FF  & $1 \times 10^{-3}$ \\
    & MAPPO-RNN & $1 \times 10^{-3}$ \\
  \midrule
  \multirow{4}{*}{Test Time Wide}
    & IPPO-FF   & $1 \times 10^{-3}$ \\
    & IPPO-RNN  & $5 \times 10^{-4}$ \\
    & MAPPO-FF  & $5 \times 10^{-4}$ \\
    & MAPPO-RNN & $3 \times 10^{-4}$ \\
  \bottomrule
  \end{tabular}
  \end{table}

  \begin{table}[h]
  \centering
  \caption{Tuned hyperparameters for SMAX V1 maps, selected from sweep ranges shown in the sub-header. Clip $\epsilon = 0.1$ was selected for all configurations. UE = Update Epochs.}
  \label{tab:tuned-hparams-SMAX V1}
  \begin{tabular}{lcccccccc}
  \toprule
   & \multicolumn{2}{c}{\textbf{IPPO-FF}} & \multicolumn{2}{c}{\textbf{IPPO-RNN}} & \multicolumn{2}{c}{\textbf{MAPPO-FF}} & \multicolumn{2}{c}{\textbf{MAPPO-RNN}} \\
  \cmidrule(lr){2-3} \cmidrule(lr){4-5} \cmidrule(lr){6-7} \cmidrule(lr){8-9}
  \textbf{Map} & \textbf{LR} & \textbf{UE} & \textbf{LR} & \textbf{UE} & \textbf{LR} & \textbf{UE} & \textbf{LR} & \textbf{UE} \\
  \multicolumn{1}{l}{\footnotesize Sweep range:} & \multicolumn{8}{c}{\footnotesize LR $\in \{10^{-4}, 3{\times}10^{-4}, 10^{-3}\}$, \; Clip $\epsilon \in \{0.1, 0.2\}$, \; UE $\in \{2,
  4\}$} \\
  \midrule
  3m              & $10^{-3}$ & 4 & $10^{-3}$ & 4 & $10^{-3}$ & 4 & $10^{-3}$ & 4 \\
  2s3z            & $10^{-3}$ & 4 & $10^{-3}$ & 4 & $10^{-3}$ & 4 & $10^{-3}$ & 4 \\
  3s5z            & $10^{-3}$ & 4 & $10^{-3}$ & 2 & $10^{-3}$ & 4 & $10^{-3}$ & 2 \\
  3s\_vs\_5z      & $10^{-3}$ & 4 & $10^{-3}$ & 2 & $10^{-3}$ & 4 & $10^{-3}$ & 4 \\
  3s5z\_vs\_3s6z  & $10^{-3}$ & 4 & $10^{-3}$ & 2 & $10^{-3}$ & 4 & $10^{-3}$ & 4 \\
  5m\_vs\_6m      & $10^{-4}$ & 2 & $10^{-3}$ & 2 & $10^{-3}$ & 4 & $10^{-3}$ & 4 \\
  8m              & $10^{-3}$ & 4 & $10^{-3}$ & 4 & $10^{-3}$ & 4 & $10^{-3}$ & 4 \\
  10m\_vs\_11m    & $10^{-4}$ & 4 & $10^{-3}$ & 2 & $10^{-4}$ & 4 & $10^{-3}$ & 4 \\
  6h\_vs\_8z      & $10^{-3}$ & 4 & $10^{-3}$ & 4 & $10^{-3}$ & 4 & $10^{-3}$ & 2 \\
  \bottomrule
  \end{tabular}
  \end{table}

  \begin{table}[h]
  \centering
  \caption{Tuned hyperparameters for SMAX V2 maps, selected from sweep ranges shown in the sub-header. Clip $\epsilon = 0.1$ and Update Epochs $= 4$ were selected for all configurations.}
  \label{tab:tuned-hparams-SMAX V2}
  \begin{tabular}{llccc}
  \toprule
  \textbf{Map} & \textbf{Algorithm} & \textbf{LR} & \textbf{Clip $\epsilon$} & \textbf{Update Epochs} \\
   & & $\{10^{-4}, 3{\times}10^{-4}, 10^{-3}\}$ & $\{0.1, 0.2\}$ & $\{2, 4\}$ \\
  \midrule
  \multirow{4}{*}{SMAX V2\_5\_units}
    & IPPO-FF   & $1 \times 10^{-3}$ & 0.1 & 4 \\
    & IPPO-RNN  & $1 \times 10^{-3}$ & 0.1 & 4 \\
    & MAPPO-FF  & $1 \times 10^{-3}$ & 0.1 & 4 \\
    & MAPPO-RNN & $1 \times 10^{-3}$ & 0.1 & 4 \\
  \midrule
  \multirow{4}{*}{SMAX V2\_10\_units}
    & IPPO-FF   & $1 \times 10^{-3}$ & 0.1 & 4 \\
    & IPPO-RNN  & $1 \times 10^{-3}$ & 0.1 & 4 \\
    & MAPPO-FF  & $1 \times 10^{-3}$ & 0.1 & 4 \\
    & MAPPO-RNN & $1 \times 10^{-3}$ & 0.1 & 4 \\
  \midrule
  \multirow{4}{*}{SMAX V2\_20\_units}
    & IPPO-FF   & $1 \times 10^{-3}$ & 0.1 & 4 \\
    & IPPO-RNN  & $1 \times 10^{-3}$ & 0.1 & 4 \\
    & MAPPO-FF  & $1 \times 10^{-3}$ & 0.1 & 4 \\
    & MAPPO-RNN & $1 \times 10^{-3}$ & 0.1 & 4 \\
  \bottomrule
  \end{tabular}
  \end{table}

\clearpage
\section{Detailed Results}\label{app:detailed_results}
\subsection{Performance Results}

\begin{table}[h]
\centering
\small
\caption{Per-environment performance deltas, reported as the median of differences. Bold values indicate statistically significant results ($p<0.05$, one-sided Wilcoxon signed rank test). We observe a statistically significant memory advantage ($\Delta > 0$) in \textbf{43.2\%} (16/37) of the tested environments. $^{\dagger}$MAPPO comparisons are omitted for Overcooked V1 as the environment is fully observable (rendering IPPO/MAPPO identical).}
\begin{tabular}{llcc}
\toprule
Environment & Scenario & $\Delta$~(RNN$-$FF) & $\Delta$~(MAPPO$-$IPPO) \\
\midrule
\multirow{3}{*}{MPE} 
 & \texttt{MPE\_simple\_reference\_v3} & \textbf{6.496} & \textbf{2.019} \\
 & \texttt{MPE\_simple\_speaker\_listener\_v4} & \textbf{14.840} & \textbf{1.596} \\
 & \texttt{MPE\_simple\_spread\_v3} & \textbf{2.500} & \textbf{0.959} \\
\cmidrule{1-4}
\multirow{9}{*}{SMAX-V1 Maps} 
 & \texttt{10m\_vs\_11m} & \textbf{0.945} & -0.150 \\
 & \texttt{2s3z} & \textbf{1.241} & \textbf{0.009} \\
 & \texttt{3m} & \textbf{0.225} & \textbf{0.025} \\
 & \texttt{3s5z} & \textbf{1.169} & \textbf{0.280} \\
 & \texttt{3s5z\_vs\_3s6z} & \textbf{0.336} & \textbf{0.282} \\
 & \texttt{3s\_vs\_5z} & \textbf{0.228} & 0.138 \\
 & \texttt{5m\_vs\_6m} & \textbf{0.358} & \textbf{0.254} \\
 & \texttt{6h\_vs\_8z} & \textbf{0.168} & \textbf{0.047} \\
 & \texttt{8m} & \textbf{1.127} & \textbf{0.102} \\
\cmidrule{1-4}
\multirow{3}{*}{SMAX-V2 Maps} 
 & \texttt{SMAX V2\_10\_units} & \textbf{0.397} & \textbf{0.327} \\
 & \texttt{SMAX V2\_20\_units} & \textbf{0.205} & \textbf{0.151} \\
 & \texttt{SMAX V2\_5\_units} & \textbf{0.459} & \textbf{0.372} \\
\cmidrule{1-4}
\multirow{5}{*}{MaBrax} 
 & \texttt{ant\_4x2} & -1028.182 & \textbf{3487.891} \\
 & \texttt{halfcheetah\_6x1} & -184.390 & \textbf{1465.603} \\
 & \texttt{hopper\_3x1} & 159.544 & \textbf{606.643} \\
 & \texttt{humanoid\_9|8} & 31.723 & \textbf{397.284} \\
 & \texttt{walker2d\_2x3} & \textbf{452.472} & 73.575 \\
\cmidrule{1-4}
Hanabi & \texttt{Two Players} & 0.279 & -1.313 \\
\cmidrule{1-4}
\multirow{5}{*}{Overcooked} 
 & asymm\_advantages & -220.000 & --- \\
 & coord\_ring & -40.000 & --- \\
 & counter\_circuit & -35.625 & --- \\
 & cramped\_room & 0.000 & --- \\
 & forced\_coord & 0.000 & --- \\
\cmidrule{1-4}
\multirow{11}{*}{Overcooked V2} 
 & asymm\_advantages & -20.000 & \textbf{46.875} \\
 & coord\_ring & 30.625 & \textbf{31.250} \\
 & counter\_circuit & -8.750 & 34.062 \\
 & cramped\_room & -25.625 & \textbf{8.125} \\
 & demo\_cook\_simple & -60.625 & 7.500 \\
 & demo\_cook\_wide & -10.938 & -7.188 \\
 & forced\_coord & -6.250 & \textbf{34.688} \\
 & grounded\_coord\_ring & -16.875 & \textbf{16.875} \\
 & grounded\_coord\_simple & -2.500 & \textbf{11.250} \\
 & test\_time\_simple & -2.500 & \textbf{10.000} \\
 & test\_time\_wide & 0.938 & 5.312 \\
\bottomrule
\end{tabular}

\label{tab:combined_wilcoxon_results}
\end{table}

\begin{figure*}[t]
  \centering

  \begin{minipage}{\textwidth}
    \centering
    \includegraphics[width=0.6\linewidth]{figures/legend.pdf}
  \end{minipage}

  \vspace{0.75em}

  \begin{subfigure}{0.32\textwidth}
    \centering
    \includegraphics[width=\linewidth]{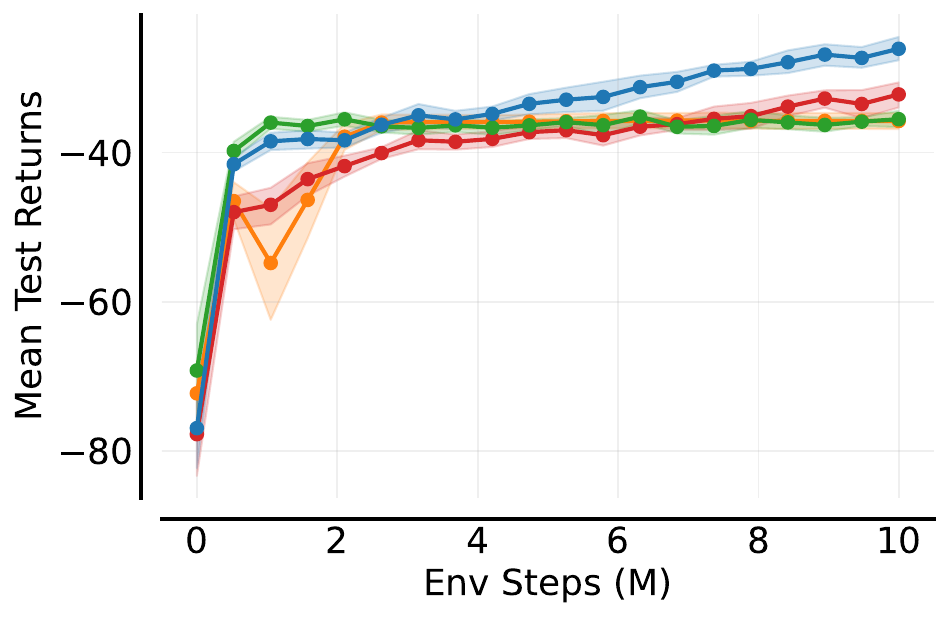}
    \caption{Reference}\label{fig:mpe_ref_perf}
  \end{subfigure}\hfill
  \begin{subfigure}{0.32\textwidth}
    \centering
    \includegraphics[width=\linewidth]{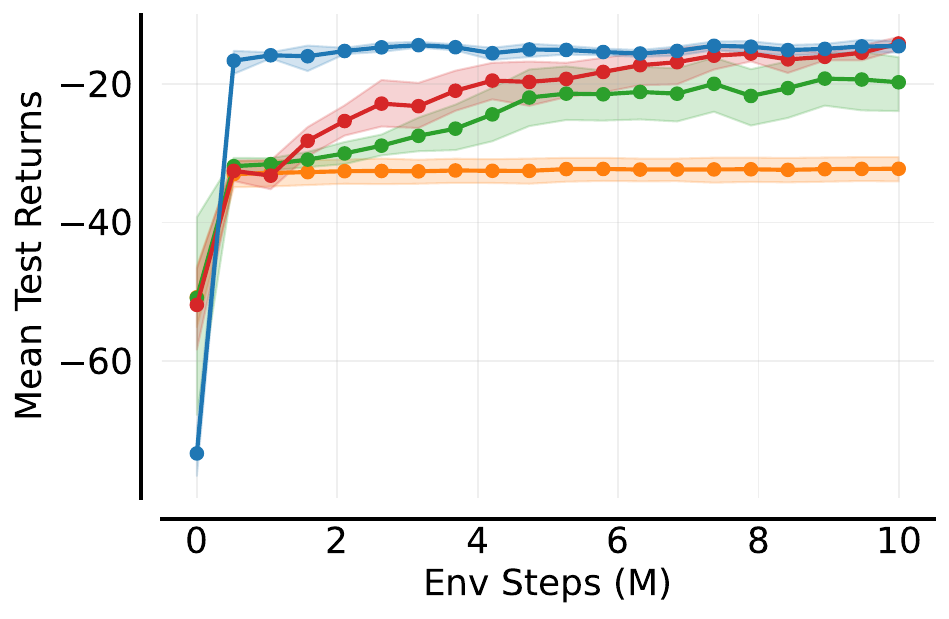}
    \caption{Speaker–Listener}\label{fig:mpe_speaker_perf}
  \end{subfigure}\hfill
  \begin{subfigure}{0.32\textwidth}
    \centering
    \includegraphics[width=\linewidth]{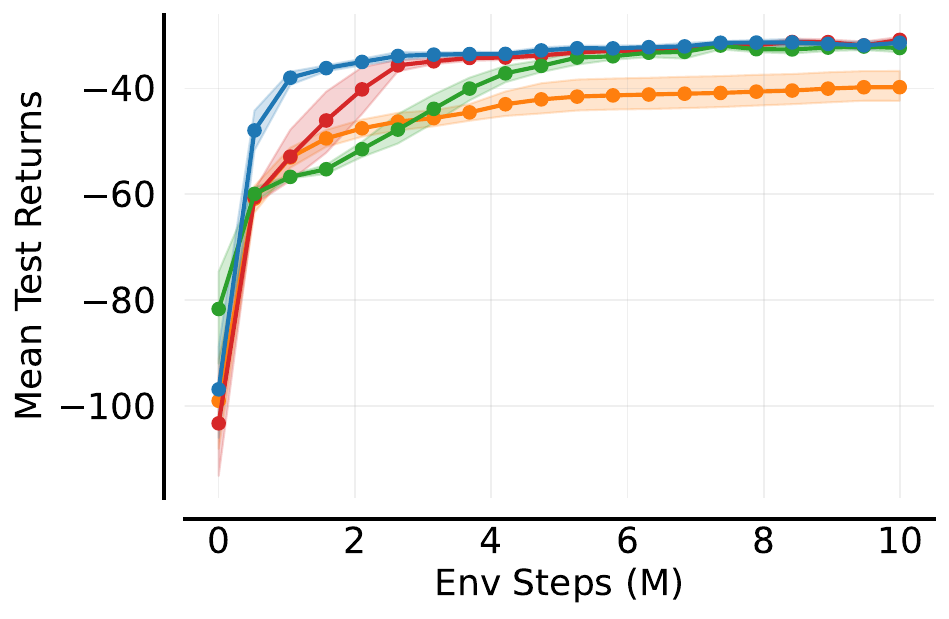}
    \caption{Spread}\label{fig:mpe_spread_perf}
  \end{subfigure}
  \caption{Mean test returns with 95\% confidence intervals in MPE.}
  \label{fig:mpe_three_wide}
\end{figure*}

\begin{figure*}[t]
  \centering

  \begin{minipage}{\textwidth}
    \centering
    \includegraphics[width=0.6\linewidth]{figures/legend.pdf}
  \end{minipage}

  \vspace{0.75em}

  \begin{subfigure}{0.32\textwidth}
    \centering
    \includegraphics[width=\linewidth]{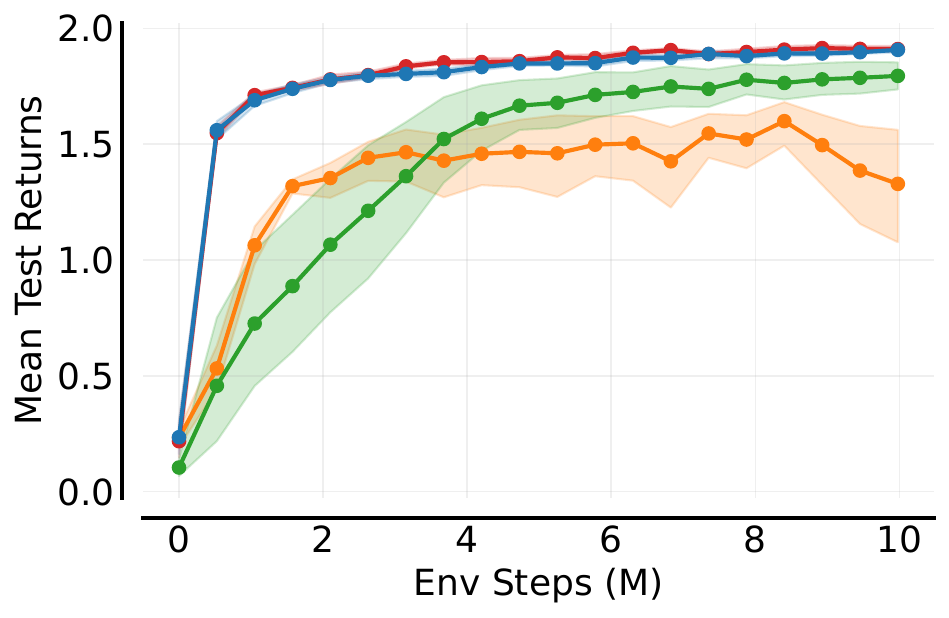}
    \caption{3m}\label{fig:perf_3m}
  \end{subfigure}\hfill
  \begin{subfigure}{0.32\textwidth}
    \centering
    \includegraphics[width=\linewidth]{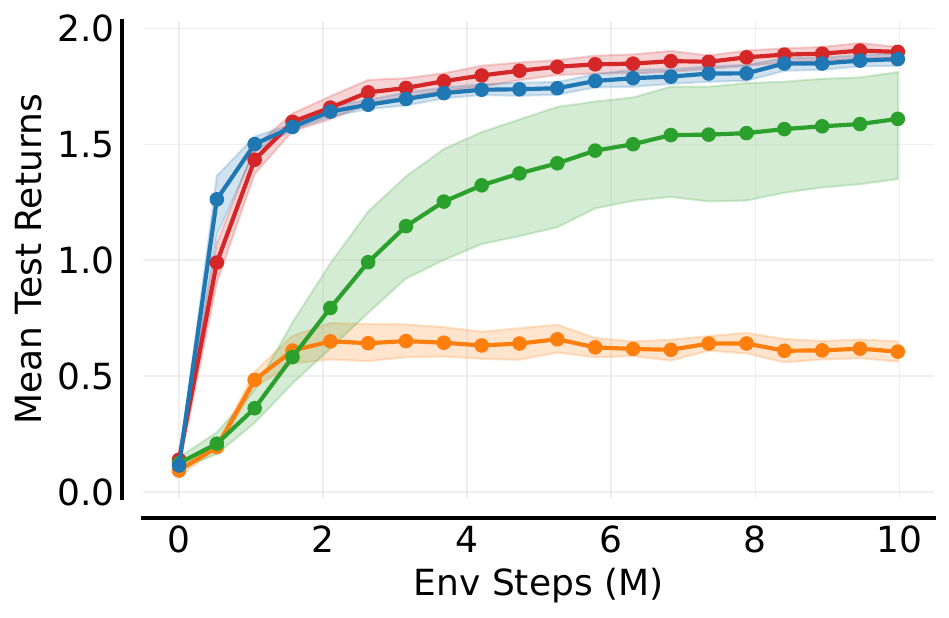}
    \caption{8m}\label{fig:perf_8m}
  \end{subfigure}\hfill
  \begin{subfigure}{0.32\textwidth}
    \centering
    \includegraphics[width=\linewidth]{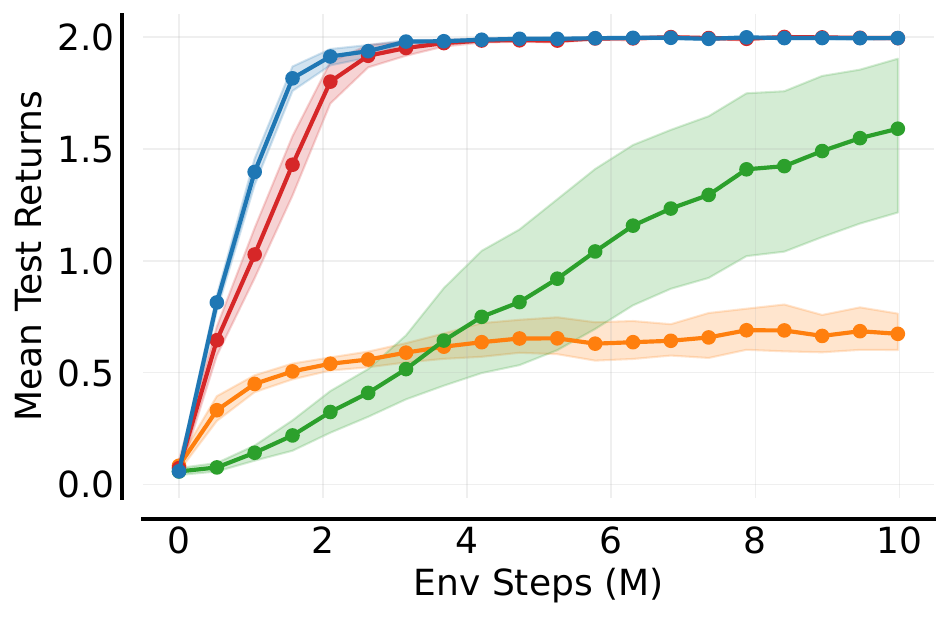}
    \caption{2s3z}\label{fig:perf_2s3z}
  \end{subfigure}

  \begin{subfigure}{0.32\textwidth}
    \centering
    \includegraphics[width=\linewidth]{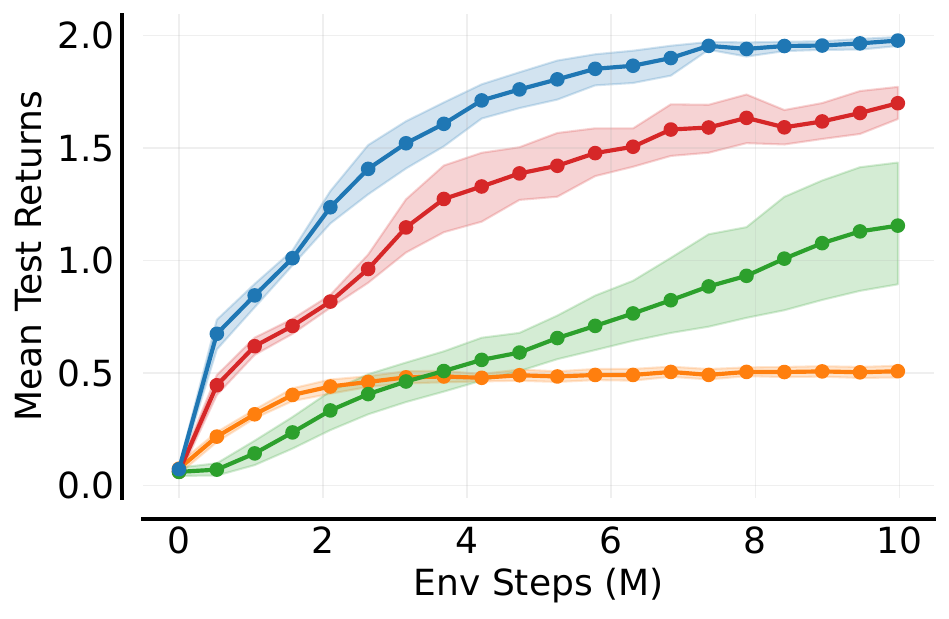}
    \caption{3s5z}\label{fig:perf_3s5z}
  \end{subfigure}\hfill
  \begin{subfigure}{0.32\textwidth}
    \centering
    \includegraphics[width=\linewidth]{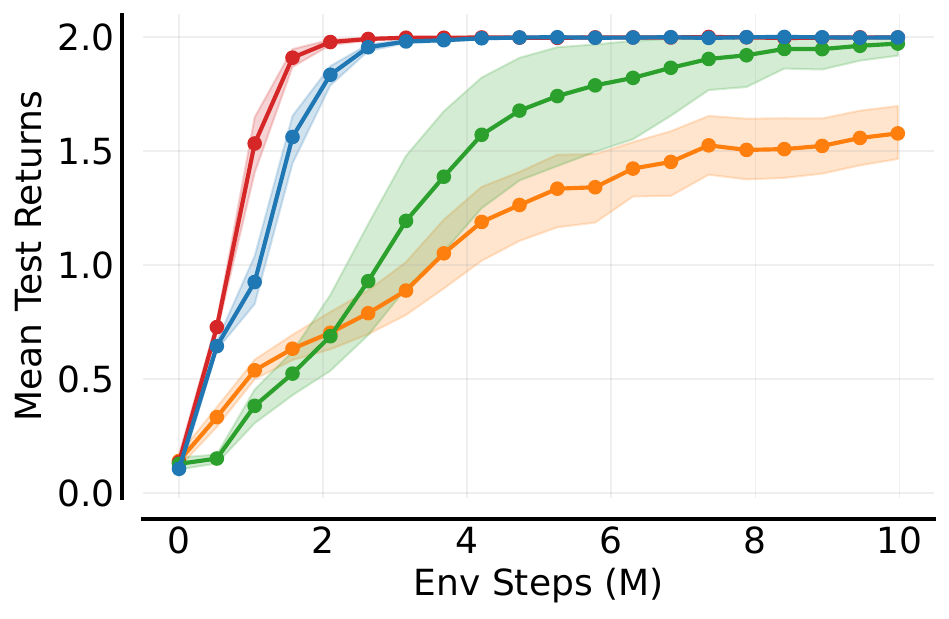}
    \caption{6h\_vs\_8z}\label{fig:perf_6h_vs_8z}
  \end{subfigure}\hfill
  \begin{subfigure}{0.32\textwidth}
    \centering
    \includegraphics[width=\linewidth]{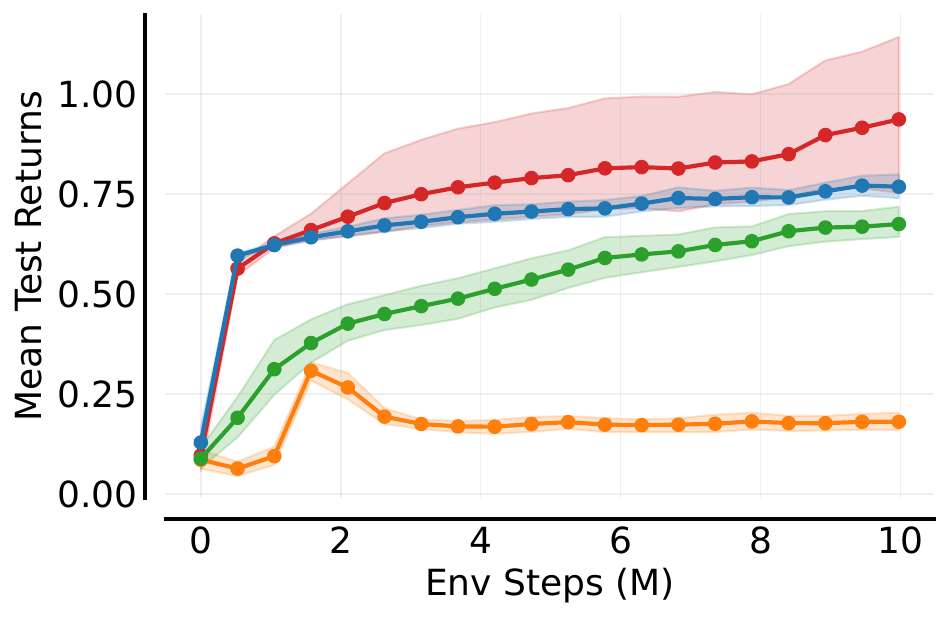}
    \caption{5m\_vs\_6m}\label{fig:perf_5m_vs_6m}
  \end{subfigure}

  \begin{subfigure}{0.32\textwidth}
    \centering
    \includegraphics[width=\linewidth]{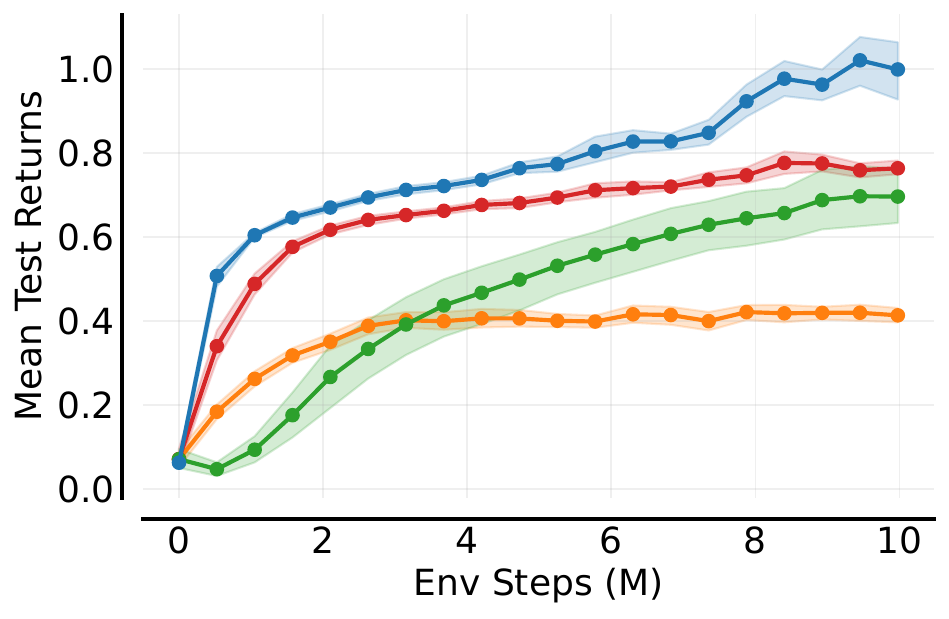}
    \caption{3s5z\_vs\_3s6z}\label{fig:perf_3s5z_vs_3s6z}
  \end{subfigure}\hfill
  \begin{subfigure}{0.32\textwidth}
    \centering
    \includegraphics[width=\linewidth]{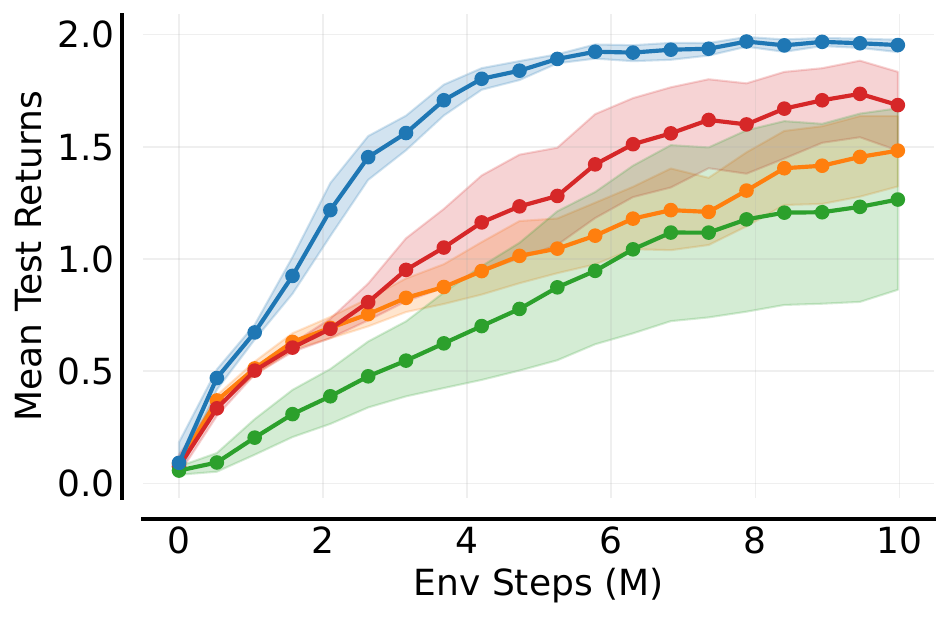}
    \caption{3s\_vs\_5z}\label{fig:perf_3s_vs_5z}
  \end{subfigure}\hfill
  \begin{subfigure}{0.32\textwidth}
    \centering
    \includegraphics[width=\linewidth]{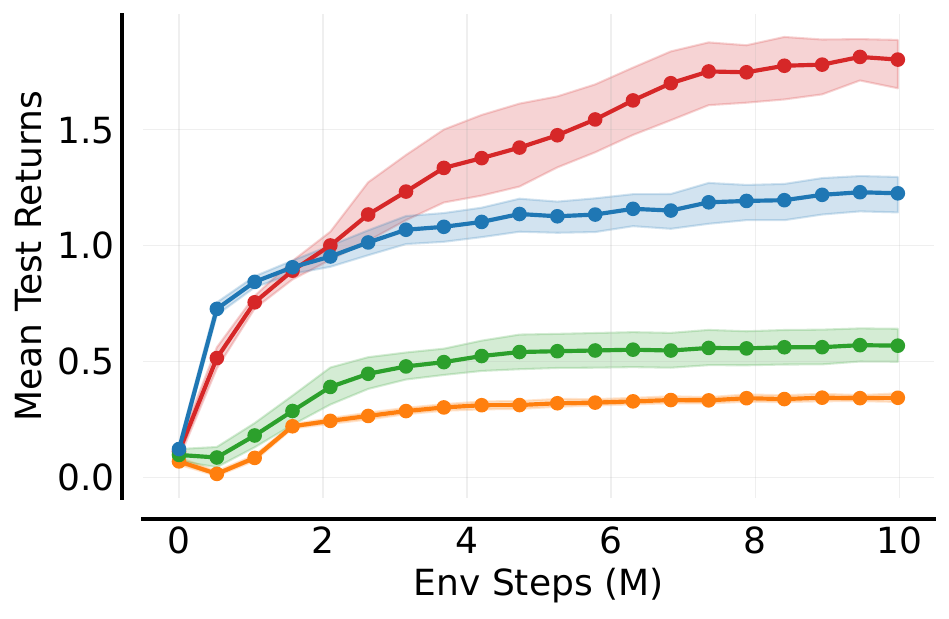}
    \caption{10m\_vs\_11m}\label{fig:perf_10m_vs_11m}
  \end{subfigure}
  \caption{Mean test returns with 95\% confidence intervals in SMAX V1.}
\end{figure*}

\begin{figure*}[t]
  \centering

  \begin{minipage}{\textwidth}
    \centering
    \includegraphics[width=0.6\linewidth]{figures/legend.pdf}
  \end{minipage}

  \vspace{0.75em}

  \begin{subfigure}{0.32\textwidth}
    \centering
    \includegraphics[width=\linewidth]{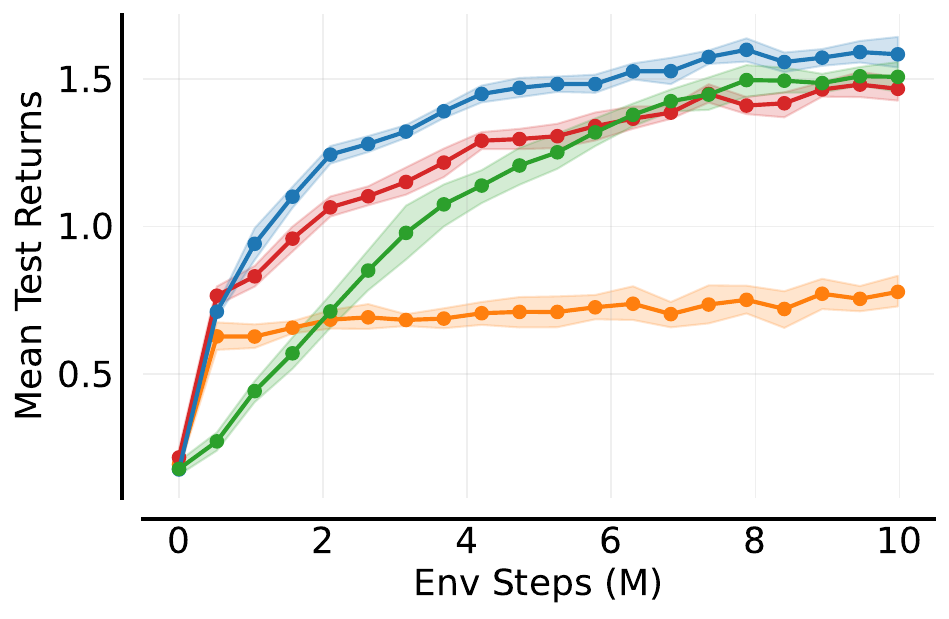}
    \caption{5 units}\label{fig:smaxv2_5_perf}
  \end{subfigure}\hfill
  \begin{subfigure}{0.32\textwidth}
    \centering
    \includegraphics[width=\linewidth]{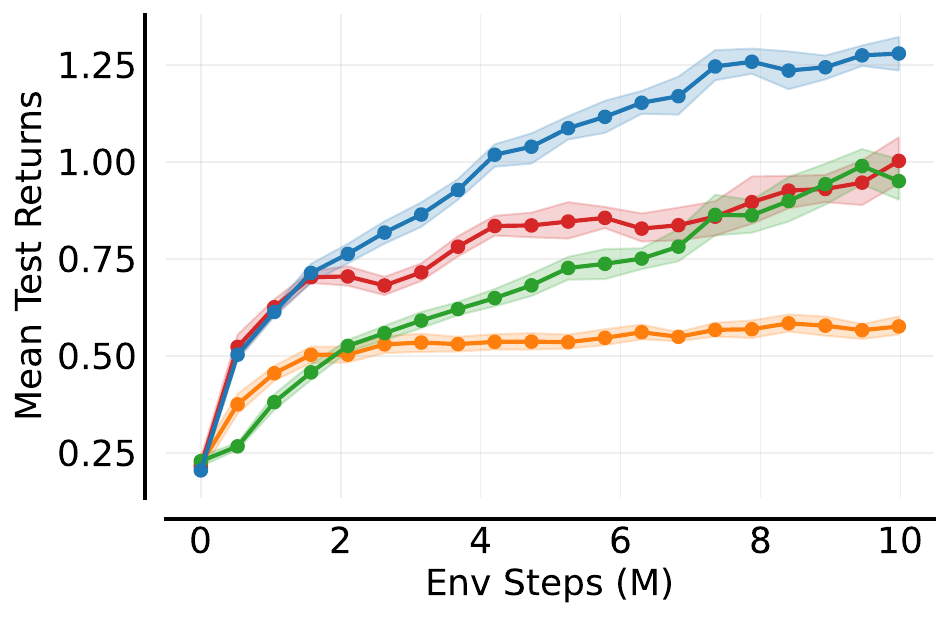}
    \caption{10 units}\label{fig:smaxv2_10_perf}
  \end{subfigure}\hfill
  \begin{subfigure}{0.32\textwidth}
    \centering
    \includegraphics[width=\linewidth]{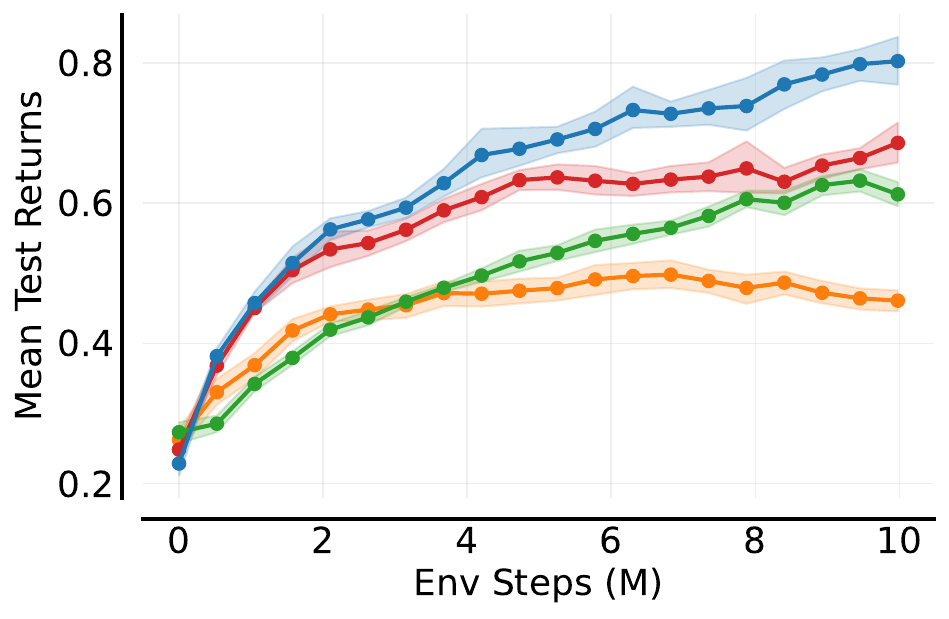}
    \caption{20 units}\label{fig:smaxv2_20_perf}
  \end{subfigure}
  \caption{Mean test returns with 95\% confidence intervals in SMAX V2.}
  \label{fig:smaxv2_three_wide}
\end{figure*}

\begin{figure*}[t]
  \centering

  \begin{minipage}{\textwidth}
    \centering
    \includegraphics[width=0.3\linewidth]{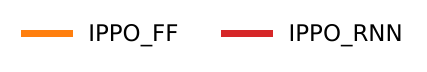}
  \end{minipage}

  \vspace{0.75em}

  \begin{subfigure}{0.32\textwidth}
    \centering
    \includegraphics[width=\linewidth]{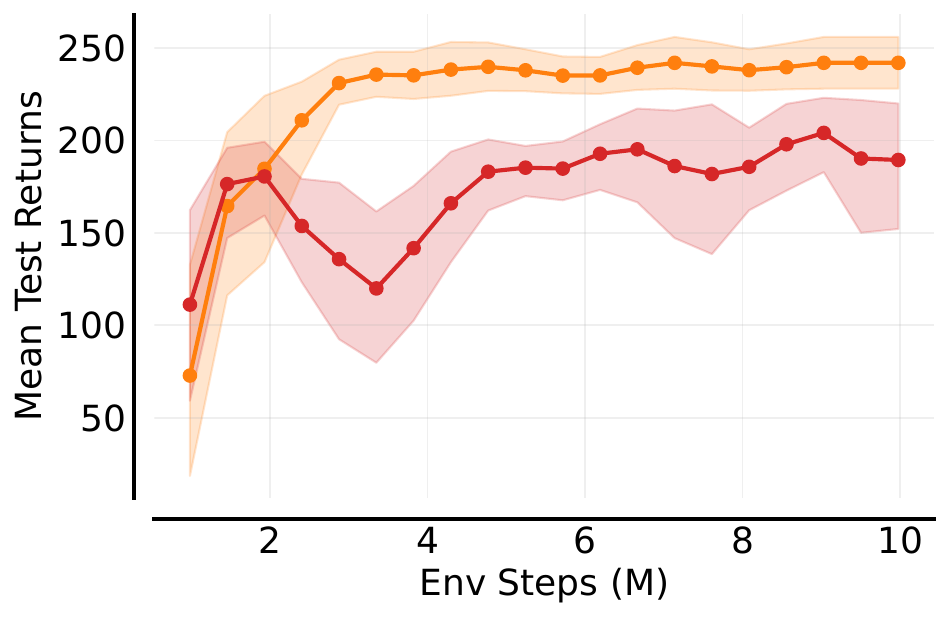}
    \caption{Coordination Ring}\label{fig:coord_ring}
  \end{subfigure}\hfill
  \begin{subfigure}{0.32\textwidth}
    \centering
    \includegraphics[width=\linewidth]{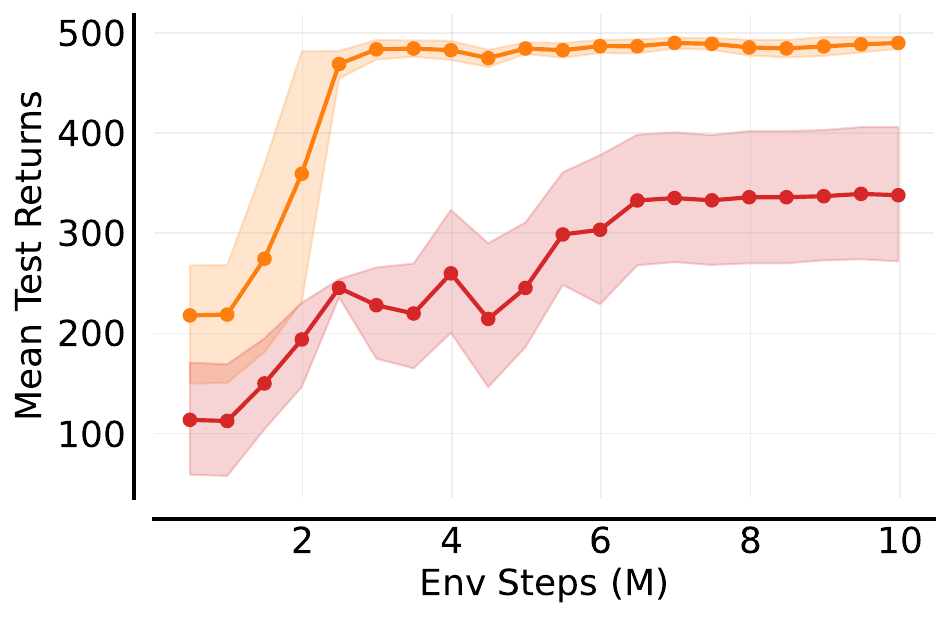}
    \caption{Asymmetric Advantages}\label{fig:asymm_advantages}
  \end{subfigure}\hfill
  \begin{subfigure}{0.32\textwidth}
    \centering
    \includegraphics[width=\linewidth]{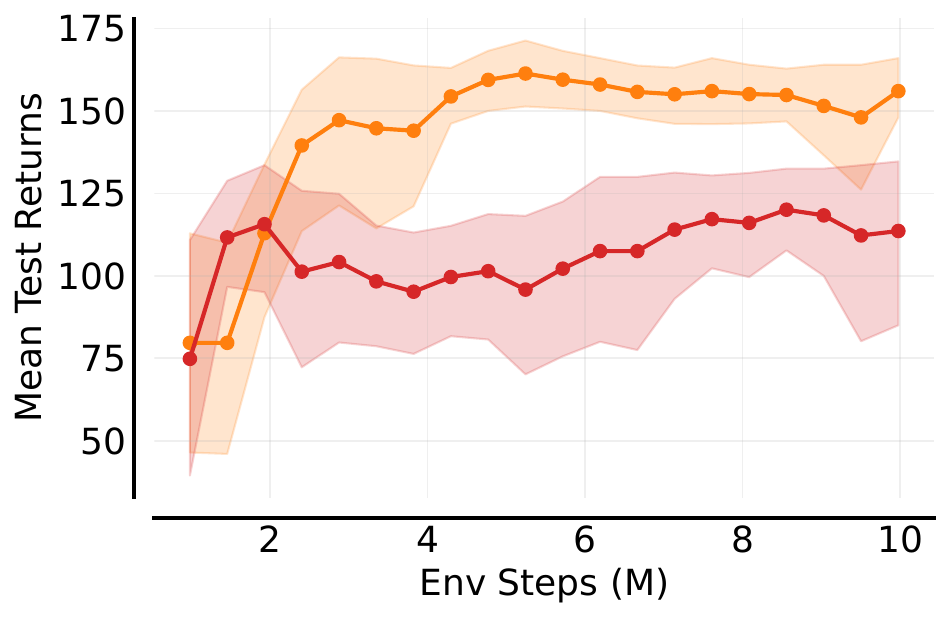}
    \caption{Counter Circuit}\label{fig:counter_circuit}
  \end{subfigure}

  \begin{subfigure}{0.32\textwidth}
    \centering
    \includegraphics[width=\linewidth]{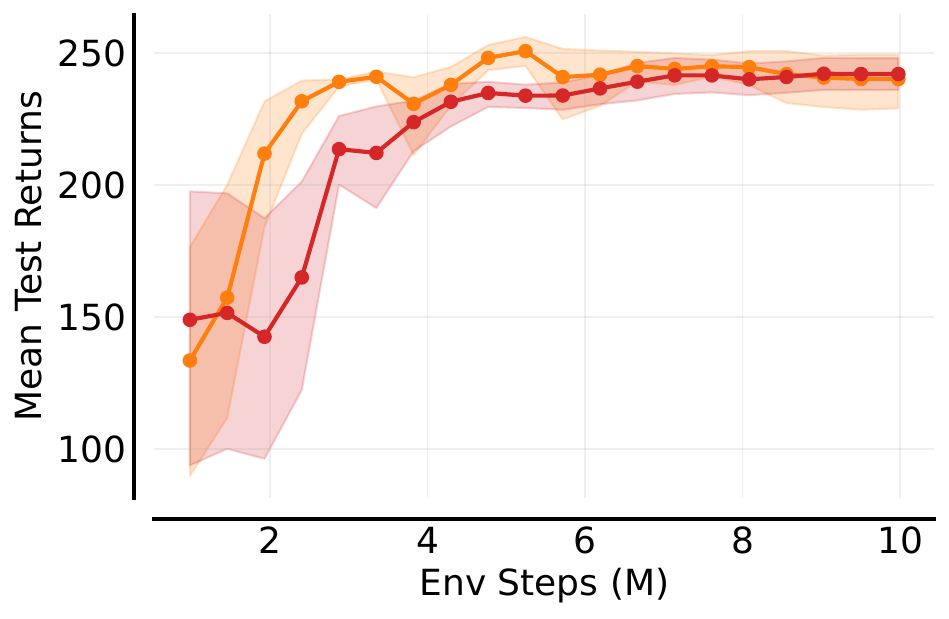}
    \caption{Cramped Room}\label{fig:cramped_room}
  \end{subfigure}\hfill
  \begin{subfigure}{0.32\textwidth}
    \centering
    \includegraphics[width=\linewidth]{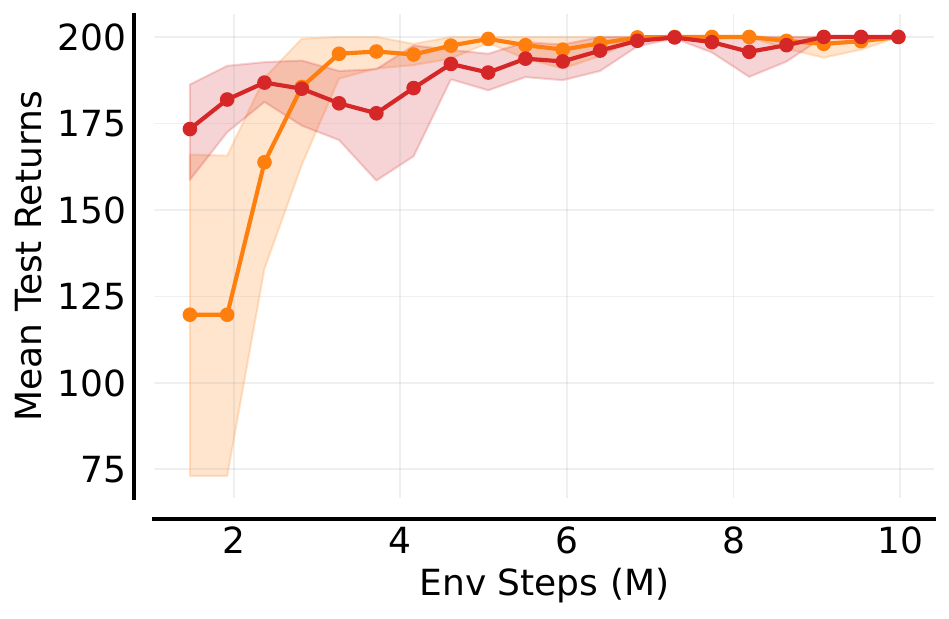}
    \caption{Forced Coordination}\label{fig:forced_coord}
  \end{subfigure}
  
  \caption{Mean test returns with 95\% confidence intervals in Overcooked V1. Overcooked V1 is fully observable, a centralised critic with concatenated global state is unnecessary here, so we report IPPO only. }
  \label{fig:overcooked_v1_results}
\end{figure*}

\begin{figure*}[t]
  \centering

  \begin{minipage}{\textwidth}
    \centering
    \includegraphics[width=0.8\linewidth]{figures/legend.pdf}
  \end{minipage}

  \vspace{0.75em}

  \begin{subfigure}{0.32\textwidth}
    \centering
    \includegraphics[width=\linewidth]{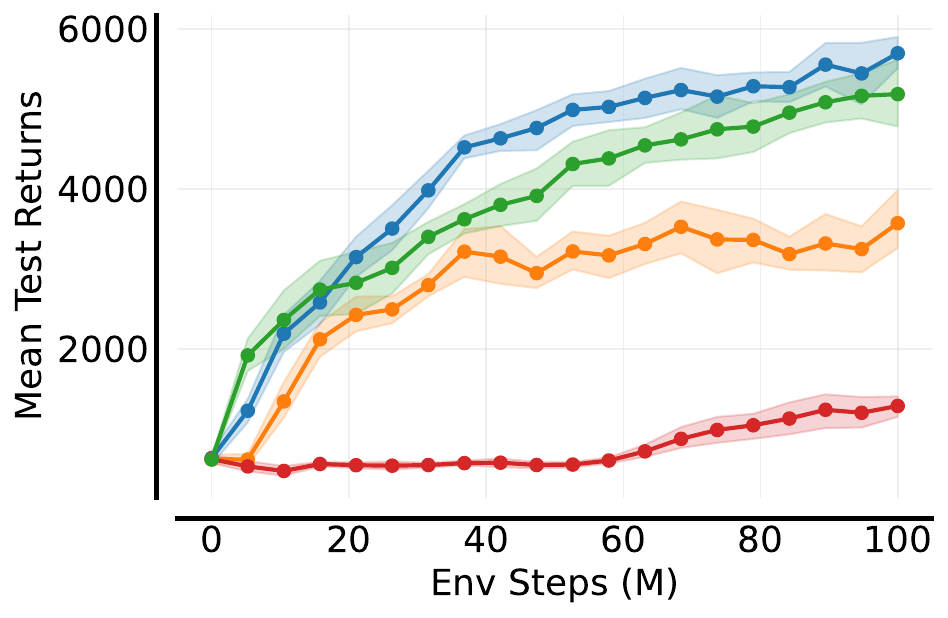}
    \caption{Ant\_4x2}\label{fig:mabrax_ant}
  \end{subfigure}\hfill
  \begin{subfigure}{0.32\textwidth}
    \centering
    \includegraphics[width=\linewidth]{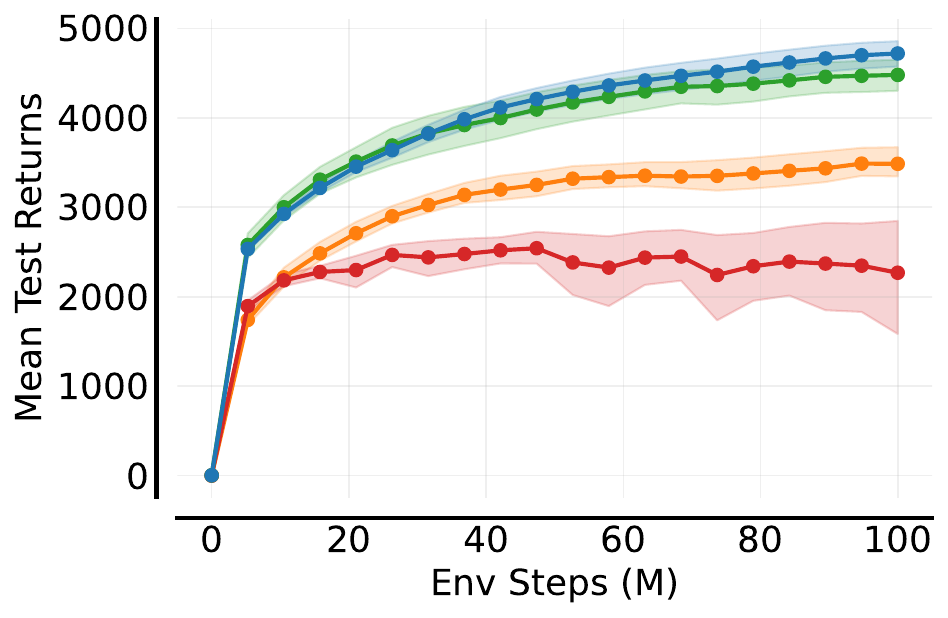}
    \caption{Halfcheetah\_6x1}\label{fig:mabrax_halfcheetah}
  \end{subfigure}\hfill
  \begin{subfigure}{0.32\textwidth}
    \centering
    \includegraphics[width=\linewidth]{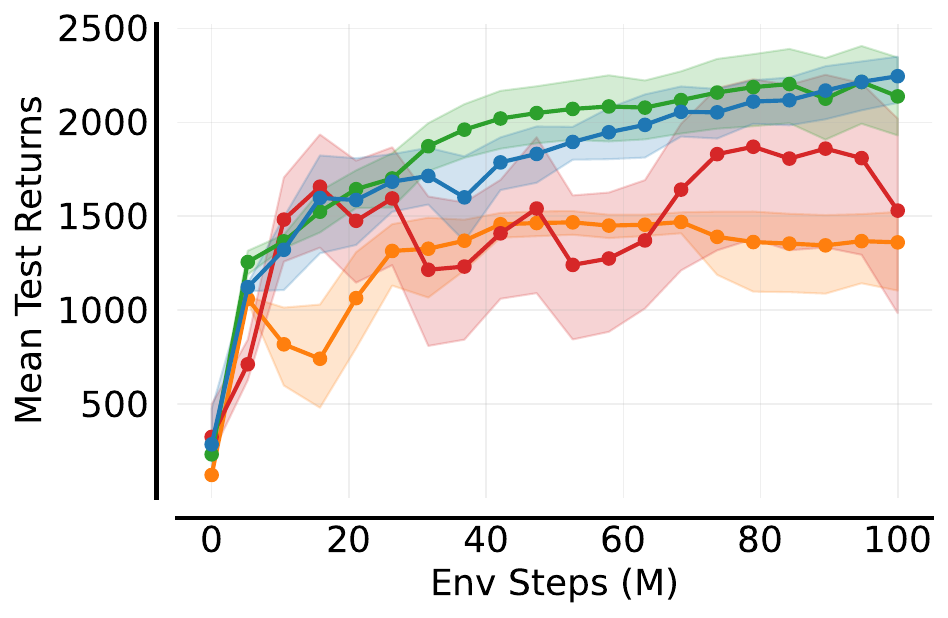}
    \caption{Hopper\_3x1}\label{fig:mabrax_hopper}
  \end{subfigure}

  \begin{subfigure}{0.32\textwidth}
    \centering
    \includegraphics[width=\linewidth]{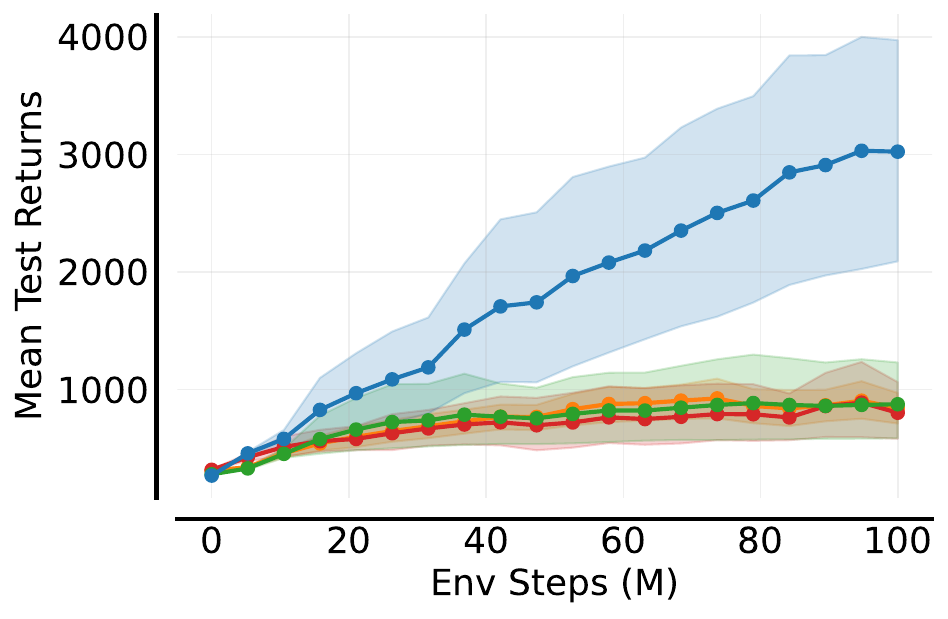}
    \caption{Walker2d\_2x3}\label{fig:mabrax_walker}
  \end{subfigure}\hfill
  \begin{subfigure}{0.32\textwidth}
    \centering
    \includegraphics[width=\linewidth]{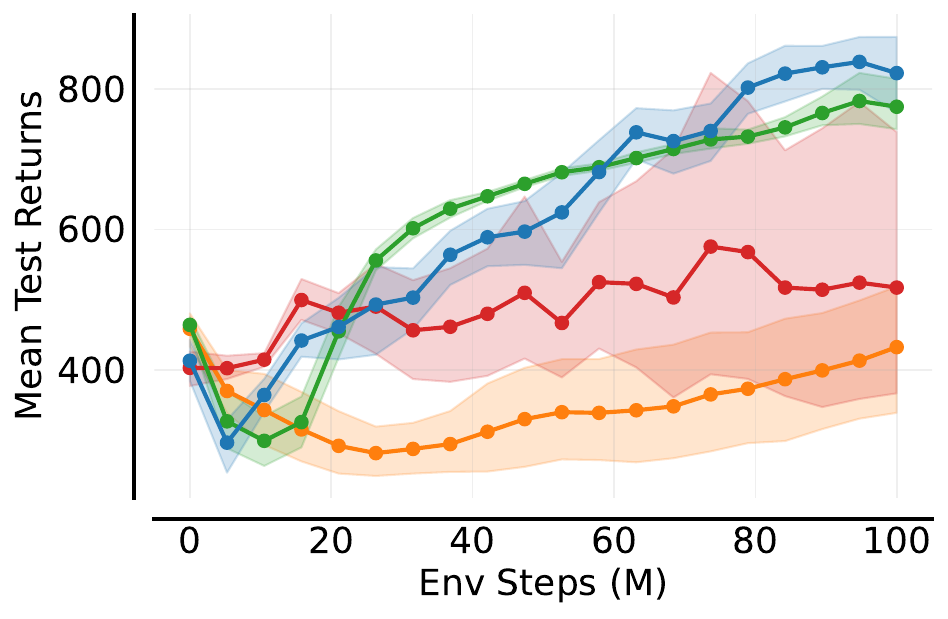}
    \caption{Humanoid\_9|8}\label{fig:mabrax_humanoid}
  \end{subfigure}
  
  \caption{Mean test returns with 95\% confidence intervals in MaBrax.}
  \label{fig:mabrax_results}
\end{figure*}

\begin{figure}
    \centering
    \includegraphics[width=0.5\linewidth]{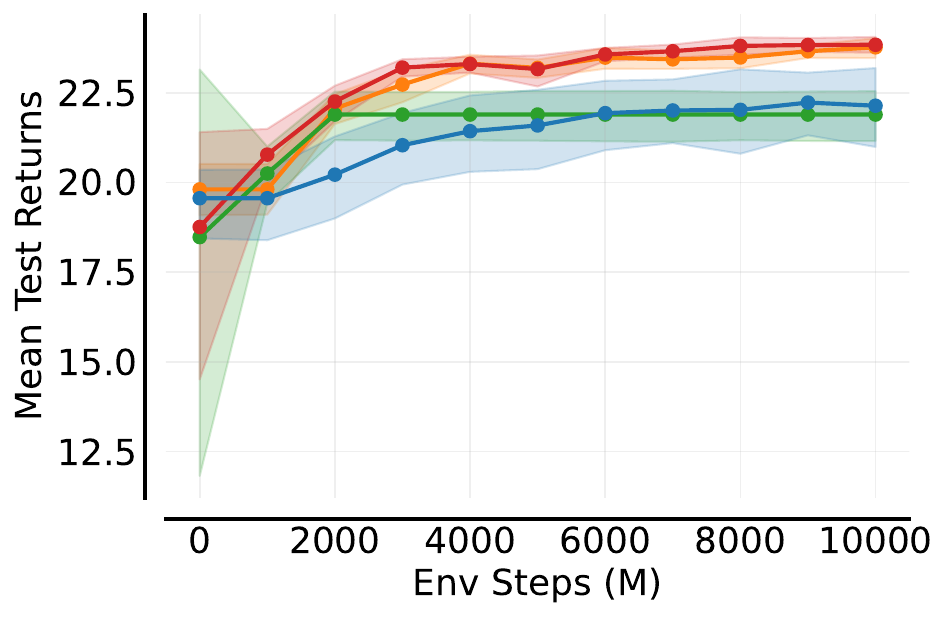}
    \caption{Mean test returns with 95\% confidence intervals in Hanabi with two players.}
    \label{fig:hanabi_results}
\end{figure}

\clearpage
\section{MPE Noise Details}
\label{sec:mpe_noise}

To add noise to our observation we define
$$
\hat{x} = x + \epsilon, \quad \epsilon \sim \mathcal{N}(0, (k \cdot \sigma_x)^2).
$$
Under this formulation, the signal-to-noise ratio (SNR) in terms of variance is
$$
\text{SNR}_{\text{power}} = \frac{\sigma_x^2}{(k \cdot \sigma_x)^2} = \frac{1}{k^2}.
$$
At the maximum scale $k=0.5$, the SNR is $4{:}1$, meaning signal variance is four times noise variance.

\clearpage 
\section{Detailed Diagnostic Measures}~\label{app:detailed_measures}

\scriptsize
\setlength{\tabcolsep}{2pt}
\renewcommand{\arraystretch}{0.95}

\begin{longtable}{lllccccc}
\caption{Normalised diagnostic metrics (mean with 95\% stratified bootstrap CI). Values exceeding the permutation-null baseline are \textbf{bolded}. For FF policies, $\mathrm{HAR}^{\mathrm{norm}}$ uses an observation-history window and $\mathrm{PIF}^{\mathrm{norm}}$/$\mathrm{DAI}^{\mathrm{norm}}$ use an observation-action history window; for RNN policies these use the hidden state. $\mathrm{AA}^{\mathrm{norm}}$ is architecture-independent.   $^{\dagger}$One or more metrics are undefined due to degenerate action entropy.} \label{tab:all_norm_metrics_long}\\
\toprule
Domain & Environment & Algorithm & $\mathrm{OAR}^{\mathrm{norm}}$ & $\mathrm{HAR}^{\mathrm{norm}}$ & $\mathrm{PIF}^{\mathrm{norm}}$ & $\mathrm{AA}^{\mathrm{norm}}$ & $\mathrm{DAI}^{\mathrm{norm}}$ \\
\midrule
\endfirsthead

\multicolumn{8}{c}{{\bfseries \tablename\ \thetable{} -- continued from previous page}} \\
\toprule
Domain & Environment & Algorithm & $\mathrm{OAR}^{\mathrm{norm}}$ & $\mathrm{HAR}^{\mathrm{norm}}$ & $\mathrm{PIF}^{\mathrm{norm}}$ & $\mathrm{AA}^{\mathrm{norm}}$ & $\mathrm{DAI}^{\mathrm{norm}}$ \\
\midrule
\endhead

\midrule
\multicolumn{8}{r}{{Continued on next page}} \\
\bottomrule
\endfoot

\bottomrule
\endlastfoot

\multirow{12}{*}{MPE} 
 & \multirow{4}{*}{\texttt{simple\_reference\_v3}} 
   & IPPO\_FF & \highlight{0.16}{$\mathbf{0.16}$ {\tiny $\mathbf{[0.15,\,0.17]}$}} & \highlight{0.10}{$\mathbf{0.10}$ {\tiny $\mathbf{[0.09,\,0.10]}$}} & $0.05$ {\tiny $[0.05,\,0.06]$} & \highlight{0.07}{$\mathbf{0.07}$ {\tiny $\mathbf{[0.07,\,0.07]}$}} & $0.06$ {\tiny $[0.05,\,0.06]$} \\
 & & IPPO\_RNN & \highlight{0.13}{$\mathbf{0.13}$ {\tiny $\mathbf{[0.11,\,0.15]}$}} & \highlight{0.10}{$\mathbf{0.10}$ {\tiny $\mathbf{[0.10,\,0.11]}$}} & \highlight{0.04}{$\mathbf{0.04}$ {\tiny $\mathbf{[0.04,\,0.04]}$}} & \highlight{0.05}{$\mathbf{0.05}$ {\tiny $\mathbf{[0.05,\,0.06]}$}} & \highlight{0.04}{$\mathbf{0.04}$ {\tiny $\mathbf{[0.04,\,0.04]}$}} \\
 & & MAPPO\_FF & \highlight{0.16}{$\mathbf{0.16}$ {\tiny $\mathbf{[0.15,\,0.17]}$}} & \highlight{0.11}{$\mathbf{0.11}$ {\tiny $\mathbf{[0.10,\,0.12]}$}} & $0.05$ {\tiny $[0.05,\,0.06]$} & \highlight{0.06}{$\mathbf{0.06}$ {\tiny $\mathbf{[0.06,\,0.06]}$}} & \highlight{0.06}{$\mathbf{0.06}$ {\tiny $\mathbf{[0.05,\,0.07]}$}} \\
 & & MAPPO\_RNN & \highlight{0.19}{$\mathbf{0.19}$ {\tiny $\mathbf{[0.17,\,0.21]}$}} & \highlight{0.07}{$\mathbf{0.07}$ {\tiny $\mathbf{[0.07,\,0.07]}$}} & \highlight{0.04}{$\mathbf{0.04}$ {\tiny $\mathbf{[0.04,\,0.04]}$}} & \highlight{0.06}{$\mathbf{0.06}$ {\tiny $\mathbf{[0.06,\,0.07]}$}} & $0.03$ {\tiny $[0.03,\,0.04]$} \\
\cmidrule{2-8}
 & \multirow{4}{*}{\texttt{simple\_speaker\_listener\_v4}} 
   & IPPO\_FF & \highlight{0.93}{$\mathbf{0.93}$ {\tiny $\mathbf{[0.80,\,1.00]}$}} & \highlight{0.10}{$\mathbf{0.10}$ {\tiny $\mathbf{[0.10,\,0.11]}$}} & \highlight{0.10}{$\mathbf{0.10}$ {\tiny $\mathbf{[0.08,\,0.11]}$}} & \highlight{0.09}{$\mathbf{0.09}$ {\tiny $\mathbf{[0.09,\,0.09]}$}} & \highlight{0.12}{$\mathbf{0.12}$ {\tiny $\mathbf{[0.10,\,0.14]}$}} \\
 & & IPPO\_RNN & \highlight{1.0}{$\mathbf{1.00}$ {\tiny $\mathbf{[1.00,\,1.00]}$}} & \highlight{0.11}{$\mathbf{0.11}$ {\tiny $\mathbf{[0.11,\,0.12]}$}} & \highlight{0.07}{$\mathbf{0.07}$ {\tiny $\mathbf{[0.06,\,0.07]}$}} & \highlight{0.07}{$\mathbf{0.07}$ {\tiny $\mathbf{[0.07,\,0.08]}$}} & \highlight{0.07}{$\mathbf{0.07}$ {\tiny $\mathbf{[0.06,\,0.08]}$}} \\
 & & MAPPO\_FF & \highlight{1.0}{$\mathbf{1.00}$ {\tiny $\mathbf{[1.00,\,1.00]}$}} & \highlight{0.12}{$\mathbf{0.12}$ {\tiny $\mathbf{[0.11,\,0.13]}$}} & \highlight{0.15}{$\mathbf{0.15}$ {\tiny $\mathbf{[0.13,\,0.17]}$}} & \highlight{0.11}{$\mathbf{0.11}$ {\tiny $\mathbf{[0.09,\,0.13]}$}} & \highlight{0.14}{$\mathbf{0.14}$ {\tiny $\mathbf{[0.12,\,0.16]}$}} \\
 & & MAPPO\_RNN & \highlight{0.92}{$\mathbf{0.92}$ {\tiny $\mathbf{[0.79,\,1.00]}$}} & \highlight{0.10}{$\mathbf{0.10}$ {\tiny $\mathbf{[0.09,\,0.10]}$}} & \highlight{0.11}{$\mathbf{0.11}$ {\tiny $\mathbf{[0.07,\,0.15]}$}} & $0.06$ {\tiny $[0.05,\,0.06]$} & \highlight{0.17}{$\mathbf{0.17}$ {\tiny $\mathbf{[0.13,\,0.21]}$}} \\
\cmidrule{2-8}
 & \multirow{4}{*}{\texttt{simple\_spread\_v3}} 
   & IPPO\_FF & \highlight{0.69}{$\mathbf{0.69}$ {\tiny $\mathbf{[0.54,\,0.80]}$}} & \highlight{0.07}{$\mathbf{0.07}$ {\tiny $\mathbf{[0.07,\,0.07]}$}} & \highlight{0.11}{$\mathbf{0.11}$ {\tiny $\mathbf{[0.10,\,0.11]}$}} & \highlight{0.07}{$\mathbf{0.07}$ {\tiny $\mathbf{[0.07,\,0.08]}$}} & \highlight{0.10}{$\mathbf{0.10}$ {\tiny $\mathbf{[0.10,\,0.11]}$}} \\
 & & IPPO\_RNN & \highlight{0.25}{$\mathbf{0.25}$ {\tiny $\mathbf{[0.24,\,0.25]}$}} & \highlight{0.09}{$\mathbf{0.09}$ {\tiny $\mathbf{[0.09,\,0.09]}$}} & \highlight{0.05}{$\mathbf{0.05}$ {\tiny $\mathbf{[0.04,\,0.05]}$}} & \highlight{0.07}{$\mathbf{0.07}$ {\tiny $\mathbf{[0.07,\,0.07]}$}} & \highlight{0.04}{$\mathbf{0.04}$ {\tiny $\mathbf{[0.04,\,0.04]}$}} \\
 & & MAPPO\_FF & \highlight{0.26}{$\mathbf{0.26}$ {\tiny $\mathbf{[0.25,\,0.28]}$}} & \highlight{0.07}{$\mathbf{0.07}$ {\tiny $\mathbf{[0.07,\,0.07]}$}} & \highlight{0.11}{$\mathbf{0.11}$ {\tiny $\mathbf{[0.11,\,0.11]}$}} & \highlight{0.07}{$\mathbf{0.07}$ {\tiny $\mathbf{[0.07,\,0.07]}$}} & \highlight{0.10}{$\mathbf{0.10}$ {\tiny $\mathbf{[0.10,\,0.10]}$}} \\
 & & MAPPO\_RNN & \highlight{0.25}{$\mathbf{0.25}$ {\tiny $\mathbf{[0.24,\,0.25]}$}} & \highlight{0.16}{$\mathbf{0.16}$ {\tiny $\mathbf{[0.16,\,0.17]}$}} & \highlight{0.18}{$\mathbf{0.18}$ {\tiny $\mathbf{[0.17,\,0.19]}$}} & \highlight{0.07}{$\mathbf{0.07}$ {\tiny $\mathbf{[0.07,\,0.07]}$}} & \highlight{0.12}{$\mathbf{0.12}$ {\tiny $\mathbf{[0.11,\,0.12]}$}} \\

\midrule
\multirow{36}{*}{SMAX-V1 maps} 
 & \multirow{4}{*}{\texttt{10m\_vs\_11m}} 
   & IPPO\_FF & \highlight{0.60}{$\mathbf{0.60}$ {\tiny $\mathbf{[0.56,\,0.64]}$}} & $0.02$ {\tiny $[0.02,\,0.02]$} & $0.01$ {\tiny $[0.01,\,0.01]$} & $0.00$ {\tiny $[0.00,\,0.00]$} & $0.01$ {\tiny $[0.01,\,0.01]$} \\
 & & IPPO\_RNN & \highlight{0.48}{$\mathbf{0.48}$ {\tiny $\mathbf{[0.47,\,0.49]}$}} & \highlight{0.06}{$\mathbf{0.06}$ {\tiny $\mathbf{[0.06,\,0.07]}$}} & $0.05$ {\tiny $[0.05,\,0.05]$} & $0.01$ {\tiny $[0.01,\,0.01]$} & $0.06$ {\tiny $[0.05,\,0.06]$} \\
 & & MAPPO\_FF & \highlight{0.73}{$\mathbf{0.73}$ {\tiny $\mathbf{[0.61,\,0.85]}$}} & \highlight{0.07}{$\mathbf{0.07}$ {\tiny $\mathbf{[0.05,\,0.11]}$}} & \highlight{0.06}{$\mathbf{0.06}$ {\tiny $\mathbf{[0.04,\,0.08]}$}} & $0.01$ {\tiny $[0.01,\,0.01]$} & \highlight{0.06}{$\mathbf{0.06}$ {\tiny $\mathbf{[0.05,\,0.08]}$}} \\
 & & MAPPO\_RNN & \highlight{0.45}{$\mathbf{0.45}$ {\tiny $\mathbf{[0.44,\,0.47]}$}} & \highlight{0.06}{$\mathbf{0.06}$ {\tiny $\mathbf{[0.06,\,0.07]}$}} & $0.04$ {\tiny $[0.04,\,0.05]$} & $0.01$ {\tiny $[0.01,\,0.01]$} & $0.07$ {\tiny $[0.07,\,0.08]$} \\
\cmidrule{2-8}
 & \multirow{4}{*}{\texttt{2s3z}} 
   & IPPO\_FF & \highlight{0.70}{$\mathbf{0.70}$ {\tiny $\mathbf{[0.67,\,0.73]}$}} & \highlight{0.08}{$\mathbf{0.08}$ {\tiny $\mathbf{[0.08,\,0.09]}$}} & $0.06$ {\tiny $[0.05,\,0.06]$} & $0.02$ {\tiny $[0.02,\,0.02]$} & $0.06$ {\tiny $[0.06,\,0.06]$} \\
 & & IPPO\_RNN & \highlight{0.56}{$\mathbf{0.56}$ {\tiny $\mathbf{[0.54,\,0.58]}$}} & \highlight{0.08}{$\mathbf{0.08}$ {\tiny $\mathbf{[0.07,\,0.08]}$}} & \highlight{0.06}{$\mathbf{0.06}$ {\tiny $\mathbf{[0.06,\,0.06]}$}} & $0.02$ {\tiny $[0.02,\,0.02]$} & \highlight{0.06}{$\mathbf{0.06}$ {\tiny $\mathbf{[0.05,\,0.06]}$}} \\
 & & MAPPO\_FF & \highlight{0.73}{$\mathbf{0.73}$ {\tiny $\mathbf{[0.66,\,0.81]}$}} & \highlight{0.08}{$\mathbf{0.08}$ {\tiny $\mathbf{[0.08,\,0.08]}$}} & $0.06$ {\tiny $[0.05,\,0.06]$} & $0.02$ {\tiny $[0.02,\,0.02]$} & \highlight{0.06}{$\mathbf{0.06}$ {\tiny $\mathbf{[0.06,\,0.07]}$}} \\
 & & MAPPO\_RNN & \highlight{0.67}{$\mathbf{0.67}$ {\tiny $\mathbf{[0.66,\,0.68]}$}} & \highlight{0.09}{$\mathbf{0.09}$ {\tiny $\mathbf{[0.08,\,0.09]}$}} & \highlight{0.06}{$\mathbf{0.06}$ {\tiny $\mathbf{[0.06,\,0.06]}$}} & $0.02$ {\tiny $[0.02,\,0.02]$} & \highlight{0.05}{$\mathbf{0.05}$ {\tiny $\mathbf{[0.05,\,0.06]}$}} \\
\cmidrule{2-8}
 & \multirow{4}{*}{\texttt{3m}} 
   & IPPO\_FF & \highlight{0.76}{$\mathbf{0.76}$ {\tiny $\mathbf{[0.72,\,0.79]}$}} & \highlight{0.09}{$\mathbf{0.09}$ {\tiny $\mathbf{[0.08,\,0.09]}$}} & $0.03$ {\tiny $[0.03,\,0.04]$} & $0.02$ {\tiny $[0.02,\,0.02]$} & $0.04$ {\tiny $[0.03,\,0.04]$} \\
 & & IPPO\_RNN & \highlight{0.66}{$\mathbf{0.66}$ {\tiny $\mathbf{[0.65,\,0.67]}$}} & \highlight{0.13}{$\mathbf{0.13}$ {\tiny $\mathbf{[0.12,\,0.13]}$}} & \highlight{0.06}{$\mathbf{0.06}$ {\tiny $\mathbf{[0.06,\,0.06]}$}} & \highlight{0.03}{$\mathbf{0.03}$ {\tiny $\mathbf{[0.03,\,0.04]}$}} & \highlight{0.06}{$\mathbf{0.06}$ {\tiny $\mathbf{[0.06,\,0.07]}$}} \\
 & & MAPPO\_FF & \highlight{0.76}{$\mathbf{0.76}$ {\tiny $\mathbf{[0.73,\,0.78]}$}} & \highlight{0.11}{$\mathbf{0.11}$ {\tiny $\mathbf{[0.10,\,0.11]}$}} & \highlight{0.06}{$\mathbf{0.06}$ {\tiny $\mathbf{[0.05,\,0.07]}$}} & \highlight{0.03}{$\mathbf{0.03}$ {\tiny $\mathbf{[0.02,\,0.03]}$}} & \highlight{0.06}{$\mathbf{0.06}$ {\tiny $\mathbf{[0.05,\,0.06]}$}} \\
 & & MAPPO\_RNN & \highlight{0.69}{$\mathbf{0.69}$ {\tiny $\mathbf{[0.68,\,0.71]}$}} & \highlight{0.12}{$\mathbf{0.12}$ {\tiny $\mathbf{[0.11,\,0.12]}$}} & \highlight{0.06}{$\mathbf{0.06}$ {\tiny $\mathbf{[0.05,\,0.06]}$}} & \highlight{0.03}{$\mathbf{0.03}$ {\tiny $\mathbf{[0.03,\,0.03]}$}} & \highlight{0.07}{$\mathbf{0.07}$ {\tiny $\mathbf{[0.06,\,0.07]}$}} \\
\cmidrule{2-8}
 & \multirow{4}{*}{\texttt{3s5z}} 
   & IPPO\_FF & \highlight{0.64}{$\mathbf{0.64}$ {\tiny $\mathbf{[0.62,\,0.67]}$}} & $0.07$ {\tiny $[0.07,\,0.07]$} & \highlight{0.06}{$\mathbf{0.06}$ {\tiny $\mathbf{[0.06,\,0.07]}$}} & $0.01$ {\tiny $[0.01,\,0.01]$} & $0.06$ {\tiny $[0.06,\,0.07]$} \\
 & & IPPO\_RNN & \highlight{0.55}{$\mathbf{0.55}$ {\tiny $\mathbf{[0.54,\,0.56]}$}} & \highlight{0.07}{$\mathbf{0.07}$ {\tiny $\mathbf{[0.06,\,0.08]}$}} & $0.07$ {\tiny $[0.07,\,0.08]$} & $0.02$ {\tiny $[0.02,\,0.02]$} & $0.07$ {\tiny $[0.06,\,0.07]$} \\
 & & MAPPO\_FF & \highlight{0.70}{$\mathbf{0.70}$ {\tiny $\mathbf{[0.61,\,0.80]}$}} & $0.07$ {\tiny $[0.07,\,0.08]$} & $0.06$ {\tiny $[0.05,\,0.06]$} & $0.02$ {\tiny $[0.01,\,0.02]$} & $0.06$ {\tiny $[0.06,\,0.08]$} \\
 & & MAPPO\_RNN & \highlight{0.71}{$\mathbf{0.71}$ {\tiny $\mathbf{[0.69,\,0.73]}$}} & \highlight{0.05}{$\mathbf{0.05}$ {\tiny $\mathbf{[0.05,\,0.06]}$}} & \highlight{0.08}{$\mathbf{0.08}$ {\tiny $\mathbf{[0.07,\,0.08]}$}} & $0.02$ {\tiny $[0.02,\,0.02]$} & $0.08$ {\tiny $[0.07,\,0.08]$} \\
\cmidrule{2-8}
 & \multirow{4}{*}{\texttt{3s5z\_vs\_3s6z}} 
   & IPPO\_FF & \highlight{0.62}{$\mathbf{0.62}$ {\tiny $\mathbf{[0.57,\,0.67]}$}} & $0.07$ {\tiny $[0.07,\,0.07]$} & $0.06$ {\tiny $[0.05,\,0.06]$} & $0.01$ {\tiny $[0.01,\,0.01]$} & $0.06$ {\tiny $[0.05,\,0.06]$} \\
 & & IPPO\_RNN & \highlight{0.53}{$\mathbf{0.53}$ {\tiny $\mathbf{[0.50,\,0.56]}$}} & \highlight{0.07}{$\mathbf{0.07}$ {\tiny $\mathbf{[0.07,\,0.08]}$}} & \highlight{0.08}{$\mathbf{0.08}$ {\tiny $\mathbf{[0.07,\,0.08]}$}} & $0.02$ {\tiny $[0.02,\,0.02]$} & \highlight{0.07}{$\mathbf{0.07}$ {\tiny $\mathbf{[0.07,\,0.08]}$}} \\
 & & MAPPO\_FF & \highlight{0.68}{$\mathbf{0.68}$ {\tiny $\mathbf{[0.59,\,0.78]}$}} & $0.07$ {\tiny $[0.07,\,0.07]$} & $0.05$ {\tiny $[0.05,\,0.05]$} & $0.01$ {\tiny $[0.01,\,0.02]$} & $0.06$ {\tiny $[0.05,\,0.06]$} \\
 & & MAPPO\_RNN & \highlight{0.60}{$\mathbf{0.60}$ {\tiny $\mathbf{[0.59,\,0.62]}$}} & \highlight{0.06}{$\mathbf{0.06}$ {\tiny $\mathbf{[0.06,\,0.06]}$}} & $0.05$ {\tiny $[0.05,\,0.05]$} & $0.01$ {\tiny $[0.01,\,0.02]$} & $0.05$ {\tiny $[0.05,\,0.05]$} \\
\cmidrule{2-8}
 & \multirow{4}{*}{\texttt{3s\_vs\_5z}} 
   & IPPO\_FF & \highlight{0.40}{$\mathbf{0.40}$ {\tiny $\mathbf{[0.38,\,0.42]}$}} & \highlight{0.09}{$\mathbf{0.09}$ {\tiny $\mathbf{[0.08,\,0.09]}$}} & \highlight{0.07}{$\mathbf{0.07}$ {\tiny $\mathbf{[0.06,\,0.07]}$}} & \highlight{0.03}{$\mathbf{0.03}$ {\tiny $\mathbf{[0.03,\,0.03]}$}} & \highlight{0.07}{$\mathbf{0.07}$ {\tiny $\mathbf{[0.06,\,0.07]}$}} \\
 & & IPPO\_RNN & \highlight{0.44}{$\mathbf{0.44}$ {\tiny $\mathbf{[0.41,\,0.47]}$}} & \highlight{0.09}{$\mathbf{0.09}$ {\tiny $\mathbf{[0.09,\,0.10]}$}} & \highlight{0.07}{$\mathbf{0.07}$ {\tiny $\mathbf{[0.06,\,0.07]}$}} & \highlight{0.03}{$\mathbf{0.03}$ {\tiny $\mathbf{[0.03,\,0.03]}$}} & \highlight{0.06}{$\mathbf{0.06}$ {\tiny $\mathbf{[0.06,\,0.06]}$}} \\
 & & MAPPO\_FF & \highlight{0.63}{$\mathbf{0.63}$ {\tiny $\mathbf{[0.46,\,0.81]}$}} & $0.08$ {\tiny $[0.08,\,0.08]$} & $0.05$ {\tiny $[0.05,\,0.05]$} & $0.03$ {\tiny $[0.02,\,0.03]$} & $0.06$ {\tiny $[0.05,\,0.06]$} \\
 & & MAPPO\_RNN & \highlight{0.41}{$\mathbf{0.41}$ {\tiny $\mathbf{[0.38,\,0.44]}$}} & \highlight{0.09}{$\mathbf{0.09}$ {\tiny $\mathbf{[0.09,\,0.09]}$}} & \highlight{0.07}{$\mathbf{0.07}$ {\tiny $\mathbf{[0.06,\,0.07]}$}} & \highlight{0.03}{$\mathbf{0.03}$ {\tiny $\mathbf{[0.03,\,0.03]}$}} & \highlight{0.06}{$\mathbf{0.06}$ {\tiny $\mathbf{[0.06,\,0.06]}$}} \\
\cmidrule{2-8}
 & \multirow{4}{*}{\texttt{5m\_vs\_6m}} 
   & IPPO\_FF & \highlight{0.68}{$\mathbf{0.68}$ {\tiny $\mathbf{[0.66,\,0.70]}$}} & $0.03$ {\tiny $[0.03,\,0.04]$} & $0.01$ {\tiny $[0.01,\,0.01]$} & \highlight{0.01}{$\mathbf{0.01}$ {\tiny $\mathbf{[0.01,\,0.01]}$}} & $0.01$ {\tiny $[0.01,\,0.01]$} \\
 & & IPPO\_RNN & \highlight{0.59}{$\mathbf{0.59}$ {\tiny $\mathbf{[0.58,\,0.61]}$}} & \highlight{0.08}{$\mathbf{0.08}$ {\tiny $\mathbf{[0.08,\,0.09]}$}} & $0.05$ {\tiny $[0.05,\,0.05]$} & $0.02$ {\tiny $[0.02,\,0.02]$} & \highlight{0.08}{$\mathbf{0.08}$ {\tiny $\mathbf{[0.08,\,0.09]}$}} \\
 & & MAPPO\_FF & \highlight{0.64}{$\mathbf{0.64}$ {\tiny $\mathbf{[0.63,\,0.66]}$}} & $0.08$ {\tiny $[0.07,\,0.08]$} & $0.05$ {\tiny $[0.05,\,0.05]$} & $0.02$ {\tiny $[0.02,\,0.02]$} & $0.05$ {\tiny $[0.05,\,0.05]$} \\
 & & MAPPO\_RNN & \highlight{0.58}{$\mathbf{0.58}$ {\tiny $\mathbf{[0.57,\,0.59]}$}} & \highlight{0.09}{$\mathbf{0.09}$ {\tiny $\mathbf{[0.08,\,0.09]}$}} & $0.05$ {\tiny $[0.04,\,0.05]$} & $0.02$ {\tiny $[0.02,\,0.02]$} & \highlight{0.10}{$\mathbf{0.10}$ {\tiny $\mathbf{[0.09,\,0.10]}$}} \\
\cmidrule{2-8}
 & \multirow{4}{*}{\texttt{6h\_vs\_8z}} 
   & IPPO\_FF & \highlight{0.57}{$\mathbf{0.57}$ {\tiny $\mathbf{[0.53,\,0.61]}$}} & \highlight{0.09}{$\mathbf{0.09}$ {\tiny $\mathbf{[0.08,\,0.09]}$}} & $0.05$ {\tiny $[0.05,\,0.06]$} & $0.02$ {\tiny $[0.02,\,0.02]$} & $0.06$ {\tiny $[0.05,\,0.06]$} \\
 & & IPPO\_RNN & \highlight{0.37}{$\mathbf{0.37}$ {\tiny $\mathbf{[0.36,\,0.38]}$}} & \highlight{0.07}{$\mathbf{0.07}$ {\tiny $\mathbf{[0.07,\,0.07]}$}} & $0.06$ {\tiny $[0.06,\,0.06]$} & \highlight{0.02}{$\mathbf{0.02}$ {\tiny $\mathbf{[0.02,\,0.02]}$}} & $0.06$ {\tiny $[0.06,\,0.06]$} \\
 & & MAPPO\_FF & \highlight{0.51}{$\mathbf{0.51}$ {\tiny $\mathbf{[0.40,\,0.65]}$}} & $0.07$ {\tiny $[0.07,\,0.08]$} & $0.05$ {\tiny $[0.05,\,0.06]$} & \highlight{0.02}{$\mathbf{0.02}$ {\tiny $\mathbf{[0.02,\,0.02]}$}} & $0.06$ {\tiny $[0.05,\,0.07]$} \\
 & & MAPPO\_RNN & \highlight{0.47}{$\mathbf{0.47}$ {\tiny $\mathbf{[0.46,\,0.48]}$}} & \highlight{0.06}{$\mathbf{0.06}$ {\tiny $\mathbf{[0.06,\,0.07]}$}} & $0.06$ {\tiny $[0.06,\,0.06]$} & $0.02$ {\tiny $[0.02,\,0.02]$} & $0.06$ {\tiny $[0.06,\,0.06]$} \\
\cmidrule{2-8}
 & \multirow{4}{*}{\texttt{8m}} 
   & IPPO\_FF & \highlight{0.61}{$\mathbf{0.61}$ {\tiny $\mathbf{[0.56,\,0.67]}$}} & $0.05$ {\tiny $[0.05,\,0.06]$} & $0.03$ {\tiny $[0.03,\,0.03]$} & $0.01$ {\tiny $[0.01,\,0.01]$} & $0.03$ {\tiny $[0.03,\,0.04]$} \\
 & & IPPO\_RNN & \highlight{0.48}{$\mathbf{0.48}$ {\tiny $\mathbf{[0.46,\,0.50]}$}} & \highlight{0.07}{$\mathbf{0.07}$ {\tiny $\mathbf{[0.07,\,0.08]}$}} & $0.05$ {\tiny $[0.05,\,0.05]$} & $0.01$ {\tiny $[0.01,\,0.02]$} & $0.06$ {\tiny $[0.06,\,0.07]$} \\
 & & MAPPO\_FF & \highlight{0.56}{$\mathbf{0.56}$ {\tiny $\mathbf{[0.54,\,0.59]}$}} & $0.07$ {\tiny $[0.06,\,0.08]$} & $0.04$ {\tiny $[0.04,\,0.05]$} & $0.01$ {\tiny $[0.01,\,0.01]$} & $0.05$ {\tiny $[0.05,\,0.05]$} \\
 & & MAPPO\_RNN & \highlight{0.50}{$\mathbf{0.50}$ {\tiny $\mathbf{[0.49,\,0.51]}$}} & \highlight{0.07}{$\mathbf{0.07}$ {\tiny $\mathbf{[0.07,\,0.08]}$}} & $0.05$ {\tiny $[0.05,\,0.05]$} & $0.01$ {\tiny $[0.01,\,0.01]$} & $0.07$ {\tiny $[0.07,\,0.07]$} \\

\midrule
\multirow{9}{*}{SMAX-V2 maps} 
 & \multirow{4}{*}{\texttt{SMAX V2\_5\_units}} 
   & IPPO\_FF & \highlight{0.29}{$\mathbf{0.29}$ {\tiny $\mathbf{[0.28,\,0.29]}$}} & \highlight{0.07}{$\mathbf{0.07}$ {\tiny $\mathbf{[0.06,\,0.07]}$}} & $0.05$ {\tiny $[0.04,\,0.06]$} & $0.01$ {\tiny $[0.01,\,0.01]$} & $0.06$ {\tiny $[0.05,\,0.07]$} \\
 & & IPPO\_RNN & \highlight{0.33}{$\mathbf{0.33}$ {\tiny $\mathbf{[0.32,\,0.33]}$}} & \highlight{0.09}{$\mathbf{0.09}$ {\tiny $\mathbf{[0.09,\,0.09]}$}} & \highlight{0.08}{$\mathbf{0.08}$ {\tiny $\mathbf{[0.08,\,0.08]}$}} & $0.01$ {\tiny $[0.01,\,0.01]$} & \highlight{0.08}{$\mathbf{0.08}$ {\tiny $\mathbf{[0.08,\,0.08]}$}} \\
 & & MAPPO\_FF & \highlight{0.36}{$\mathbf{0.36}$ {\tiny $\mathbf{[0.35,\,0.37]}$}} & \highlight{0.11}{$\mathbf{0.11}$ {\tiny $\mathbf{[0.11,\,0.11]}$}} & \highlight{0.16}{$\mathbf{0.16}$ {\tiny $\mathbf{[0.15,\,0.16]}$}} & $0.02$ {\tiny $[0.02,\,0.02]$} & \highlight{0.14}{$\mathbf{0.14}$ {\tiny $\mathbf{[0.14,\,0.15]}$}} \\
 & & MAPPO\_RNN & \highlight{0.35}{$\mathbf{0.35}$ {\tiny $\mathbf{[0.34,\,0.35]}$}} & \highlight{0.12}{$\mathbf{0.12}$ {\tiny $\mathbf{[0.12,\,0.13]}$}} & \highlight{0.09}{$\mathbf{0.09}$ {\tiny $\mathbf{[0.09,\,0.09]}$}} & $0.02$ {\tiny $[0.02,\,0.02]$} & \highlight{0.09}{$\mathbf{0.09}$ {\tiny $\mathbf{[0.09,\,0.09]}$}} \\
\cmidrule{2-8}
& \multirow{4}{*}{\texttt{SMAX V2\_10\_units}} 
   & IPPO\_FF & \highlight{0.12}{$\mathbf{0.12}$ {\tiny $\mathbf{[0.10,\,0.14]}$}} & $0.04$ {\tiny $[0.03,\,0.04]$} & $0.03$ {\tiny $[0.03,\,0.04]$} & $0.00$ {\tiny $[0.00,\,0.00]$} & $0.04$ {\tiny $[0.03,\,0.04]$} \\
 & & IPPO\_RNN & \highlight{0.23}{$\mathbf{0.23}$ {\tiny $\mathbf{[0.22,\,0.23]}$}} & \highlight{0.07}{$\mathbf{0.07}$ {\tiny $\mathbf{[0.06,\,0.08]}$}} & $0.06$ {\tiny $[0.06,\,0.06]$} & $0.01$ {\tiny $[0.01,\,0.01]$} & $0.06$ {\tiny $[0.06,\,0.06]$} \\
 & & MAPPO\_FF & \highlight{0.23}{$\mathbf{0.23}$ {\tiny $\mathbf{[0.22,\,0.24]}$}} & \highlight{0.11}{$\mathbf{0.11}$ {\tiny $\mathbf{[0.10,\,0.13]}$}} & \highlight{0.08}{$\mathbf{0.08}$ {\tiny $\mathbf{[0.07,\,0.08]}$}} & $0.01$ {\tiny $[0.01,\,0.01]$} & \highlight{0.08}{$\mathbf{0.08}$ {\tiny $\mathbf{[0.07,\,0.08]}$}} \\
 & & MAPPO\_RNN & \highlight{0.26}{$\mathbf{0.26}$ {\tiny $\mathbf{[0.25,\,0.26]}$}} & \highlight{0.09}{$\mathbf{0.09}$ {\tiny $\mathbf{[0.08,\,0.09]}$}} & $0.07$ {\tiny $[0.07,\,0.07]$} & $0.01$ {\tiny $[0.01,\,0.01]$} & $0.07$ {\tiny $[0.07,\,0.08]$} \\
\cmidrule{2-8}
 & \multirow{4}{*}{\texttt{SMAX V2\_20\_units}} 
 & IPPO\_FF & $0.00$ {\tiny $[0.00,\,0.00]$} & $0.02$ {\tiny $[0.02,\,0.02]$} & $0.01$ {\tiny $[0.01,\,0.02]$} & $0.00$ {\tiny $[0.00,\,0.00]$} & $0.02$ {\tiny $[0.02,\,0.02]$} \\
 & & IPPO\_RNN & \highlight{0.01}{$\mathbf{0.01}$ {\tiny $\mathbf{[0.00,\,0.02]}$}} & \highlight{0.05}{$\mathbf{0.05}$ {\tiny $\mathbf{[0.04,\,0.05]}$}} & $0.04$ {\tiny $[0.04,\,0.04]$} & $0.00$ {\tiny $[0.00,\,0.00]$} & $0.04$ {\tiny $[0.04,\,0.04]$} \\
 & & MAPPO\_FF & \highlight{0.00}{$\mathbf{0.00}$ {\tiny $\mathbf{[0.00,\,0.00]}$}} & \highlight{0.06}{$\mathbf{0.06}$ {\tiny $\mathbf{[0.05,\,0.07]}$}} & $0.02$ {\tiny $[0.01,\,0.02]$} & $0.00$ {\tiny $[0.00,\,0.00]$} & $0.02$ {\tiny $[0.02,\,0.02]$} \\
 & & MAPPO\_RNN & \highlight{0.00}{$\mathbf{0.00}$ {\tiny $\mathbf{[0.00,\,0.01]}$}} & \highlight{0.06}{$\mathbf{0.06}$ {\tiny $\mathbf{[0.05,\,0.07]}$}} & $0.04$ {\tiny $[0.04,\,0.05]$} & $0.00$ {\tiny $[0.00,\,0.00]$} & $0.04$ {\tiny $[0.04,\,0.05]$} \\

\pagebreak[4]
\multirow{20}{*}{MaBrax} 
& \multirow{4}{*}{\texttt{ant\_4x2}} 
   & IPPO\_FF\_NoPS & \highlight{0.65}{$\mathbf{0.65}$ {\tiny $\mathbf{[0.51,\,0.79]}$}} & $0.46$ {\tiny $[0.35,\,0.61]$} & $0.50$ {\tiny $[0.39,\,0.63]$} & $0.00$ {\tiny $[0.00,\,0.01]$} & $0.50$ {\tiny $[0.38,\,0.63]$} \\
 & & IPPO\_RNN\_NoPS & \highlight{0.31}{$\mathbf{0.31}$ {\tiny $\mathbf{[0.16,\,0.49]}$}} & \highlight{0.02}{$\mathbf{0.02}$ {\tiny $\mathbf{[0.00,\,0.03]}$}} & \highlight{0.01}{$\mathbf{0.01}$ {\tiny $\mathbf{[0.00,\,0.02]}$}} & \highlight{0.01}{$\mathbf{0.01}$ {\tiny $\mathbf{[0.00,\,0.02]}$}} & \highlight{0.04}{$\mathbf{0.04}$ {\tiny $\mathbf{[0.01,\,0.08]}$}} \\
 & & MAPPO\_FF\_NoPS & \highlight{0.58}{$\mathbf{0.58}$ {\tiny $\mathbf{[0.50,\,0.70]}$}} & $0.31$ {\tiny $[0.21,\,0.47]$} & $0.38$ {\tiny $[0.27,\,0.54]$} & $0.00$ {\tiny $[0.00,\,0.00]$} & $0.37$ {\tiny $[0.26,\,0.52]$} \\
 & & MAPPO\_RNN\_NoPS & \highlight{0.51}{$\mathbf{0.51}$ {\tiny $\mathbf{[0.42,\,0.63]}$}} & $0.01$ {\tiny $[0.01,\,0.01]$} & \highlight{0.01}{$\mathbf{0.01}$ {\tiny $\mathbf{[0.00,\,0.01]}$}} & $0.00$ {\tiny $[0.00,\,0.00]$} & \highlight{0.02}{$\mathbf{0.02}$ {\tiny $\mathbf{[0.02,\,0.03]}$}} \\
\cmidrule{2-8}
 & \multirow{4}{*}{\texttt{halfcheetah\_6x1}} 
   & IPPO\_FF\_NoPS & \highlight{0.13}{$\mathbf{0.13}$ {\tiny $\mathbf{[0.11,\,0.15]}$}} & \highlight{0.05}{$\mathbf{0.05}$ {\tiny $\mathbf{[0.05,\,0.06]}$}} & $0.06$ {\tiny $[0.06,\,0.06]$} & $0.00$ {\tiny $[0.00,\,0.00]$} & $0.06$ {\tiny $[0.06,\,0.06]$} \\
 & & IPPO\_RNN\_NoPS & \highlight{1.0}{$\mathbf{1.00}$ {\tiny $\mathbf{[1.00,\,1.00]}$}} & $0.03$ {\tiny $[0.01,\,0.05]$} & $0.02$ {\tiny $[0.01,\,0.03]$} & $0.03$ {\tiny $[0.01,\,0.05]$} & $0.05$ {\tiny $[0.02,\,0.09]$} \\
 & & MAPPO\_FF\_NoPS & ---$^{\dagger}$ & \highlight{0.53}{$\mathbf{0.53}$ {\tiny $\mathbf{[0.30,\,0.79]}$}} & \highlight{0.62}{$\mathbf{0.62}$ {\tiny $\mathbf{[0.37,\,0.87]}$}} & ---$^{\dagger}$ & $0.59$ {\tiny $[0.36,\,0.82]$} \\
 & & MAPPO\_RNN\_NoPS & \highlight{1.0}{$\mathbf{1.00}$ {\tiny $\mathbf{[1.00,\,1.00]}$}} & $0.02$ {\tiny $[0.01,\,0.02]$} & $0.02$ {\tiny $[0.01,\,0.04]$} & $0.01$ {\tiny $[0.01,\,0.01]$} & \highlight{0.09}{$\mathbf{0.09}$ {\tiny $\mathbf{[0.03,\,0.17]}$}} \\
\cmidrule{2-8}
 & \multirow{4}{*}{\texttt{hopper\_3x1}} 
   & IPPO\_FF\_NoPS & \highlight{0.29}{$\mathbf{0.29}$ {\tiny $\mathbf{[0.20,\,0.40]}$}} & \highlight{0.12}{$\mathbf{0.12}$ {\tiny $\mathbf{[0.10,\,0.13]}$}} & $0.14$ {\tiny $[0.12,\,0.16]$} & $0.00$ {\tiny $[0.00,\,0.00]$} & $0.12$ {\tiny $[0.10,\,0.13]$} \\
 & & IPPO\_RNN\_NoPS & ---$^{\dagger}$ & ---$^{\dagger}$ & ---$^{\dagger}$ & ---$^{\dagger}$ & \highlight{0.10}{$\mathbf{0.10}$ ${\dagger}$} \\
 & & MAPPO\_FF\_NoPS & \highlight{1.0}{$\mathbf{1.00}$ {\tiny $\mathbf{[1.00,\,1.00]}$}} & \highlight{0.28}{$\mathbf{0.28}$ {\tiny $\mathbf{[0.21,\,0.34]}$}} & \highlight{0.35}{$\mathbf{0.35}$ {\tiny $\mathbf{[0.25,\,0.47]}$}} & \highlight{0.02}{$\mathbf{0.02}$ {\tiny $\mathbf{[0.00,\,0.03]}$}} & \highlight{0.36}{$\mathbf{0.36}$ {\tiny $\mathbf{[0.25,\,0.47]}$}} \\
 & & MAPPO\_RNN\_NoPS & \highlight{0.97}{$\mathbf{0.97}$ {\tiny $\mathbf{[0.93,\,1.00]}$}} & \highlight{0.01}{$\mathbf{0.01}$ {\tiny $\mathbf{[0.01,\,0.02]}$}} & \highlight{0.02}{$\mathbf{0.02}$ {\tiny $\mathbf{[0.00,\,0.03]}$}} & $0.00$ {\tiny $[0.00,\,0.01]$} & \highlight{0.05}{$\mathbf{0.05}$ {\tiny $\mathbf{[0.02,\,0.11]}$}} \\
\cmidrule{2-8}
 & \multirow{4}{*}{\texttt{humanoid\_9|8}} 
   & IPPO\_FF\_NoPS & $0.00$ {\tiny $[0.00,\,0.00]$} & $0.03$ {\tiny $[0.03,\,0.04]$} & \highlight{0.03}{$\mathbf{0.03}$ {\tiny $\mathbf{[0.03,\,0.04]}$}} & $0.00$ {\tiny $[0.00,\,0.00]$} & \highlight{0.03}{$\mathbf{0.03}$ {\tiny $\mathbf{[0.03,\,0.04]}$}} \\
 & & IPPO\_RNN\_NoPS & ---$^{\dagger}$ & ---$^{\dagger}$ & ---$^{\dagger}$ & ---$^{\dagger}$ & ---$^{\dagger}$ \\
 & & MAPPO\_FF\_NoPS & ---$^{\dagger}$ & $0.07$ {\tiny $[0.06,\,0.09]$} & \highlight{0.16}{$\mathbf{0.16}$ {\tiny $\mathbf{[0.06,\,0.35]}$}} & ---$^{\dagger}$ & $0.07$ {\tiny $[0.06,\,0.09]$} \\
 & & MAPPO\_RNN\_NoPS & \highlight{0.15}{$\mathbf{0.15}$ {\tiny $\mathbf{[0.06,\,0.25]}$}} & \highlight{0.01}{$\mathbf{0.01}$ {\tiny $\mathbf{[0.00,\,0.02]}$}} & $0.01$ {\tiny $[0.00,\,0.01]$} & $0.01$ {\tiny $[0.00,\,0.02]$} & \highlight{0.08}{$\mathbf{0.08}$ {\tiny $\mathbf{[0.01,\,0.18]}$}} \\
\cmidrule{2-8}
 & \multirow{4}{*}{\texttt{walker2d\_2x3}} 
   & IPPO\_FF\_NoPS & \highlight{0.56}{$\mathbf{0.56}$ {\tiny $\mathbf{[0.46,\,0.69]}$}} & \highlight{0.12}{$\mathbf{0.12}$ {\tiny $\mathbf{[0.11,\,0.14]}$}} & \highlight{0.15}{$\mathbf{0.15}$ {\tiny $\mathbf{[0.12,\,0.19]}$}} & $0.00$ {\tiny $[0.00,\,0.00]$} & \highlight{0.14}{$\mathbf{0.14}$ {\tiny $\mathbf{[0.11,\,0.16]}$}} \\
 & & IPPO\_RNN\_NoPS & \highlight{1.0}{$\mathbf{1.00}$ {\tiny $\mathbf{[1.00,\,1.00]}$}} & $0.01$ {\tiny $[0.00,\,0.03]$} & \highlight{0.01}{$\mathbf{0.01}$ {\tiny $\mathbf{[0.01,\,0.03]}$}} & \highlight{0.02}{$\mathbf{0.02}$ {\tiny $\mathbf{[0.00,\,0.04]}$}} & $0.04$ {\tiny $[0.02,\,0.10]$} \\
 & & MAPPO\_FF\_NoPS & \highlight{1.0}{$\mathbf{1.00}$ ${\dagger}$} & $0.22$ {\tiny $[0.15,\,0.30]$} & $0.35$ {\tiny $[0.23,\,0.49]$} & \highlight{0.13}{$\mathbf{0.13}$ ${\dagger}$} & \highlight{0.29}{$\mathbf{0.29}$ {\tiny $\mathbf{[0.21,\,0.40]}$}} \\
 & & MAPPO\_RNN\_NoPS & \highlight{1.0}{$\mathbf{1.00}$ ${\dagger}$} & \highlight{0.03}{$\mathbf{0.03}$ ${\dagger}$} & $0.01$ ${\dagger}$ & $0.00$ ${\dagger}$ & \highlight{0.10}{$\mathbf{0.10}$ {\tiny $\mathbf{[0.03,\,0.17]}$}} \\

\midrule
\multirow{4}{*}{Hanabi} 
 & \multirow{4}{*}{\texttt{Two Players}} 
   & IPPO\_FF & \highlight{0.11}{$\mathbf{0.11}$ {\tiny $\mathbf{[0.10,\,0.12]}$}} & \highlight{0.04}{$\mathbf{0.04}$ {\tiny $\mathbf{[0.04,\,0.05]}$}} & $0.02$ {\tiny $[0.02,\,0.02]$} & $0.01$ {\tiny $[0.01,\,0.01]$} & $0.02$ {\tiny $[0.02,\,0.02]$} \\
 & & IPPO\_RNN & \highlight{0.10}{$\mathbf{0.10}$ {\tiny $\mathbf{[0.10,\,0.11]}$}} & \highlight{0.05}{$\mathbf{0.05}$ {\tiny $\mathbf{[0.04,\,0.05]}$}} & $0.03$ {\tiny $[0.03,\,0.03]$} & $0.01$ {\tiny $[0.01,\,0.01]$} & \highlight{0.05}{$\mathbf{0.05}$ {\tiny $\mathbf{[0.05,\,0.05]}$}} \\
 & & MAPPO\_FF & \highlight{0.12}{$\mathbf{0.12}$ {\tiny $\mathbf{[0.10,\,0.13]}$}} & \highlight{0.04}{$\mathbf{0.04}$ {\tiny $\mathbf{[0.04,\,0.05]}$}} & $0.02$ {\tiny $[0.02,\,0.02]$} & $0.01$ {\tiny $[0.01,\,0.01]$} & $0.02$ {\tiny $[0.02,\,0.02]$} \\
 & & MAPPO\_RNN & \highlight{0.14}{$\mathbf{0.14}$ {\tiny $\mathbf{[0.13,\,0.15]}$}} & \highlight{0.05}{$\mathbf{0.05}$ {\tiny $\mathbf{[0.04,\,0.05]}$}} & $0.03$ {\tiny $[0.03,\,0.03]$} & $0.01$ {\tiny $[0.00,\,0.01]$} & \highlight{0.05}{$\mathbf{0.05}$ {\tiny $\mathbf{[0.05,\,0.06]}$}} \\

\midrule
\multirow{10}{*}{\shortstack[l]{Overcooked\\V1 }} 
 & \multirow{2}{*}{\texttt{asymm\_advantages}} 
   & IPPO\_FF & \highlight{0.77}{$\mathbf{0.77}$ {\tiny $\mathbf{[0.77,\,0.77]}$}} & \highlight{0.10}{$\mathbf{0.10}$ {\tiny $\mathbf{[0.09,\,0.10]}$}} & \highlight{0.08}{$\mathbf{0.08}$ {\tiny $\mathbf{[0.08,\,0.09]}$}} & \highlight{0.01}{$\mathbf{0.01}$ {\tiny $\mathbf{[0.01,\,0.01]}$}} & \highlight{0.08}{$\mathbf{0.08}$ {\tiny $\mathbf{[0.08,\,0.08]}$}} \\
 & & IPPO\_RNN & \highlight{0.73}{$\mathbf{0.73}$ {\tiny $\mathbf{[0.71,\,0.75]}$}} & \highlight{0.19}{$\mathbf{0.19}$ {\tiny $\mathbf{[0.17,\,0.22]}$}} & \highlight{0.06}{$\mathbf{0.06}$ {\tiny $\mathbf{[0.03,\,0.09]}$}} & \highlight{0.01}{$\mathbf{0.01}$ {\tiny $\mathbf{[0.01,\,0.01]}$}} & \highlight{0.07}{$\mathbf{0.07}$ {\tiny $\mathbf{[0.05,\,0.10]}$}} \\
\cmidrule{2-8}
 & \multirow{2}{*}{\texttt{coord\_ring}} 
   & IPPO\_FF & \highlight{0.74}{$\mathbf{0.74}$ {\tiny $\mathbf{[0.74,\,0.75]}$}} & \highlight{0.17}{$\mathbf{0.17}$ {\tiny $\mathbf{[0.16,\,0.18]}$}} & $0.02$ {\tiny $[0.02,\,0.03]$} & \highlight{0.01}{$\mathbf{0.01}$ {\tiny $\mathbf{[0.01,\,0.01]}$}} & $0.03$ {\tiny $[0.03,\,0.03]$} \\
 & & IPPO\_RNN & \highlight{0.74}{$\mathbf{0.74}$ {\tiny $\mathbf{[0.71,\,0.76]}$}} & \highlight{0.09}{$\mathbf{0.09}$ {\tiny $\mathbf{[0.07,\,0.10]}$}} & $0.03$ {\tiny $[0.02,\,0.04]$} & \highlight{0.01}{$\mathbf{0.01}$ {\tiny $\mathbf{[0.01,\,0.01]}$}} & $0.03$ {\tiny $[0.02,\,0.04]$} \\
\cmidrule{2-8}
 & \multirow{2}{*}{\texttt{counter\_circuit}} 
   & IPPO\_FF & \highlight{0.75}{$\mathbf{0.75}$ {\tiny $\mathbf{[0.74,\,0.75]}$}} & \highlight{0.14}{$\mathbf{0.14}$ {\tiny $\mathbf{[0.13,\,0.15]}$}} & $0.01$ {\tiny $[0.01,\,0.02]$} & \highlight{0.01}{$\mathbf{0.01}$ {\tiny $\mathbf{[0.01,\,0.01]}$}} & $0.02$ {\tiny $[0.02,\,0.03]$} \\
 & & IPPO\_RNN & \highlight{0.65}{$\mathbf{0.65}$ {\tiny $\mathbf{[0.55,\,0.73]}$}} & \highlight{0.11}{$\mathbf{0.11}$ {\tiny $\mathbf{[0.07,\,0.14]}$}} & $0.03$ {\tiny $[0.02,\,0.04]$} & \highlight{0.01}{$\mathbf{0.01}$ {\tiny $\mathbf{[0.01,\,0.01]}$}} & \highlight{0.03}{$\mathbf{0.03}$ {\tiny $\mathbf{[0.02,\,0.05]}$}} \\
\cmidrule{2-8}
 & \multirow{2}{*}{\texttt{cramped\_room}} 
   & IPPO\_FF & \highlight{0.73}{$\mathbf{0.73}$ {\tiny $\mathbf{[0.72,\,0.74]}$}} & \highlight{0.16}{$\mathbf{0.16}$ {\tiny $\mathbf{[0.15,\,0.17]}$}} & $0.03$ {\tiny $[0.02,\,0.03]$} & \highlight{0.02}{$\mathbf{0.02}$ {\tiny $\mathbf{[0.01,\,0.02]}$}} & $0.04$ {\tiny $[0.03,\,0.04]$} \\
 & & IPPO\_RNN & \highlight{0.72}{$\mathbf{0.72}$ {\tiny $\mathbf{[0.71,\,0.72]}$}} & \highlight{0.08}{$\mathbf{0.08}$ {\tiny $\mathbf{[0.07,\,0.09]}$}} & $0.03$ {\tiny $[0.02,\,0.04]$} & \highlight{0.01}{$\mathbf{0.01}$ {\tiny $\mathbf{[0.01,\,0.02]}$}} & $0.04$ {\tiny $[0.02,\,0.06]$} \\
\cmidrule{2-8}
 & \multirow{2}{*}{\texttt{forced\_coord}} 
   & IPPO\_FF & \highlight{0.74}{$\mathbf{0.74}$ {\tiny $\mathbf{[0.74,\,0.75]}$}} & \highlight{0.15}{$\mathbf{0.15}$ {\tiny $\mathbf{[0.14,\,0.16]}$}} & $0.05$ {\tiny $[0.04,\,0.05]$} & \highlight{0.02}{$\mathbf{0.02}$ {\tiny $\mathbf{[0.02,\,0.02]}$}} & $0.06$ {\tiny $[0.05,\,0.06]$} \\
 & & IPPO\_RNN & \highlight{0.73}{$\mathbf{0.73}$ {\tiny $\mathbf{[0.72,\,0.74]}$}} & \highlight{0.12}{$\mathbf{0.12}$ {\tiny $\mathbf{[0.11,\,0.13]}$}} & $0.01$ {\tiny $[0.01,\,0.01]$} & \highlight{0.01}{$\mathbf{0.01}$ {\tiny $\mathbf{[0.01,\,0.02]}$}} & $0.01$ {\tiny $[0.01,\,0.02]$} \\

\midrule
\multirow{44}{*}{Overcooked V2} 
 & \multirow{4}{*}{\texttt{asymm\_advantages}} 
   & IPPO\_FF & \highlight{0.63}{$\mathbf{0.63}$ {\tiny $\mathbf{[0.47,\,0.75]}$}} & \highlight{0.24}{$\mathbf{0.24}$ {\tiny $\mathbf{[0.22,\,0.26]}$}} & \highlight{0.12}{$\mathbf{0.12}$ {\tiny $\mathbf{[0.11,\,0.14]}$}} & \highlight{0.02}{$\mathbf{0.02}$ {\tiny $\mathbf{[0.01,\,0.02]}$}} & \highlight{0.13}{$\mathbf{0.13}$ {\tiny $\mathbf{[0.11,\,0.14]}$}} \\
 & & IPPO\_RNN & \highlight{0.63}{$\mathbf{0.63}$ {\tiny $\mathbf{[0.47,\,0.75]}$}} & \highlight{0.28}{$\mathbf{0.28}$ {\tiny $\mathbf{[0.21,\,0.35]}$}} & \highlight{0.10}{$\mathbf{0.10}$ {\tiny $\mathbf{[0.05,\,0.16]}$}} & \highlight{0.01}{$\mathbf{0.01}$ {\tiny $\mathbf{[0.01,\,0.01]}$}} & \highlight{0.11}{$\mathbf{0.11}$ {\tiny $\mathbf{[0.05,\,0.17]}$}} \\
 & & MAPPO\_FF & \highlight{0.77}{$\mathbf{0.77}$ {\tiny $\mathbf{[0.77,\,0.78]}$}} & \highlight{0.24}{$\mathbf{0.24}$ {\tiny $\mathbf{[0.22,\,0.25]}$}} & \highlight{0.13}{$\mathbf{0.13}$ {\tiny $\mathbf{[0.11,\,0.15]}$}} & \highlight{0.02}{$\mathbf{0.02}$ {\tiny $\mathbf{[0.02,\,0.02]}$}} & \highlight{0.13}{$\mathbf{0.13}$ {\tiny $\mathbf{[0.12,\,0.15]}$}} \\
 & & MAPPO\_RNN & \highlight{0.76}{$\mathbf{0.76}$ {\tiny $\mathbf{[0.75,\,0.76]}$}} & \highlight{0.43}{$\mathbf{0.43}$ {\tiny $\mathbf{[0.41,\,0.46]}$}} & \highlight{0.06}{$\mathbf{0.06}$ {\tiny $\mathbf{[0.05,\,0.07]}$}} & \highlight{0.01}{$\mathbf{0.01}$ {\tiny $\mathbf{[0.01,\,0.01]}$}} & \highlight{0.06}{$\mathbf{0.06}$ {\tiny $\mathbf{[0.05,\,0.08]}$}} \\
\cmidrule{2-8}
 & \multirow{4}{*}{\texttt{coord\_ring}} 
   & IPPO\_FF & \highlight{0.73}{$\mathbf{0.73}$ {\tiny $\mathbf{[0.72,\,0.74]}$}} & \highlight{0.19}{$\mathbf{0.19}$ {\tiny $\mathbf{[0.16,\,0.22]}$}} & $0.03$ {\tiny $[0.03,\,0.04]$} & $0.01$ {\tiny $[0.00,\,0.01]$} & $0.04$ {\tiny $[0.03,\,0.04]$} \\
 & & IPPO\_RNN & \highlight{0.65}{$\mathbf{0.65}$ {\tiny $\mathbf{[0.50,\,0.75]}$}} & \highlight{0.18}{$\mathbf{0.18}$ {\tiny $\mathbf{[0.12,\,0.25]}$}} & $0.02$ {\tiny $[0.01,\,0.03]$} & $0.00$ {\tiny $[0.00,\,0.00]$} & \highlight{0.03}{$\mathbf{0.03}$ {\tiny $\mathbf{[0.02,\,0.05]}$}} \\
 & & MAPPO\_FF & \highlight{0.74}{$\mathbf{0.74}$ {\tiny $\mathbf{[0.73,\,0.75]}$}} & \highlight{0.21}{$\mathbf{0.21}$ {\tiny $\mathbf{[0.19,\,0.24]}$}} & $0.03$ {\tiny $[0.02,\,0.03]$} & $0.01$ {\tiny $[0.00,\,0.01]$} & $0.04$ {\tiny $[0.03,\,0.05]$} \\
 & & MAPPO\_RNN & \highlight{0.74}{$\mathbf{0.74}$ {\tiny $\mathbf{[0.72,\,0.75]}$}} & \highlight{0.22}{$\mathbf{0.22}$ {\tiny $\mathbf{[0.20,\,0.24]}$}} & $0.02$ {\tiny $[0.02,\,0.03]$} & $0.00$ {\tiny $[0.00,\,0.01]$} & $0.03$ {\tiny $[0.02,\,0.03]$} \\
\cmidrule{2-8}
 & \multirow{4}{*}{\texttt{counter\_circuit}} 
   & IPPO\_FF & \highlight{0.72}{$\mathbf{0.72}$ {\tiny $\mathbf{[0.70,\,0.73]}$}} & \highlight{0.19}{$\mathbf{0.19}$ {\tiny $\mathbf{[0.14,\,0.23]}$}} & $0.05$ {\tiny $[0.04,\,0.06]$} & \highlight{0.01}{$\mathbf{0.01}$ {\tiny $\mathbf{[0.01,\,0.01]}$}} & \highlight{0.06}{$\mathbf{0.06}$ {\tiny $\mathbf{[0.05,\,0.08]}$}} \\
 & & IPPO\_RNN & \highlight{0.68}{$\mathbf{0.68}$ {\tiny $\mathbf{[0.61,\,0.73]}$}} & \highlight{0.25}{$\mathbf{0.25}$ {\tiny $\mathbf{[0.19,\,0.31]}$}} & \highlight{0.04}{$\mathbf{0.04}$ {\tiny $\mathbf{[0.03,\,0.07]}$}} & $0.00$ {\tiny $[0.00,\,0.01]$} & \highlight{0.06}{$\mathbf{0.06}$ {\tiny $\mathbf{[0.03,\,0.09]}$}} \\
 & & MAPPO\_FF & \highlight{0.70}{$\mathbf{0.70}$ {\tiny $\mathbf{[0.67,\,0.73]}$}} & \highlight{0.12}{$\mathbf{0.12}$ {\tiny $\mathbf{[0.08,\,0.16]}$}} & $0.04$ {\tiny $[0.02,\,0.05]$} & $0.00$ {\tiny $[0.00,\,0.01]$} & $0.04$ {\tiny $[0.03,\,0.06]$} \\
 & & MAPPO\_RNN & \highlight{0.69}{$\mathbf{0.69}$ {\tiny $\mathbf{[0.65,\,0.72]}$}} & \highlight{0.27}{$\mathbf{0.27}$ {\tiny $\mathbf{[0.24,\,0.29]}$}} & \highlight{0.08}{$\mathbf{0.08}$ {\tiny $\mathbf{[0.05,\,0.10]}$}} & \highlight{0.01}{$\mathbf{0.01}$ {\tiny $\mathbf{[0.01,\,0.01]}$}} & \highlight{0.09}{$\mathbf{0.09}$ {\tiny $\mathbf{[0.06,\,0.12]}$}} \\
\cmidrule{2-8}
 & \multirow{4}{*}{\texttt{cramped\_room}} 
   & IPPO\_FF & \highlight{0.72}{$\mathbf{0.72}$ {\tiny $\mathbf{[0.71,\,0.73]}$}} & \highlight{0.20}{$\mathbf{0.20}$ {\tiny $\mathbf{[0.18,\,0.22]}$}} & $0.03$ {\tiny $[0.03,\,0.04]$} & \highlight{0.01}{$\mathbf{0.01}$ {\tiny $\mathbf{[0.01,\,0.02]}$}} & $0.04$ {\tiny $[0.03,\,0.05]$} \\
 & & IPPO\_RNN & \highlight{0.69}{$\mathbf{0.69}$ {\tiny $\mathbf{[0.65,\,0.73]}$}} & \highlight{0.19}{$\mathbf{0.19}$ {\tiny $\mathbf{[0.13,\,0.25]}$}} & $0.02$ {\tiny $[0.01,\,0.03]$} & $0.01$ {\tiny $[0.00,\,0.01]$} & \highlight{0.03}{$\mathbf{0.03}$ {\tiny $\mathbf{[0.02,\,0.05]}$}} \\
 & & MAPPO\_FF & \highlight{0.72}{$\mathbf{0.72}$ {\tiny $\mathbf{[0.71,\,0.74]}$}} & \highlight{0.19}{$\mathbf{0.19}$ {\tiny $\mathbf{[0.17,\,0.21]}$}} & $0.04$ {\tiny $[0.03,\,0.06]$} & \highlight{0.01}{$\mathbf{0.01}$ {\tiny $\mathbf{[0.01,\,0.02]}$}} & $0.05$ {\tiny $[0.04,\,0.06]$} \\
 & & MAPPO\_RNN & \highlight{0.64}{$\mathbf{0.64}$ {\tiny $\mathbf{[0.50,\,0.72]}$}} & \highlight{0.24}{$\mathbf{0.24}$ {\tiny $\mathbf{[0.18,\,0.29]}$}} & $0.01$ {\tiny $[0.01,\,0.02]$} & \highlight{0.01}{$\mathbf{0.01}$ {\tiny $\mathbf{[0.01,\,0.01]}$}} & $0.02$ {\tiny $[0.01,\,0.02]$} \\
\cmidrule{2-8}
 & \multirow{4}{*}{\texttt{forced\_coord}} 
   & IPPO\_FF & \highlight{0.73}{$\mathbf{0.73}$ {\tiny $\mathbf{[0.72,\,0.74]}$}} & \highlight{0.20}{$\mathbf{0.20}$ {\tiny $\mathbf{[0.19,\,0.21]}$}} & \highlight{0.07}{$\mathbf{0.07}$ {\tiny $\mathbf{[0.06,\,0.08]}$}} & \highlight{0.01}{$\mathbf{0.01}$ {\tiny $\mathbf{[0.01,\,0.02]}$}} & \highlight{0.09}{$\mathbf{0.09}$ {\tiny $\mathbf{[0.07,\,0.10]}$}} \\
 & & IPPO\_RNN & \highlight{0.43}{$\mathbf{0.43}$ {\tiny $\mathbf{[0.24,\,0.61]}$}} & \highlight{0.18}{$\mathbf{0.18}$ {\tiny $\mathbf{[0.10,\,0.26]}$}} & \highlight{0.02}{$\mathbf{0.02}$ {\tiny $\mathbf{[0.01,\,0.03]}$}} & \highlight{0.01}{$\mathbf{0.01}$ {\tiny $\mathbf{[0.00,\,0.01]}$}} & \highlight{0.04}{$\mathbf{0.04}$ {\tiny $\mathbf{[0.02,\,0.06]}$}} \\
 & & MAPPO\_FF & \highlight{0.73}{$\mathbf{0.73}$ {\tiny $\mathbf{[0.72,\,0.75]}$}} & \highlight{0.20}{$\mathbf{0.20}$ {\tiny $\mathbf{[0.17,\,0.22]}$}} & \highlight{0.08}{$\mathbf{0.08}$ {\tiny $\mathbf{[0.07,\,0.08]}$}} & \highlight{0.02}{$\mathbf{0.02}$ {\tiny $\mathbf{[0.01,\,0.02]}$}} & \highlight{0.08}{$\mathbf{0.08}$ {\tiny $\mathbf{[0.08,\,0.09]}$}} \\
 & & MAPPO\_RNN & \highlight{0.71}{$\mathbf{0.71}$ {\tiny $\mathbf{[0.65,\,0.75]}$}} & \highlight{0.36}{$\mathbf{0.36}$ {\tiny $\mathbf{[0.33,\,0.39]}$}} & $0.01$ {\tiny $[0.00,\,0.01]$} & \highlight{0.01}{$\mathbf{0.01}$ {\tiny $\mathbf{[0.01,\,0.02]}$}} & $0.01$ {\tiny $[0.00,\,0.01]$} \\
\cmidrule{2-8}
 & \multirow{4}{*}{\texttt{test\_time\_simple}} 
   & IPPO\_FF & \highlight{0.73}{$\mathbf{0.73}$ {\tiny $\mathbf{[0.71,\,0.74]}$}} & \highlight{0.24}{$\mathbf{0.24}$ {\tiny $\mathbf{[0.19,\,0.27]}$}} & \highlight{0.09}{$\mathbf{0.09}$ {\tiny $\mathbf{[0.07,\,0.11]}$}} & \highlight{0.01}{$\mathbf{0.01}$ {\tiny $\mathbf{[0.01,\,0.01]}$}} & \highlight{0.10}{$\mathbf{0.10}$ {\tiny $\mathbf{[0.08,\,0.11]}$}} \\
 & & IPPO\_RNN & \highlight{0.63}{$\mathbf{0.63}$ {\tiny $\mathbf{[0.57,\,0.69]}$}} & \highlight{0.24}{$\mathbf{0.24}$ {\tiny $\mathbf{[0.19,\,0.28]}$}} & \highlight{0.08}{$\mathbf{0.08}$ {\tiny $\mathbf{[0.06,\,0.11]}$}} & $0.01$ {\tiny $[0.00,\,0.01]$} & \highlight{0.09}{$\mathbf{0.09}$ {\tiny $\mathbf{[0.07,\,0.11]}$}} \\
 & & MAPPO\_FF & \highlight{0.72}{$\mathbf{0.72}$ {\tiny $\mathbf{[0.71,\,0.73]}$}} & \highlight{0.26}{$\mathbf{0.26}$ {\tiny $\mathbf{[0.25,\,0.27]}$}} & \highlight{0.10}{$\mathbf{0.10}$ {\tiny $\mathbf{[0.09,\,0.11]}$}} & \highlight{0.01}{$\mathbf{0.01}$ {\tiny $\mathbf{[0.01,\,0.01]}$}} & \highlight{0.12}{$\mathbf{0.12}$ {\tiny $\mathbf{[0.10,\,0.13]}$}} \\
 & & MAPPO\_RNN & \highlight{0.61}{$\mathbf{0.61}$ {\tiny $\mathbf{[0.59,\,0.64]}$}} & \highlight{0.28}{$\mathbf{0.28}$ {\tiny $\mathbf{[0.25,\,0.31]}$}} & \highlight{0.09}{$\mathbf{0.09}$ {\tiny $\mathbf{[0.05,\,0.14]}$}} & $0.01$ {\tiny $[0.01,\,0.01]$} & \highlight{0.13}{$\mathbf{0.13}$ {\tiny $\mathbf{[0.09,\,0.18]}$}} \\
\cmidrule{2-8}
 & \multirow{4}{*}{\texttt{test\_time\_wide}} 
   & IPPO\_FF & \highlight{0.70}{$\mathbf{0.70}$ {\tiny $\mathbf{[0.67,\,0.72]}$}} & \highlight{0.19}{$\mathbf{0.19}$ {\tiny $\mathbf{[0.18,\,0.21]}$}} & \highlight{0.10}{$\mathbf{0.10}$ {\tiny $\mathbf{[0.09,\,0.10]}$}} & \highlight{0.01}{$\mathbf{0.01}$ {\tiny $\mathbf{[0.01,\,0.01]}$}} & \highlight{0.11}{$\mathbf{0.11}$ {\tiny $\mathbf{[0.10,\,0.12]}$}} \\
 & & IPPO\_RNN & \highlight{0.58}{$\mathbf{0.58}$ {\tiny $\mathbf{[0.51,\,0.64]}$}} & \highlight{0.21}{$\mathbf{0.21}$ {\tiny $\mathbf{[0.15,\,0.28]}$}} & $0.03$ {\tiny $[0.01,\,0.04]$} & $0.00$ {\tiny $[0.00,\,0.01]$} & \highlight{0.05}{$\mathbf{0.05}$ {\tiny $\mathbf{[0.03,\,0.07]}$}} \\
 & & MAPPO\_FF & \highlight{0.73}{$\mathbf{0.73}$ {\tiny $\mathbf{[0.72,\,0.73]}$}} & \highlight{0.21}{$\mathbf{0.21}$ {\tiny $\mathbf{[0.19,\,0.23]}$}} & \highlight{0.10}{$\mathbf{0.10}$ {\tiny $\mathbf{[0.09,\,0.11]}$}} & \highlight{0.01}{$\mathbf{0.01}$ {\tiny $\mathbf{[0.01,\,0.01]}$}} & \highlight{0.11}{$\mathbf{0.11}$ {\tiny $\mathbf{[0.10,\,0.13]}$}} \\
 & & MAPPO\_RNN & \highlight{0.64}{$\mathbf{0.64}$ {\tiny $\mathbf{[0.62,\,0.66]}$}} & \highlight{0.28}{$\mathbf{0.28}$ {\tiny $\mathbf{[0.25,\,0.30]}$}} & \highlight{0.10}{$\mathbf{0.10}$ {\tiny $\mathbf{[0.08,\,0.12]}$}} & $0.01$ {\tiny $[0.01,\,0.01]$} & \highlight{0.13}{$\mathbf{0.13}$ {\tiny $\mathbf{[0.10,\,0.16]}$}} \\
\cmidrule{2-8}
 & \multirow{4}{*}{\texttt{demo\_cook\_simple}} 
   & IPPO\_FF & \highlight{0.71}{$\mathbf{0.71}$ {\tiny $\mathbf{[0.70,\,0.72]}$}} & \highlight{0.22}{$\mathbf{0.22}$ {\tiny $\mathbf{[0.21,\,0.23]}$}} & \highlight{0.10}{$\mathbf{0.10}$ {\tiny $\mathbf{[0.09,\,0.11]}$}} & \highlight{0.01}{$\mathbf{0.01}$ {\tiny $\mathbf{[0.01,\,0.01]}$}} & \highlight{0.11}{$\mathbf{0.11}$ {\tiny $\mathbf{[0.10,\,0.11]}$}} \\
 & & IPPO\_RNN & \highlight{0.52}{$\mathbf{0.52}$ {\tiny $\mathbf{[0.48,\,0.55]}$}} & \highlight{0.21}{$\mathbf{0.21}$ {\tiny $\mathbf{[0.18,\,0.24]}$}} & $0.02$ {\tiny $[0.02,\,0.03]$} & $0.00$ {\tiny $[0.00,\,0.00]$} & \highlight{0.03}{$\mathbf{0.03}$ {\tiny $\mathbf{[0.02,\,0.04]}$}} \\
 & & MAPPO\_FF & \highlight{0.70}{$\mathbf{0.70}$ {\tiny $\mathbf{[0.67,\,0.72]}$}} & \highlight{0.21}{$\mathbf{0.21}$ {\tiny $\mathbf{[0.18,\,0.24]}$}} & \highlight{0.10}{$\mathbf{0.10}$ {\tiny $\mathbf{[0.08,\,0.12]}$}} & $0.01$ {\tiny $[0.01,\,0.01]$} & \highlight{0.10}{$\mathbf{0.10}$ {\tiny $\mathbf{[0.08,\,0.12]}$}} \\
 & & MAPPO\_RNN & \highlight{0.57}{$\mathbf{0.57}$ {\tiny $\mathbf{[0.56,\,0.59]}$}} & \highlight{0.23}{$\mathbf{0.23}$ {\tiny $\mathbf{[0.22,\,0.25]}$}} & \highlight{0.11}{$\mathbf{0.11}$ {\tiny $\mathbf{[0.09,\,0.14]}$}} & $0.01$ {\tiny $[0.01,\,0.01]$} & \highlight{0.13}{$\mathbf{0.13}$ {\tiny $\mathbf{[0.11,\,0.15]}$}} \\
\cmidrule{2-8}
 & \multirow{4}{*}{\texttt{demo\_cook\_wide}} 
   & IPPO\_FF & \highlight{0.72}{$\mathbf{0.72}$ {\tiny $\mathbf{[0.71,\,0.73]}$}} & \highlight{0.18}{$\mathbf{0.18}$ {\tiny $\mathbf{[0.16,\,0.19]}$}} & \highlight{0.09}{$\mathbf{0.09}$ {\tiny $\mathbf{[0.08,\,0.10]}$}} & \highlight{0.01}{$\mathbf{0.01}$ {\tiny $\mathbf{[0.01,\,0.01]}$}} & \highlight{0.09}{$\mathbf{0.09}$ {\tiny $\mathbf{[0.08,\,0.10]}$}} \\
 & & IPPO\_RNN & \highlight{0.56}{$\mathbf{0.56}$ {\tiny $\mathbf{[0.51,\,0.62]}$}} & \highlight{0.18}{$\mathbf{0.18}$ {\tiny $\mathbf{[0.11,\,0.24]}$}} & \highlight{0.06}{$\mathbf{0.06}$ {\tiny $\mathbf{[0.03,\,0.08]}$}} & $0.00$ {\tiny $[0.00,\,0.00]$} & \highlight{0.06}{$\mathbf{0.06}$ {\tiny $\mathbf{[0.04,\,0.08]}$}} \\
 & & MAPPO\_FF & \highlight{0.72}{$\mathbf{0.72}$ {\tiny $\mathbf{[0.71,\,0.72]}$}} & \highlight{0.18}{$\mathbf{0.18}$ {\tiny $\mathbf{[0.17,\,0.20]}$}} & \highlight{0.09}{$\mathbf{0.09}$ {\tiny $\mathbf{[0.08,\,0.10]}$}} & \highlight{0.01}{$\mathbf{0.01}$ {\tiny $\mathbf{[0.01,\,0.01]}$}} & \highlight{0.09}{$\mathbf{0.09}$ {\tiny $\mathbf{[0.09,\,0.10]}$}} \\
 & & MAPPO\_RNN & \highlight{0.61}{$\mathbf{0.61}$ {\tiny $\mathbf{[0.57,\,0.65]}$}} & \highlight{0.23}{$\mathbf{0.23}$ {\tiny $\mathbf{[0.21,\,0.26]}$}} & \highlight{0.12}{$\mathbf{0.12}$ {\tiny $\mathbf{[0.10,\,0.15]}$}} & \highlight{0.01}{$\mathbf{0.01}$ {\tiny $\mathbf{[0.01,\,0.01]}$}} & \highlight{0.14}{$\mathbf{0.14}$ {\tiny $\mathbf{[0.11,\,0.16]}$}} \\
\pagebreak[4]
 & \multirow{4}{*}{\texttt{grounded\_coord\_ring}} 
   & IPPO\_FF & \highlight{0.71}{$\mathbf{0.71}$ {\tiny $\mathbf{[0.70,\,0.73]}$}} & \highlight{0.23}{$\mathbf{0.23}$ {\tiny $\mathbf{[0.22,\,0.25]}$}} & \highlight{0.07}{$\mathbf{0.07}$ {\tiny $\mathbf{[0.06,\,0.08]}$}} & $0.00$ {\tiny $[0.00,\,0.01]$} & \highlight{0.08}{$\mathbf{0.08}$ {\tiny $\mathbf{[0.08,\,0.09]}$}} \\
 & & IPPO\_RNN & \highlight{0.47}{$\mathbf{0.47}$ {\tiny $\mathbf{[0.33,\,0.59]}$}} & \highlight{0.12}{$\mathbf{0.12}$ {\tiny $\mathbf{[0.09,\,0.15]}$}} & \highlight{0.11}{$\mathbf{0.11}$ {\tiny $\mathbf{[0.06,\,0.16]}$}} & $0.00$ {\tiny $[0.00,\,0.01]$} & \highlight{0.13}{$\mathbf{0.13}$ {\tiny $\mathbf{[0.08,\,0.19]}$}} \\
 & & MAPPO\_FF & \highlight{0.70}{$\mathbf{0.70}$ {\tiny $\mathbf{[0.66,\,0.72]}$}} & \highlight{0.24}{$\mathbf{0.24}$ {\tiny $\mathbf{[0.23,\,0.25]}$}} & \highlight{0.08}{$\mathbf{0.08}$ {\tiny $\mathbf{[0.07,\,0.08]}$}} & $0.00$ {\tiny $[0.00,\,0.01]$} & \highlight{0.09}{$\mathbf{0.09}$ {\tiny $\mathbf{[0.08,\,0.10]}$}} \\
 & & MAPPO\_RNN & \highlight{0.62}{$\mathbf{0.62}$ {\tiny $\mathbf{[0.60,\,0.64]}$}} & \highlight{0.29}{$\mathbf{0.29}$ {\tiny $\mathbf{[0.27,\,0.30]}$}} & \highlight{0.11}{$\mathbf{0.11}$ {\tiny $\mathbf{[0.09,\,0.13]}$}} & $0.00$ {\tiny $[0.00,\,0.01]$} & \highlight{0.15}{$\mathbf{0.15}$ {\tiny $\mathbf{[0.13,\,0.17]}$}} \\
\cmidrule{2-8} 
 & \multirow{4}{*}{\texttt{grounded\_coord\_simple}} 
   & IPPO\_FF & \highlight{0.72}{$\mathbf{0.72}$ {\tiny $\mathbf{[0.72,\,0.73]}$}} & \highlight{0.26}{$\mathbf{0.26}$ {\tiny $\mathbf{[0.25,\,0.27]}$}} & \highlight{0.11}{$\mathbf{0.11}$ {\tiny $\mathbf{[0.10,\,0.12]}$}} & \highlight{0.01}{$\mathbf{0.01}$ {\tiny $\mathbf{[0.01,\,0.01]}$}} & \highlight{0.12}{$\mathbf{0.12}$ {\tiny $\mathbf{[0.10,\,0.13]}$}} \\
 & & IPPO\_RNN & \highlight{0.62}{$\mathbf{0.62}$ {\tiny $\mathbf{[0.52,\,0.70]}$}} & \highlight{0.24}{$\mathbf{0.24}$ {\tiny $\mathbf{[0.18,\,0.28]}$}} & \highlight{0.07}{$\mathbf{0.07}$ {\tiny $\mathbf{[0.05,\,0.09]}$}} & $0.01$ {\tiny $[0.00,\,0.01]$} & \highlight{0.08}{$\mathbf{0.08}$ {\tiny $\mathbf{[0.05,\,0.11]}$}} \\
 & & MAPPO\_FF & \highlight{0.72}{$\mathbf{0.72}$ {\tiny $\mathbf{[0.71,\,0.73]}$}} & \highlight{0.25}{$\mathbf{0.25}$ {\tiny $\mathbf{[0.23,\,0.27]}$}} & \highlight{0.10}{$\mathbf{0.10}$ {\tiny $\mathbf{[0.08,\,0.11]}$}} & \highlight{0.01}{$\mathbf{0.01}$ {\tiny $\mathbf{[0.01,\,0.01]}$}} & \highlight{0.11}{$\mathbf{0.11}$ {\tiny $\mathbf{[0.09,\,0.12]}$}} \\
 & & MAPPO\_RNN & \highlight{0.65}{$\mathbf{0.65}$ {\tiny $\mathbf{[0.62,\,0.68]}$}} & \highlight{0.29}{$\mathbf{0.29}$ {\tiny $\mathbf{[0.27,\,0.31]}$}} & \highlight{0.11}{$\mathbf{0.11}$ {\tiny $\mathbf{[0.09,\,0.14]}$}} & $0.01$ {\tiny $[0.01,\,0.01]$} & \highlight{0.16}{$\mathbf{0.16}$ {\tiny $\mathbf{[0.13,\,0.18]}$}} \\
\end{longtable}

\end{document}